\documentclass[10pt,twocolumn,letterpaper]{article}

\usepackage[pagenumbers]{wacv25toolkit/wacv} 

\usepackage{graphicx}
\usepackage{amsmath}
\usepackage{amssymb}
\usepackage{booktabs}
\usepackage{microtype}

\usepackage[pagebackref,breaklinks,colorlinks]{hyperref}

\usepackage{graphicx}
\usepackage{booktabs}
\usepackage[accsupp]{axessibility}

\usepackage{xcolor}
\usepackage{amsmath,amssymb,amsfonts}
\usepackage{multirow,multicol}
\usepackage{algorithmic}
\usepackage{textcomp}
\usepackage{xspace}
\usepackage{url}
\usepackage{dsfont}
\usepackage{balance}
\usepackage{soul}
\usepackage{todonotes}
\usepackage{pifont}
\usepackage{colortbl}
\usepackage{overpic}
\usepackage{enumitem}
\usepackage{nccmath}
\usepackage{courier}
\usepackage{array}
\usepackage{tabularray}
\usepackage{microtype}
\usepackage{makecell}
\usepackage{tikz}
\usepackage{pgfplots}
\usepackage{bbding}
\usepackage{wasysym}

\pgfplotsset{compat=1.18}
\usepackage{orcidlink}
\usepackage{subcaption}

\usepackage[capitalize]{cleveref}
\crefname{section}{Sec.}{Secs.}
\Crefname{section}{Section}{Sections}
\Crefname{table}{Table}{Tables}
\crefname{table}{Tab.}{Tabs.}

\begin{document}

    \newcommand{\cmark}{\ding{51}}
\newcommand{\xmark}{\ding{55}}
\newcommand{\warning}[1]{\textbf{\color{red!90}{#1}}}
\newcommand{\myparagraph}[1]{\vspace{5pt}\noindent\textbf{{#1}}}

\definecolor{myblue}{rgb}{0.0000,0.7490,1.0000}
\definecolor{myyellow}{rgb}{1.0000,0.7530,0.0000}
\definecolor{mygray}{rgb}{0.9000,0.9000,0.9000}
\definecolor{myazure}{rgb}{0.8509,0.8980,0.9412}
\definecolor{mygreen}{rgb}{0.0,0.6,0.0}
\definecolor{mylightyellow}{rgb}{1.0000,0.9294,0.6902}

\definecolor{part_blue}{rgb}{0.12156862745098039, 0.4666666666666667, 0.7058823529411765}
\definecolor{part_orange}{rgb}{1.0, 0.4980392156862745, 0.054901960784313725}
\definecolor{part_green}{rgb}{0.17254901960784313, 0.6274509803921569, 0.17254901960784313}
\definecolor{part_red}{rgb}{0.8392156862745098, 0.15294117647058825, 0.1568627450980392}

\newcommand{\acronym}{COPS\xspace}
\newcommand{\spacing}{\vspace{0.8pt}}

    \def\eg{\emph{e.g}\onedot,\xspace} \def\Eg{\emph{E.g}\onedot,\xspace}
    \def\ie{\emph{i.e}\onedot,\xspace} \def\Ie{\emph{I.e}\onedot,\xspace}

    \title{3D Part Segmentation via Geometric Aggregation of 2D Visual Features}

    \author{
        \begin{minipage}[t]{0.33\textwidth}
        \centering
        Marco Garosi\\
        University of Trento\\
        {\tt\small marco.garosi@unitn.it}
        \end{minipage}%
        \begin{minipage}[t]{0.33\textwidth}
        \centering
        Riccardo Tedoldi\\
        University of Trento\\
        {\tt\small riccardo.tedoldi@unitn.it}
        \end{minipage}
        \begin{minipage}[t]{0.33\textwidth}
        \centering
        Davide Boscaini\\
        Fondazione Bruno Kessler\\
        {\tt\small dboscaini@fbk.eu}
        \end{minipage}
        \and
        \begin{minipage}[t]{0.33\textwidth}
        \centering
        Massimiliano Mancini\\
        University of Trento\\
        {\tt\small massimiliano.mancini@unitn.it}
        \end{minipage}
        \begin{minipage}[t]{0.33\textwidth}
        \centering
        Nicu Sebe\\
        University of Trento\\
        {\tt\small sebe@disi.unitn.it}
        \end{minipage}
        \begin{minipage}[t]{0.33\textwidth}
        \centering
        Fabio Poiesi\\
        Fondazione Bruno Kessler\\
        {\tt\small poiesi@fbk.eu}
        \end{minipage}
    }

    \maketitle

    \begin{abstract}
Supervised 3D part segmentation models are tailored for a fixed set of objects and parts, limiting their transferability to open-set, real-world scenarios.
Recent works have explored vision-language models (VLMs) as a promising alternative, using multi-view rendering and textual prompting to identify object parts.
However, naively applying VLMs in this context introduces several drawbacks, such as the need for meticulous prompt engineering, and fails to leverage the 3D geometric structure of objects.
To address these limitations, we propose \acronym, a COmprehensive model for Parts Segmentation that blends the semantics extracted from visual concepts and 3D geometry to effectively identify object parts.
\acronym renders a point cloud from multiple viewpoints, extracts 2D features, projects them back to 3D, and uses a novel geometric-aware feature aggregation procedure to ensure spatial and semantic consistency.
Finally, it clusters points into parts and labels them.
We demonstrate that \acronym is efficient, scalable, and achieves zero-shot state-of-the-art performance across five datasets, covering synthetic and real-world data, texture-less and coloured objects, as well as rigid and non-rigid shapes.
The code is available at \href{https://3d-cops.github.io}{https://3d-cops.github.io}.
\end{abstract}
    \section{Introduction}\label{sec:intro}

\begin{figure*}[t]
\centering
\begin{overpic}[width=0.92\textwidth,trim=210 0 210 90,clip]{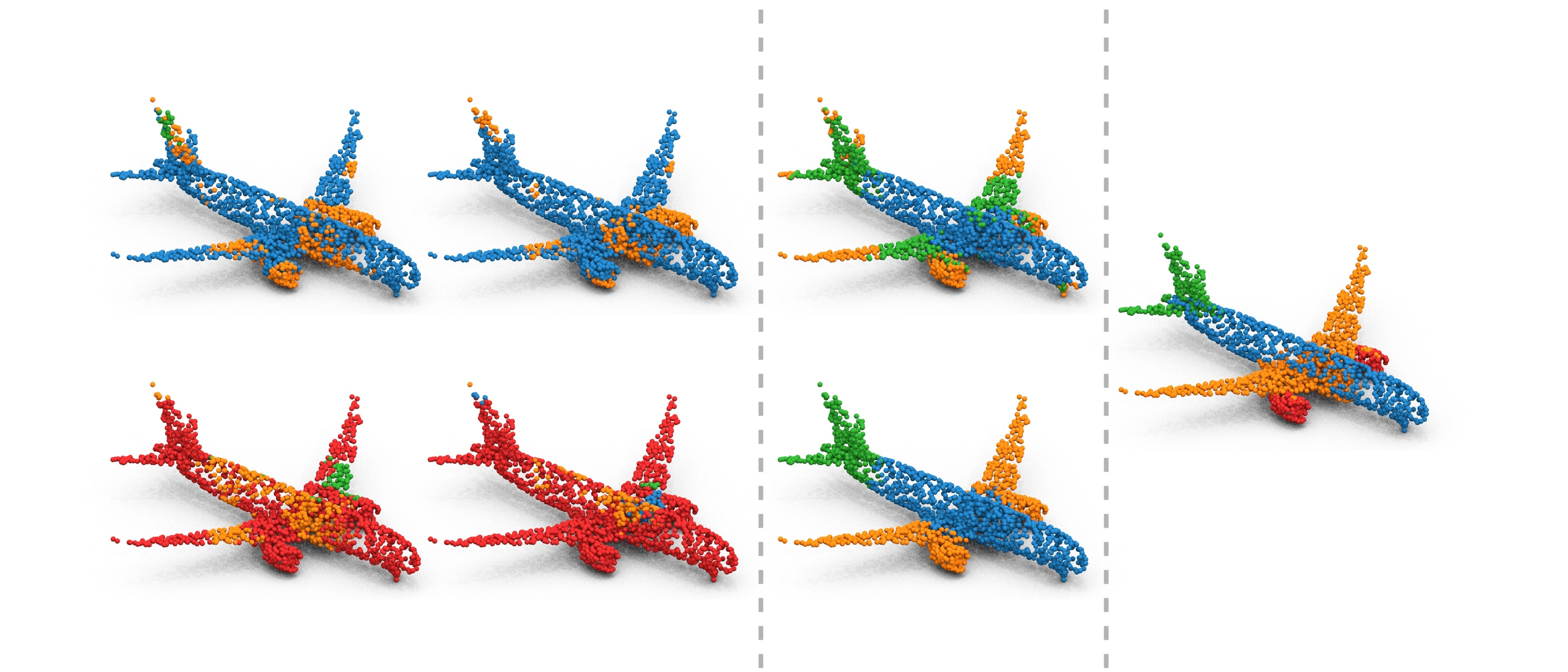}
    \put(16,43){PointCLIPv2~\cite{zhu_pointclip_2023}}
    \put(6,25){Original prompts}
    \put(7.5,22.5){mIoU = 18.7}
    \put(30.5,25){GPT-generated}
    \put(31.5,22.5){mIoU = 14.5}
    \put(5.5,4){Template-based}
    \put(7.5,1.5){mIoU = 9.9}
    \put(29.3,4){Part names only}
    \put(31.5,1.5){mIoU = 5.1}
    \put(55.5,43){\acronym (ours)}
    \put(52,25){CLIP visual features}
    \put(55.5,22.5){mIoU = 24.3}
    \put(53.6,4){DINOv2 features}
    \put(55.5,1.5){mIoU = 41.9}
    \put(80,43){Ground-truth}
    \put(75,14){{\color{part_blue} \large $\bullet$} Fuselage}
    \put(75,11){{\color{part_orange} \large $\bullet$} Wing}
    \put(87.5,14){{\color{part_red} \large $\bullet$} Engine}
    \put(87.5,11){{\color{part_green} \large $\bullet$} Tail}
\end{overpic}

\vspace{-3mm}
\caption{
    The quality of part descriptions significantly affects the segmentation performance of methods based on vision-language models.
    For example, the performance of PointCLIPv2~\cite{zhu_pointclip_2023} (left) deteriorates rapidly when replacing the default textual prompt with a GPT-generated description, the template ``This is a depth image of an airplane's [part]'', or simply using part names.
    In contrast, our pipeline (center) achieves more accurate segmentations by disentangling part decomposition from part classification.
    The improvement is evident when using the same CLIP visual features as PointCLIPv2 (top) and becomes even more pronounced when using DINOv2~\cite{oquab2023dinov2} features (bottom), the default choice of \acronym.
    \acronym generates more uniform segments with sharper boundaries, resulting in higher segmentation quality.
}%
\label{fig:teaser}
\end{figure*}
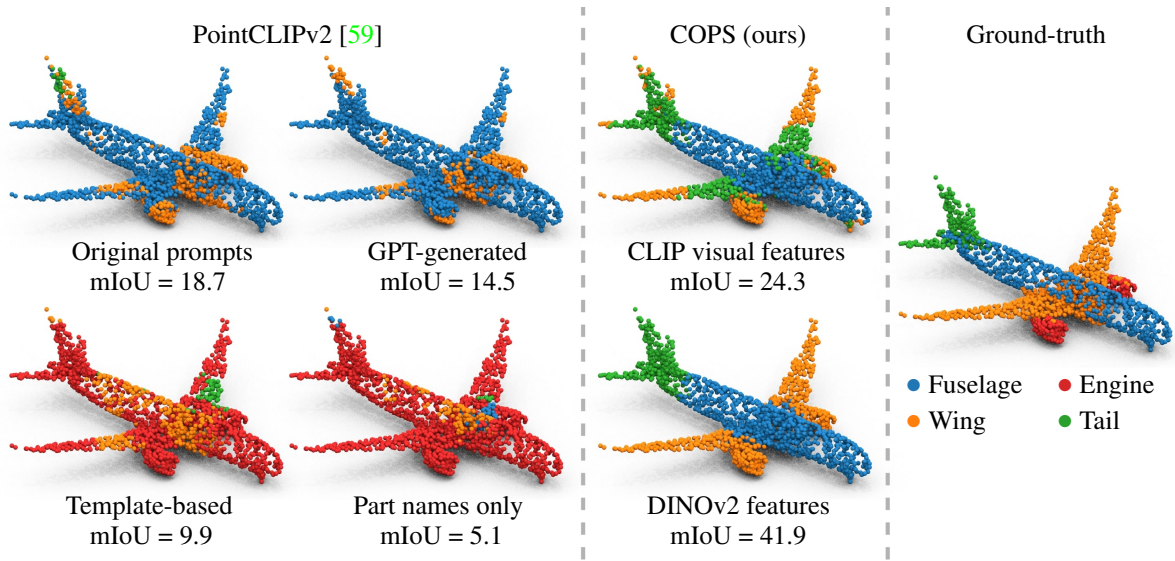

Recognising object parts has raised widespread attention in both industry and academia, with applications ranging from robotic manipulation~\cite{gao2023physically, brohan2023rt2} to augmented reality~\cite{chen2024spatialvlm}.
Understanding and interacting with objects requires identifying their building components. For example, we open a bottle by detecting its cap, or hold a mug by its handle.

Currently, well-annotated 3D part segmentation datasets contain only a limited amount of samples in a restricted number of categories.
This limits the transferability of models, as their knowledge is narrowed to the closed-set parts and objects seen during training.
In contrast, vision-language models (VLMs) can handle objects and parts that were not seen during training, making them more versatile in open-vocabulary settings~\cite{radford2021learning, li2023blip2, li2022grounded}.

In this context, methods such as PointCLIP~\cite{zhang_pointclip_2021, zhu_pointclip_2023}, PartSLIP~\cite{liu_partslip_2023, zhou2023partslip}, ZeroPS~\cite{xue2023zerops}, and SATR~\cite{abdelreheem2023satr} leverage VLMs like CLIP~\cite{radford2021learning} or GLIP~\cite{li2022grounded} in a two-step procedure to tackle open-vocabulary settings.
First, a point cloud is rendered into multiple views, which are then processed by the VLM's visual encoder to extract visual features.
Second, these visual features are projected back into 3D and compared with the text features derived from prompts describing object parts to produce the final segmentation.
This approach suffers from two drawbacks: (i) it does not incorporate 3D geometric knowledge, and (ii) its performance depends on the quality of the input prompts (\cref{fig:teaser}), as some parts can be challenging to describe in natural language.
The latter issue can be partially addressed by crafting better textual prompts with the assistance of large language models such as ChatGPT~\cite{liu_partslip_2023, zhou2023partslip}, or through prompt ensembles to reduce description ambiguity~\cite{radford2021learning}.
However, the inherent ambiguity of natural language remains a significant challenge, particularly when generalising to unseen objects and parts.

In this paper, we blend features from a 2D vision foundation model (VFM) with 3D geometric understanding to detect object parts.
\acronym follows a five-step procedure that involves:
(i) rendering the object from multiple views that cover a large as possible portion of the object,
(ii) extracting pixel-level features using DINOv2~\cite{oquab2023dinov2} and upsampling its output,
(iii) back-projecting these features onto the points,
(iv) aggregating point-level features from multiple views using the geometric feature aggregation (GFA) module, and
(v) segmenting object parts through unsupervised clustering of point-level features, followed by zero-shot classification.
For tasks requiring only part decomposition without knowledge of part categories, the zero-shot classifier of \acronym can be disabled to reduce computational costs and avoid inaccuracies caused by natural language ambiguities~\cite{udandarao2023sus, NEURIPS2022_702f4db7}.

In summary, our contributions are:
\begin{itemize}[noitemsep,nolistsep,leftmargin=*]
\item A novel training-free method for 3D part segmentation, which disentangles part segmentation from semantic labelling;
\item A geometric feature aggregation module that fuses 3D structural information with semantic knowledge from VFMs;
\item A thorough evaluation on five benchmarks, establishing a unified evaluation that future zero-shot part segmentation methods can use for comparison.
\end{itemize}

    \section{Related works}\label{sec:related}

We identify two main approaches to 3D point cloud part segmentation: using networks specifically designed to handle 3D data natively and adapting 2D VLMs to the 3D task.

\noindent\textbf{3D networks for point cloud part segmentation.}
A general approach to point cloud part segmentation~\cite{Bogo:CVPR:2014, yi2016shapenetpart, Mo_2019_CVPR, uy2019scanobjectnn} is to leverage point-based methods~\cite{qi2017pointnet, qi2017pointnetplus} that extract a semantic representation for each point.
These works focus either on supervised or self-supervised methods~\cite{Boscaini2015, NEURIPS2018_f5f8590c, masci2018geodesic, veličković2018graph, Guo2021, BOSCAINI2023104768}, and they produce feature representations that transfer well to the part segmentation task~\cite{qi2017pointnet, qi2017pointnetplus, simonovsky2017dynamic, thomas2019kpconv}.
Other works focus directly on part segmentation. They group points belonging to the same part by leveraging different strategies.
The most common techniques are prototype-based, modelling and matching shape parts~\cite{he2020learning, yi2019gspn, zhao2022number, qin2024unified}, grouping~\cite{wang2018sgpn, jiang2020pointgroup, vu2022softgroup, zhang2021point, liu2020self}, and directly training for point clustering.
However, datasets with 3D annotated shapes are scarce and costly to produce. Therefore, a thread of works explored the use of alternative supervision signals, such as bounding boxes~\cite{chibane2022box2mask}, assembly manuals~\cite{wang2022ikea}, and reference games~\cite{koo2022partglot}. Similarly, other works introduced unsupervised objectives, such as learning hierarchies~\cite{niu2022rim}, co-segmentation \cite{zhu2020adacoseg, yang2021unsupervised, chen2019bae}, abstraction \cite{tulsiani2017learning, paschalidou2019superquadrics, paschalidou2021neural, genova2020local, deng2020cvxnet}.
In addition to data scarcity, some methods have to face costly training procedures, which results in limited transfer learning capabilities and robustness to domain gaps and unseen objects. With \acronym, we overcome all these limitations by offering a training-free, efficient, and versatile solution.

\noindent\textbf{Adapting 2D VLMs to 3D tasks.}
Several VLM-based methods aim to segment 3D objects into parts with a three-stage approach:
(i) generating multiple 2D views of the object,
(ii) extracting the features from each view,
(iii) projecting 2D features onto 3D points.
However, aligning visual and text modalities requires carefully tuned prompts, which must be effectively matched to visual prompts.
For example, PointCLIP~\cite{zhang_pointclip_2021} and PointCLIPv2~\cite{zhu_pointclip_2023} utilise CLIP~\cite{radford2021learning} for segmenting objects into parts.
DILF~\cite{ning_dilf_2024} extended PointCLIPv2~\cite{zhu_pointclip_2023} by incorporating a fully differentiable renderer to learn optimal camera poses for specific datasets.
In contrast to these methods, the segmentation quality of \acronym does not depend on prompts, as we utilise them only to label already-segmented parts.
PartSLIP~\cite{liu_partslip_2023} is a bounding-box-based framework for part segmentation that uses GLIP~\cite{li2022grounded}. Its successor~\cite{zhou2023partslip} introduces SAM~\cite{kirillov2023segment} and an expectation-maximisation algorithm to improve predictions. Subsequent works~\cite{umam2023partdistill, xue2023zerops, abdelreheem2023satr} build upon PartSLIP to enhance the part segmentation quality.
These detection-based methods rely on bounding boxes to back-project categorical labels to the points, while \acronym projects features to points.

    \section{Our approach}\label{sec:method}

\begin{figure*}[t]
\centering
\begin{overpic}[width=0.95\textwidth,trim=400 0 220 80,clip]{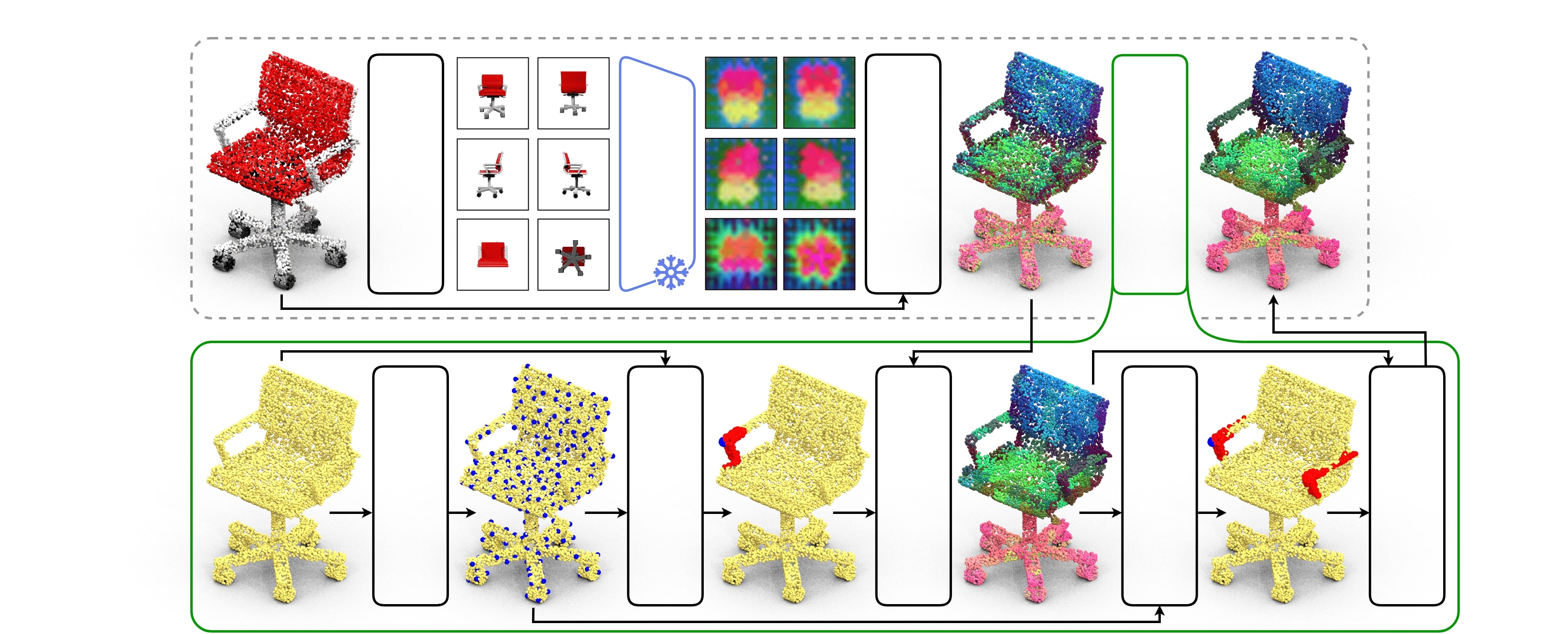}
    %
    %
    \put(-2.5,28.7){\rotatebox{90}{\color{gray}{Feat. extractor $\Phi$}}}
    \put(2,42.5){ $\mathcal{X}$}
    \put(80,42.5){ $\mathcal{F}$}
    \put(14.8,32){\rotatebox{90}{3D-to-2D}}
    \put(17.1,31.5){\rotatebox{90}{projection}}
    \put(34.5,32.8){\rotatebox{90}{Frozen}}
    \put(37,31.8){\rotatebox{90}{DINOv2}}
    \put(53.7,31.7){\rotatebox{90}{2D-to-3D}}
    \put(56.1,31.6){\rotatebox{90}{back-proj.}}
    \put(73,31.1){\rotatebox{90}{Geometric}}
    \put(75.3,30.1){\rotatebox{90}{feature aggr.}}
    %
    %
    \put(-2.5,6.5){\rotatebox{90}{\color{gray}{\color{mygreen} GFA details}}}
    \put(1.5,17.5){ $\mathcal{X}^\textrm{P}$}
    \put(15,5.5){\rotatebox{90}{Farthest point}}
    \put(17.3,7.3){\rotatebox{90}{sampling}}
    \put(35.3,8.5){\rotatebox{90}{Spatial}}
    \put(37.6,7.5){\rotatebox{90}{attention}}
    \put(54.6,5.5){\rotatebox{90}{Spatially-cons.}}
    \put(56.9,6.1){\rotatebox{90}{feature aggr.}}
    \put(73.9,7.5){\rotatebox{90}{Semantic}}
    \put(76.2,7.6){\rotatebox{90}{attention}}
    \put(93.2,4.7){\rotatebox{90}{Semantic.-cons.}}
    \put(95.5,5.8){\rotatebox{90}{feature aggr.}}
\end{overpic}

\vspace{-2mm}
\caption{
    Overview of \acronym's feature extractor.
    $\Phi$ (top) extracts point-level features by (i) rendering multiple views of the object, (ii) processing them with DINOv2, (iii) lifting them in 3D.
    The Geometric Feature Aggregation module (GFA, bottom) further refines these features by extracting super points (blue points in the second row) and their neighbouring points (red points in the second row) to obtain spatially consistent centroids.
    These centroids are used to perform spatial- and semantic-consistent feature aggregation, ensuring that the features are both locally consistent and similar across large distances when describing the same part (\eg the armrest).
    \vspace{-5pt}
}%
\label{fig:diagram}
\vspace{-3mm}
\end{figure*}
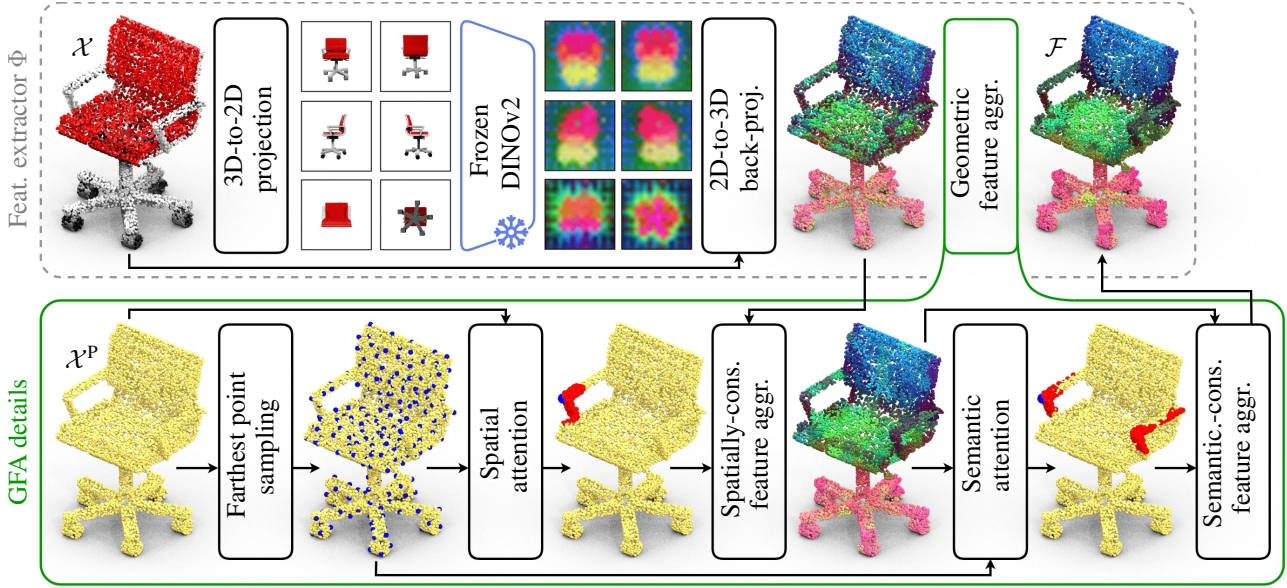

\acronym consists of five-stages, as depicted in \cref{fig:diagram}.
First, the point cloud is rendered into multiple 2D images.
Next, we extract features with a 2D vision foundation model and subsequently project them back onto 3D points, obtaining point-level semantic features.
These steps eliminate the need for models specifically crafted and trained for 3D tasks, thus avoiding the need for annotated data.
We introduce a geometric feature aggregation module, or GFA, that combines visual features from geometrically neighbouring and semantically similar points to produce novel features that incorporate both the point's semantic (\ie what part it belongs to) and geometry (\ie how it relates to other points).
Lastly, we attain open-vocabulary by integrating predictions from a VLM, which assigns a semantic label to each part. Differently from our competitors~\cite{zhang_pointclip_2021, zhu_pointclip_2023}, we do not require meticulously crafted prompts, as the VLM is leveraged for part-level labelling, rather than point-level labelling.

\subsection{Problem formulation}
Given a point cloud of an object and a list of parts, our goal is to segment the point cloud into its constituent parts and label them according to the given list.
Formally, let
$\mathcal{X} = \left( \mathcal{X}^\text{P} \,|\, \mathcal{X}^\text{C} \right) \in \mathbb{R}^{N\times6}$
denote a point cloud with $N$ vertices, where $\mathcal{X}^\text{P} \in \mathbb{R}^{N\times3}$ encodes positional information as XYZ coordinates, $\mathcal{X}^\text{C} \in \mathbb{R}^{N\times3}$ encodes photometric information (when available) as RGB colours, and $|$ denotes concatenation along the last dimension.
We assume $\mathcal{X}$ to have $P$ parts, \ie $\mathcal{X} = \bigcup_{p=1}^P \mathcal{X}_{p}$, and we denote the set of point-level part labels as $\mathcal{Y} \in \{1, \dots, P\}^{N}$.
\acronym consists of two main components: a feature extractor $\Phi$, and a segmenter $\Psi$.
$\Phi \colon \mathcal{X} \mapsto \mathcal{F} \in \mathbb{R}^{N \times d}$ is a function that maps input points into $d$-dimensional features.
Next, $\Psi \colon \mathcal{F} \mapsto \tilde{\mathcal{Y}} \in \{1, \dots, P\}^{N}$ converts point-wise features into categorical labels that identify the input's parts.
Following~\cite{zhang_pointclip_2021, zhu_pointclip_2023}, we measure the segmentation quality as the mean intersection-over-union (mIoU) between our prediction $\tilde{\mathcal{Y}}$ and the ground truth $\mathcal{Y}$.

\vspace{-2mm}
\subsection{Feature extraction}
\vspace{-2mm}
The feature extractor $\Phi$ consists of two parts: a feature encoder $\Phi_{\mathcal{E}}$ and a geometric feature aggregation module $\Phi_{\mathtt{GFA}}$. Thus, $\Phi = \Phi_{\mathtt{GFA}} \circ \Phi_{\mathcal{E}}$. This section discusses $\Phi_{\mathcal{E}}$, while the following one presents $\Phi_{\mathtt{GFA}}$.
$\Phi_{\mathcal{E}}$ comprises three parts: (i) point cloud rasterization, which turns the input into a collection of 2D images; (ii) 2D feature extraction; and (iii) feature back-projection onto 3D points.

\noindent\textbf{Point cloud rasterization.}
One of the goals of \acronym is to avoid training. Therefore, we employ a frozen pre-trained 2D vision model, DINOv2~\cite{oquab2023dinov2} in our case, to extract features. However, these models are designed for image inputs and cannot operate directly on 3D data.
To overcome this limitation, we rasterize the input point cloud into a collection of images, capturing the input from various angles.
We rasterize $\mathcal{X}$ from $R$ pre-defined camera viewpoints sampled from a unit sphere around the object to obtain the set of images $\{\textbf{I}_r\}_{r=1}^R$.
If $\mathcal{X}$ contains only coordinate information, $\textbf{I}_r \in [0, 1]^{H \times W}$ are greyscale (non-metric) depth maps.
If $\mathcal{X}$ also contains photometric information associated with each point, $\textbf{I}_r \in [0, 1]^{H \times W \times 3}$ are RGB images.  

Unlike~\cite{zhang_pointclip_2021, zhu_pointclip_2023}, during rendering we do not subsample the point cloud. As a consequence, our model produces higher-quality renders that look more photorealistic and close the gap with the VFM's training data.
Additionally, we have chosen to retain the background of the point cloud when it is provided (\eg in ScanObjectNN~\cite{uy2019scanobjectnn}). This increases the task's difficulty, as it can add noise, either surrounding the point cloud or introducing other objects in the scene.

\noindent\textbf{2D feature extraction.}
We can now extract features from the images obtained in the previous step.
Let us denote with $\Gamma_\Theta$ the VFM with parameters $\Theta$.
We implement $\Gamma_\Theta$ with DINOv2~\cite{oquab2023dinov2}, keeping its parameters $\Theta$ frozen.
To perform part segmentation, we require dense features, but most VFMs, including DINOv2, are ViT-based~\cite{dosovitskiy2021image}, thus providing only patch-level features.
Therefore, we obtain pixel-level features for each image $\textbf{I}_r$ by interpolating the patch-level ones with a bicubic interpolator $\Pi$, resulting in $\Pi \circ \Gamma_\Theta (\textbf{I}_r) = \textbf{H}_r \in \mathbb{R}^{H \times W \times d}$, where $H \times W$ is the resolution of $\textbf{I}_r$ and $d$ is the feature size.

\noindent\textbf{Feature back-projection.}
As a last step, we need to bring the 2D features to the 3D points.
We achieve this via back-projection, which exploits the correspondence $\zeta \colon \textbf{I}_r \to \mathcal{X}$ between the pixels $\textbf{p}_{ij}$ of a rendered 2D image and the 3D points $\textbf{x}_k$ of $\mathcal{X}$.
The $\zeta$ is a byproduct of the rasterization procedure, requiring no additional computation.
Let $\textbf{x}_{i} \in \mathcal{X}$ be the $i^{th}$ point of the point cloud and  $\mathcal{Z}_i \in \zeta$ the set of pixel-to-point mappings for the $i^{th}$ point, $\textbf{z}_{ir}$ being the mapping for the $i^{th}$ point in the rendered image $\textbf{I}_r$.
The feature vector $\textbf{f}_{i} \in \mathcal{R}^d$ for $\textbf{x}_{i}$ is computed as:
\begin{equation}\label{eq:backprojection}
    \textbf{f}_{i} = \frac{1}{|\mathcal{Z}_i|} \sum_{\mathbf{z}_{ir} \in \mathcal{Z}_i} \textbf{H}_r[\mathbf{z}_{ir}],
\end{equation}
which gets the pixel coordinates of $\textbf{x}_j$ in each view and averages the features of such pixels.
This back-projection mechanism provides a linear combination of features, therefore being scalable and computationally cheaper than other approaches that filter foreground content (\eg \cite{abdelreheem2023satr, zhou2023partslip, liu_partslip_2023, kim2024partstad}). 
Lastly, some points might be invisible in all views. According to geometric consistency, \ie points that are close in space likely belong to the same part, \acronym estimates features for hidden points by averaging those of their $L = 20$ nearest neighbours in 3D space.

\subsection{Geometric feature aggregation (GFA)}
The features extracted with $\mathcal{F} = \Phi_{\mathcal{E}}$ are derived from 2D views, thus lacking spatial and semantic consistency that are keys to creating coherent and robust 3D representations.
In the following, we describe how we refine $\mathcal{F}$ with two consecutive steps, spatially- and semantically-consistent GFA.
In \cref{sec:ablation}, we refer to \cref{eq:backprojection} as ``baseline''.

\noindent\textbf{Spatially-consistent GFA}
performs geometric reasoning at the coordinate level, \ie $\mathcal{X}^P$.
It follows two main principles:
(i) model point cloud regions via centroids,
(ii) propagate features from super points to the original ones. 
We extract a set of $M$ centroids $\mathcal{C} = \{ \textbf{x}_m \in \mathcal{X}^P \}_{m=1}^M$ by farthest point sampling (FPS) on $\mathcal{X}^P$.
For each centroid $\textbf{c} \in \mathcal{C}$, we collect the features associated with its $K$ nearest neighbours $\mathcal{X}_c = \{\textbf{x}_j\}_{j=1}^K \in \mathcal{X}^P$, and update the centroid's features by computing their average:
\begin{equation}
    \textbf{f}_c = \frac{1}{K} \sum_{\textbf{x}_i\in\mathcal{X}_c} \textbf{f}_i.
\end{equation}
Once the centroid features are updated, we update the features of the original points, \ie $\mathcal{X} \setminus \mathcal{C}$, via interpolation:
\begin{equation}
\label{eq:spatial-aggregation}
    \textbf{f}^\text{ spatial}_i = \frac{1}{K'}\sum_{\textbf{c}\in\mathcal{C}_{K'}} \textbf{f}_c,
\end{equation}
where $\mathcal{C}_{K'}$ are the $K'$ closest centroids $\textbf{c}$ to point $i$. Therefore, the feature vector of each point is influenced by the closest centroids, obtaining spatially consistent features (\cref{eq:spatial-aggregation}) that encode the geometrical relationships of points.

\noindent\textbf{Semantically-consistent GFA} refines the spatially consistent features by performing geometric reasoning at the feature level, \ie $\mathcal{F}$.
Similarly to the spatially-consistent counterpart, we still extract $\mathcal{C}$ via FPS, but the $K$ nearest neighbours are computed in the feature space rather than in the coordinate space, \ie $\mathcal{F}_c = \{\textbf{f}_j\}_{j=1}^K \in \mathcal{F}$.
The features of the centroid $c$ are updated as:
\begin{equation}
    \textbf{f}^\text{ semantic}_c = \frac{1}{K} \sum_{\textbf{f}_i\in\mathcal{F}_c} \textbf{f}^\text{ spatial}_i,
\end{equation}
where the closest points are retrieved directly in the feature space. The final point-level features are obtained as:
\begin{equation}
\label{eq:semantic-aggregation}
    \hat{\textbf{f}}_i = \frac{1}{K'}\sum_{\textbf{f}_c\in\mathcal{C}_{K'}} \textbf{f}^\text{ semantic}_c,
\end{equation}
where $\mathcal{C}_{K'}$ are the $K'$ closest centroids $\textbf{c}$ to point $i$.
We term this aggregation mechanism semantically-consistent because neighbouring features correspond to points that have a similar semantic but that can be distant in the coordinate space (\eg the wheels of an office chair).
This module operates under the assumption that the VFM maps semantically similar points to close regions of the latent space.

\subsection{Zero-shot part segmentation head}
\vspace{-2mm}
This section discusses $\Psi \colon \mathcal{F} \mapsto \tilde{\mathcal{Y}} \in \{1, \dots, P\}^{N}$, which requires no specific training for our task.
$\Psi$ enables zero-shot 3D part segmentation by performing:
(i) \emph{shape-level clustering},
(ii) \emph{language-driven semantic anchors extraction}, and
(iii) \emph{label alignment} to classify parts.

\noindent\textbf{Shape-level clustering.}
We cluster the point-level features $\mathcal{F}$ to obtain candidate parts.
Given a target of $P$ parts, we apply K-means to split the point cloud into $P$ regions $\{\mathcal{X}_1, \cdots, \mathcal{X}_P\}$.
Clustering takes advantage of the spatially and semantically consistent features extracted via \cref{eq:semantic-aggregation}.

\noindent\textbf{Semantic anchors extraction.}
At this stage, the obtained parts lack a direct link to natural language, as there is no mapping from cluster index to categorical label.
We exploit the capabilities of VLMs to provide such a mapping, which is required to compare fairly with our competitors.
We note that this step is not strictly necessary, \eg, for applications that do not require this mapping but simply require parts.

We employ a VLM, CLIP in our case following~\cite{zhang_pointclip_2021, zhu_pointclip_2023}, to encode both the rendered views of the object and textual descriptions of the target parts.
These can be only the part name, manually annotated descriptions, or LLM-based ones: we analyse all these options in our ablation study.
Next, we compute the cosine similarity between the 2D visual features and the textual embeddings.
We back-project these scores to 3D exploiting the same function $\zeta$ defined in \cref{eq:backprojection}, and we classify the point with the label with the highest similarity.
Lastly, we group points by label, obtaining $P$ segmentation masks, denoted as $\{\hat{\mathcal{Y}}_1, \cdots, \hat{\mathcal{Y}}_P\}$.

\noindent\textbf{Final segmentation.}
The last step is to assign semantic labels to the clusters by mapping each cluster $\mathcal{X}_i$ to a segmentation mask $\hat{\mathcal{Y}}_j$.
To ensure each mask is assigned to only one cluster, we use the Hungarian algorithm to measure the overlap between the set of points for each pair $(\mathcal{X}_i, \hat{\mathcal{Y}}_j)$.

    \section{Experimental results}%
\label{sec:experiments}

\subsection{Benchmark details}%
\label{ss:dataset}
\vspace{-2mm}
\spacing\noindent\textbf{Datasets.}
We evaluate \acronym on five part segmentation datasets, including the synthetic datasets ShapeNetPart~\cite{yi2016shapenetpart}, PartNet~\cite{Mo_2019_CVPR}, and PartNetE~\cite{liu_partslip_2023}, and the real-world datasets ScanObjectNN~\cite{uy2019scanobjectnn} and FAUST~\cite{Bogo:CVPR:2014}.
ShapeNetPart~\cite{yi2016shapenetpart} contains 2,874 point clouds, spread across 16 categories. It provides no photometric information.
Next, PartNet~\cite{Mo_2019_CVPR} is a larger and more diverse dataset with 24 categories comprising 26,671 objects, of which 5,169 are in the test set. Points provide colour information.
Following PartNetE~\cite{liu_partslip_2023} is a subset of PartNet containing 1,906 test set point clouds with colour.
Fourth, ScanObjectNN~\cite{uy2019scanobjectnn} provides 15 categories with a total of 2,902 real-world textured point clouds (580 in the test set).
Point clouds are provided both without (\texttt{OBJ-ONLY}) and with (\texttt{OBJ-BG}) background, which can provide context but also introduce noise.
Finally, FAUST~\cite{Bogo:CVPR:2014} contains 300 untextured real human scans of 10 different subjects in various poses. We utilise the coarse-grained part annotations proposed in SATR~\cite{abdelreheem2023satr}.

\noindent\textbf{Metrics.} 
We measure performance as the mean Intersection over Union (mIoU) between our prediction and the ground truth, reporting both:
(i) class-average, which averages the mIoU of point clouds in the same category, and then averages class IoUs;
and (ii) instance-average IoU, which averages the mIoU of all objects, regardless of their class.
Following our competitors in \cref{tab:zs_partnete}, we use the average IoU (aIoU)~\cite{xue2023zerops}.
We also report results when measuring segmentation quality without assigning semantic labels to parts, indicating it with ``$\uparrow$''.
This shows the upper bound \acronym can achieve when a \emph{perfect} labeller is employed.
This setting may be useful when the downstream task does not require semantic labels, but only which parts compose the object.

\vspace{1pt}
\spacing\noindent\textbf{Baselines.}
We compare \acronym to other methods that leverage VLMs.
When there are no direct competitors, we use PointCLIPv2 as our baseline.
Because there are different implementations of evaluation metrics, we meticulously checked other methods' source code to ensure comparable results.

\vspace{1pt}
\spacing\noindent\textbf{Implementation details.}
We use DINOv2 base~\cite{oquab2023dinov2} as VFM and CLIP ViT-B/16~\cite{radford2021learning} for semantic labelling.
We render 48 RGBD images per point cloud, with a point size of 0.01 or 0.04 depending on the density of the dataset.

\subsection{Quantitative results}
\vspace{-2mm}

\begin{table*}[t!]
\centering
\tabcolsep 3pt

\resizebox{2\columnwidth}{!}{%
\begin{tabular}{crl|cc|cccccccccccccccc}
    \toprule
    \color{gray} \rotatebox{0}{} & & Method & \rotatebox{0}{$\textrm{mIoU}_\textrm{I}$} & \rotatebox{0}{$\textrm{mIoU}_\textrm{C}$} & \rotatebox{0}{Airplane} & \rotatebox{0}{Bag} & \rotatebox{0}{Cap} & \rotatebox{0}{Car} & \rotatebox{0}{Chair} & \rotatebox{0}{Earph.} & \rotatebox{0}{Guitar} & \rotatebox{0}{Knife} & \rotatebox{0}{Lamp} & \rotatebox{0}{Laptop} & \rotatebox{0}{Motor.} & \rotatebox{0}{Mug} & \rotatebox{0}{Pistol} & \rotatebox{0}{Rocket} & \rotatebox{0}{Skate} & \rotatebox{0}{Table} \\
    \toprule
    & & \color{gray} Instances & \multicolumn{2}{c|}{\color{gray} 2874} & \color{gray} 341 & \color{gray} 14 & \color{gray} 11 & \color{gray} 158 & \color{gray} 704 & \color{gray} 14 & \color{gray} 159 & \color{gray} 80 & \color{gray} 286 & \color{gray} 83 & \color{gray} 51 & \color{gray} 38 & \color{gray} 44 & \color{gray} 12 & \color{gray} 31 & \color{gray} 848 \\
    \color{gray} \multirow{2}{*}{\rotatebox{90}{mesh}} & \color{gray} 1 & 3DH \cite{decatur2022highlighter} & 9.6 & 5.7 & 5.8 & 2.1 & 2.9 & 2.9 & 15.5 & 9.6 & 0.9 & 1.6 & 13.2 & 1.8 & 5.6 & 0.7 & 1.4 & 10.4 & 6.4 & 10.8 \\
    & \color{gray} 2 & SATR \cite{abdelreheem2023satr} & 32.8 & 31.9 & 38.5 & 44.6 & 24.0 & 19.6 & 33.2 & 16.9 & 40.2 & 45.9 & 30.2 & 37.8 & 15.7 & 52.3 & 20.9 & 28.4 & 30.8 & 31.4 \\
    \midrule
    \color{gray} \multirow{3}{*}{\rotatebox{90}{pcd}} & \color{gray} 3 & PointCLIP \cite{zhang_pointclip_2021} & 24.8 & 29.8 & 22.0 & 44.8 & 13.4 & - & 18.7 & 28.3 & 22.7 & 24.8 & - & 22.9 & - & 48.6 & - & 22.7 & 42.7 & 45.5 \\
    & \color{gray} 4 & PointCLIPv2~\cite{zhu_pointclip_2023} & 51.8 & 51.0 & 33.4 & 60.4 & 52.9 & 27.2 & 51.6 & 56.5 & \textbf{71.5} & \textbf{76.7} & 44.7 & 61.5 & \textbf{31.5} & 48.0 & 46.1 & 49.6 & 43.9 & 61.1 \\
    \rowcolor{myazure} \color{white} & \color{gray} 5 & \acronym & \textbf{62.5} & \textbf{57.3} & \textbf{51.4} & \textbf{67.8} & \textbf{68.6} & \textbf{30.1} & \textbf{69.0} & \textbf{60.9} & 63.5 & 76.3 & \textbf{48.2} & \textbf{70.0} & 22.8 & \textbf{53.7} & \textbf{51.1} & \textbf{50.5} & \textbf{59.1} & \textbf{73.6} \\
    \midrule
    \rowcolor{mylightyellow} \color{white} & \color{gray} 6 & \acronym $\uparrow$ & 72.1 & 65.0 & 62.3 & 72.2 & 78.7 & 38.7 & 80.8 & 72.3 & 62.6 & 78.0 & 77.2 & 73.4 & 26.5 & 65.3 & 52.5 & 53.6 & 66.4 & 78.8 \\
    \midrule
    %
    & \color{gray} 7 & Improvement & \color{mygreen} +10.7 & \color{mygreen} +6.3 & \color{mygreen} +18.0 & \color{mygreen} +7.4 & \color{mygreen} +15.7 & \color{mygreen} +2.9 & \color{mygreen} +17.4 & \color{mygreen} +4.4 & \color{red} -8.0 & \color{red} -0.4 & \color{mygreen} +3.5 & \color{mygreen} +8.5 & \color{red} -8.7 & \color{mygreen} +5.7 & \color{mygreen} +5.0 & \color{mygreen} +0.9 & \color{mygreen} +15.2 & \color{mygreen} +12.5 \\
    \bottomrule
\end{tabular}
}
\vspace{-3mm}
\caption{
    Zero-shot part segmentation results on ShapeNetPart~\cite{yi2016shapenetpart}.
    In this Table and \cref{tab:zs_partnet}, \cref{tab:zs_partnete}, and \cref{tab:zs_sonn} we report results in terms of mean IoU across instances ($\textrm{mIoU}_\textrm{I}$) and categories ($\textrm{mIoU}_\textrm{C}$).
    Some categories for PointCLIP are missing since the authors of PointCLIPv2, who adapted it to part segmentation, do not report them.
    Row 7 shows our improvement over row 4, the best-performing competitor.
}%
\label{tab:zs_shapenetpart}

\end{table*}

\begin{table*}[t!]
\centering
\tabcolsep 3pt

\resizebox{2\columnwidth}{!}{%
\begin{tabular}{rl|cc|cccccccccccc}
    \toprule
    & Method & \rotatebox{0}{$\textrm{mIoU}_\textrm{I}$} & \rotatebox{0}{$\textrm{mIoU}_\textrm{C}$} & \rotatebox{0}{Bottle} & \rotatebox{0}{Bowl} & \rotatebox{0}{Clock} & \rotatebox{0}{Displ.} & \rotatebox{0}{Door} & \rotatebox{0}{Earph.} & \rotatebox{0}{Faucet} & \rotatebox{0}{Knife} & \rotatebox{0}{Lamp} & \rotatebox{0}{Micro.} & \rotatebox{0}{Table} & \rotatebox{0}{Vase} \\
    
    \toprule
    
    & \color{gray} Instances & \multicolumn{2}{c|}{\color{gray} 5169} & \color{gray} 1668 & \color{gray} 39 & \color{gray} 98 & \color{gray} 191 & \color{gray} 51 & \color{gray} 53 & \color{gray} 132 & \color{gray} 77 & \color{gray} 419 & \color{gray} 39 & \color{gray} 1668 & \color{gray} 233 \\
    \color{gray} 1 & PointCLIPv2 \cite{zhu_pointclip_2023} & 34.0 & 35.1 & 39.0 & 25.6 & 17.8 & 47.3 & 43.7 & 24.3 & 21.1 & 34.1 & 54.3 & \textbf{57.5} & 46.0 & 30.8 \\
    
    \rowcolor{myazure} \color{gray} 2 & \acronym & \textbf{38.8} & \textbf{38.9} & \textbf{44.6} & \textbf{29.2} & \textbf{22.0} & \textbf{54.7} & \textbf{49.8} & \textbf{31.8} & \textbf{22.6} & \textbf{35.3} & \textbf{55.0} & 55.1 & \textbf{46.3} & \textbf{36.8} \\

    \midrule

    \rowcolor{mylightyellow} \color{gray} 3 & \acronym $\uparrow$ & 85.6 & 79.1 & 79.8 & 89.6 & 91.4 & 85.3 & 81.6 & 90.6 & 82.7 & 93.2 & 95.9 & 80.6 & 94.5 & 92.1 \\
    
    \midrule 
    
    \color{gray} 4 & Improvement & \color{mygreen} +4.8 & \color{mygreen} +3.8 & \color{mygreen} +5.6 & \color{mygreen} +3.6 & \color{mygreen} +4.2 & \color{mygreen} +7.4 & \color{mygreen} +6.1 & \color{mygreen} +7.5 & \color{mygreen} +1.5 & \color{mygreen} +1.2 & \color{mygreen} +0.7 & \color{red} -2.4 & \color{mygreen} +0.3 & \color{mygreen} +6.0 \\
    \bottomrule
\end{tabular}
}

\vspace{-3mm}
\caption{
    Zero-shot part segmentation results on PartNet~\cite{Mo_2019_CVPR}.
    We show details about 12 out of the 24 total categories.
}%
\label{tab:zs_partnet}

\end{table*}

\begin{table*}[t!]
\centering
\tabcolsep 3pt

\resizebox{2\columnwidth}{!}{%
\begin{tabular}{rl|cc|cccccccccccccc}
    \toprule
    & Method & \rotatebox{0}{$\textrm{aIoU}_\textrm{I}$} & \rotatebox{0}{$\textrm{aIoU}_\textrm{C}$} & \rotatebox{0}{Clock} & \rotatebox{0}{Door} & \rotatebox{0}{Knife} & \rotatebox{0}{Table} & \rotatebox{0}{Box} & \rotatebox{0}{Lighter} & \rotatebox{0}{Oven} & \rotatebox{0}{Pen} & \rotatebox{0}{Safe} & \rotatebox{0}{Stapler} & \rotatebox{0}{Suitcase} & \rotatebox{0}{Toaster} \\
    
    \toprule
    
    & \color{gray} Instances & \multicolumn{2}{c|}{\color{gray} 1906} & \color{gray} 23 & \color{gray} 28 & \color{gray} 36 & \color{gray} 93 & \color{gray} 20 & \color{gray} 20 & \color{gray} 22 & \color{gray} 40 & \color{gray} 22 & \color{gray} 15 & \color{gray} 16 & \color{gray} 17 \\
    \color{gray} 1 & PartSLIP~\cite{zhou2023partslip} & - & 36.4 & 17.8 & 27.9 & 18.3 & 46.9 & 38.8 & 53.2 & 22.8 & 44.6 & 17.4 & 27.3 & 65.8 & 19.7 \\
    \color{gray} 2 & ZeroPS~\cite{xue2023zerops} & - & 56.0 & 33.8 & 37.8 & 68.7 & 53.3 & 63.1 & \textbf{64.4} & 37.0 & 71.7 & 26.2 & \textbf{80.7} & 62.2 & \textbf{62.7} \\
    \rowcolor{myazure} \color{gray} 3 & \acronym & \textbf{55.8} & \textbf{56.2} & \textbf{41.8} & \textbf{46.0} & \textbf{69.4} & \textbf{63.5} & \textbf{87.3} & 60.0 & \textbf{53.4} & \textbf{73.9} & \textbf{50.6} & 58.1 & \textbf{65.2} & 34.5 \\
    
    \midrule
    
    \color{gray} 4 & Improvement & \color{mygreen} - & \color{mygreen} +0.2 & \color{mygreen} +8.0 & \color{mygreen} +8.2 & \color{mygreen} +0.7 & \color{mygreen} +10.2 & \color{mygreen} +24.2 & \color{red} -4.4 & \color{mygreen} +16.4 & \color{mygreen} +2.2 & \color{mygreen} +24.4 & \color{red} -22.6 & \color{mygreen} +3.0 & \color{red} -27.7 \\
    \bottomrule
\end{tabular}
}

\vspace{-3mm}
\caption{
    Zero-shot part segmentation results on PartNetE~\cite{liu_partslip_2023}.
    We report 12 out of 45 categories.
    Following our competitors, we measure performance with the Average IoU (aIoU).
    Row 4 shows our improvement over row 2, the best-performing competitor.
}%
\label{tab:zs_partnete}

\end{table*}
\begin{table*}[t!]
\centering
\tabcolsep 3pt

\resizebox{2\columnwidth}{!}{%
\begin{tabular}{cl|cc|ccccccccccccccc}
    \toprule
    & Method & \rotatebox{0}{$\textrm{mIoU}_\textrm{I}$} & \rotatebox{0}{$\textrm{mIoU}_\textrm{C}$} & \rotatebox{0}{Bag} & \rotatebox{0}{Bed} & \rotatebox{0}{Bin} & \rotatebox{0}{Box} & \rotatebox{0}{Cabin.} & \rotatebox{0}{Chair} & \rotatebox{0}{Desk} & \rotatebox{0}{Displ.} & \rotatebox{0}{Door} & \rotatebox{0}{Pillow} & \rotatebox{0}{Shelf} & \rotatebox{0}{Sink} & \rotatebox{0}{Sofa} & \rotatebox{0}{Table} & \rotatebox{0}{Toilet} \\

    \toprule
    
    & \color{gray} Instances & \multicolumn{2}{c|}{\color{gray} 583} & \color{gray} 14 & \color{gray} 45 & \color{gray} 20 & \color{gray} 68 & \color{gray} 109 & \color{gray} 34 & \color{gray} 27 & \color{gray} 41 & \color{gray} 57 & \color{gray} 44 & \color{gray} 26 & \color{gray} 40 & \color{gray} 20 & \color{gray} 17 & \color{gray} 21 \\
    
    \color{gray} 1 & PointCLIPv2 \cite{zhu_pointclip_2023} & 10.6 & 11.6 & 12.7 & 5.1 & 3.5 & 8.3 & 9.5 & 9.1 & 21.4 & 19.6 & 5.4 & 11.8 & 17.6 & 30.0 & 8.4 & 6.5 & 5.0 \\
    
    \rowcolor{myazure} \color{gray} 2 & \acronym  & \textbf{20.2} & \textbf{20.3} & \textbf{19.1} & \textbf{9.3} & \textbf{10.1} & \textbf{14.0} & \textbf{25.2} & \textbf{17.5} & \textbf{31.4} & \textbf{24.8} & \textbf{14.1} & \textbf{23.3} & \textbf{31.7} & \textbf{39.3} & \textbf{13.6} & \textbf{20.7} & \textbf{10.8} \\

    \midrule

    \rowcolor{mylightyellow} \color{gray} 3 & \acronym $\uparrow$ & 63.0 & 62.2 & 69.9 & 74.7 & 81.8 & 75.7 & 65.3 & 46.9 & 52.3 & 54.1 & 65.2 & 59.6 & 50.0 & 59.2 & 57.4 & 51.5 & 69.5 \\
    
    \midrule

    \color{gray} 4 & Improvement & \color{mygreen} +9.6 & \color{mygreen} +8.7 & \color{mygreen} +6.4 & \color{mygreen} +4.2 & \color{mygreen} +6.6 & \color{mygreen} +5.7 & \color{mygreen} +15.7 & \color{mygreen} +8.4 & \color{mygreen} +10.0 & \color{mygreen} +5.2 & \color{mygreen} +8.7 & \color{mygreen} +11.5 & \color{mygreen} +14.1 & \color{mygreen} +9.3 & \color{mygreen} +5.2 & \color{mygreen} +14.2 & \color{mygreen} +5.8 \\

    \bottomrule
\end{tabular}
}

\vspace{-3mm}
\caption{
    Zero-shot part segmentation results on the OBJ-BG variant of  ScanObjectNN~\cite{uy2019scanobjectnn}.
    All 15 categories are reported.
}%
\label{tab:zs_sonn}

\end{table*}

\cref{tab:zs_shapenetpart} shows results on ShapeNetPart.
\acronym outperforms PointCLIPv2, the previous state-of-the-art model, by a significant margin of +10.7\% in instance average mIoU and +6.3\% in class average.
We show results on all 16 categories, observing a consistent increase over 13 categories, scoring +17.4\% on Chair, +15.7\% on Cap, and +18.0\% on Airplane.
\cref{tab:zs_partnet} reports scores on the PartNet dataset. Since no other model is competing in our setting, \ie zero-shot part segmentation, on this dataset, we provide a baseline by running PointCLIPv2.
Results are obtained by using GPT-generated prompts, similar to~\cite{zhu_pointclip_2023} but excluding the post-search optimisation phase.
We show detailed results for 12 out of 24 categories.
\acronym outperforms the baseline on 21 categories, achieving a considerable margin of +6.1\% on Door, +7.4\% on Display, and +7.5\% on Earphone.
Overall, our model surpasses PointCLIPv2 by +4.8\% in instance-average and +3.8\% in class-average.
We also compare with PartNetE~\cite{liu_partslip_2023} in \cref{tab:zs_partnete}, reporting 12 out of 45 categories.
\acronym surpasses both PartSLIP~\cite{liu_partslip_2023} and ZeroPS~\cite{xue2023zerops} by +0.2\%, with improvements of up to +24.2\% on Box and +24.4\% on Safe.
\cref{tab:zs_sonn} presents a detailed comparison on the challenging \texttt{OBJ-BG} setting of the real-world ScanObjectNN dataset.
\acronym outperforms across all categories, with gains as high as +15.7\% on Cabinet. The model effectively reduces the labeller's noise, achieving improvements of +9.6\% in instance-average mIoU and +8.7\% in class-average mIoU.
\cref{tab:zs_faust} shows performance on FAUST~\cite{Bogo:CVPR:2014}, with detailed results on the four coarse-grained parts annotated by SATR~\cite{abdelreheem2023satr}.
\acronym surpasses PointCLIPv2 by +16.8\%.
We do not compare with 3DH~\cite{decatur2022highlighter} and SATR~\cite{abdelreheem2023satr}, as they operate with triangular meshes that capture finer details compared to point clouds, making a direct comparison unfair.

\begin{table}[htpb!]
\centering
\tabcolsep 3pt

\small{
\begin{tabular}{crl|c|cccc}
    \toprule
    
    \color{gray} Input & & Method & mIoU & Arm & Head & Leg & Torso \\
    
    \toprule
    
    \color{gray} \multirow{2}{*}{\rotatebox{0}{mesh}} & \color{gray} 1 & 3DH \cite{decatur2022highlighter} & 16.5 & 28.6  & 14.2 & 14.9 & 8.2 \\
    & \color{gray} 2 & SATR \cite{abdelreheem2023satr} & 82.5 & 85.9 & 90.6 & 85.8 & 67.6 \\
    
    \midrule
    
    \color{gray} \multirow{3}{*}{\rotatebox{0}{pcd}} & \color{gray} 3 & PointCLIPv2 \cite{zhu_pointclip_2023} & 13.6 & 14.0 & 15.8 & 24.1 & 0.3 \\
    \rowcolor{myazure} \color{white}& \color{gray} 4 & \acronym & \textbf{30.4} & \textbf{29.8} & \textbf{33.4} & \textbf{48.2} & \textbf{10.2} \\
    \rowcolor{mylightyellow} \color{white} & \color{gray} 5 & \acronym $\uparrow$ & 64.7 & 72.4 & 79.7 & 62.2 & 42.3 \\
    
    \midrule

    & \color{gray} 6 & Improvement & \color{mygreen} +16.8 & \color{mygreen} +15.8 & \color{mygreen} +17.6 & \color{mygreen} +24.1 & \color{mygreen} +9.9 \\
    \bottomrule
\end{tabular}
}
\vspace{-3mm}
\caption{
    Zero-shot part segmentation results on the coarse-grained variant of FAUST~\cite{abdelreheem2023satr} in terms of mean IoU (mIoU).
    We show details about all the four parts, and directly compare with the only competitor that uses point clouds data.
}%
\label{tab:zs_faust}

\end{table}

\subsection{Qualitative results}
\begin{figure*}[!htb]
\centering

%
%
\begin{minipage}{0.12\textwidth}
    \begin{overpic}[width=\textwidth,trim=610 185 635 440,clip]{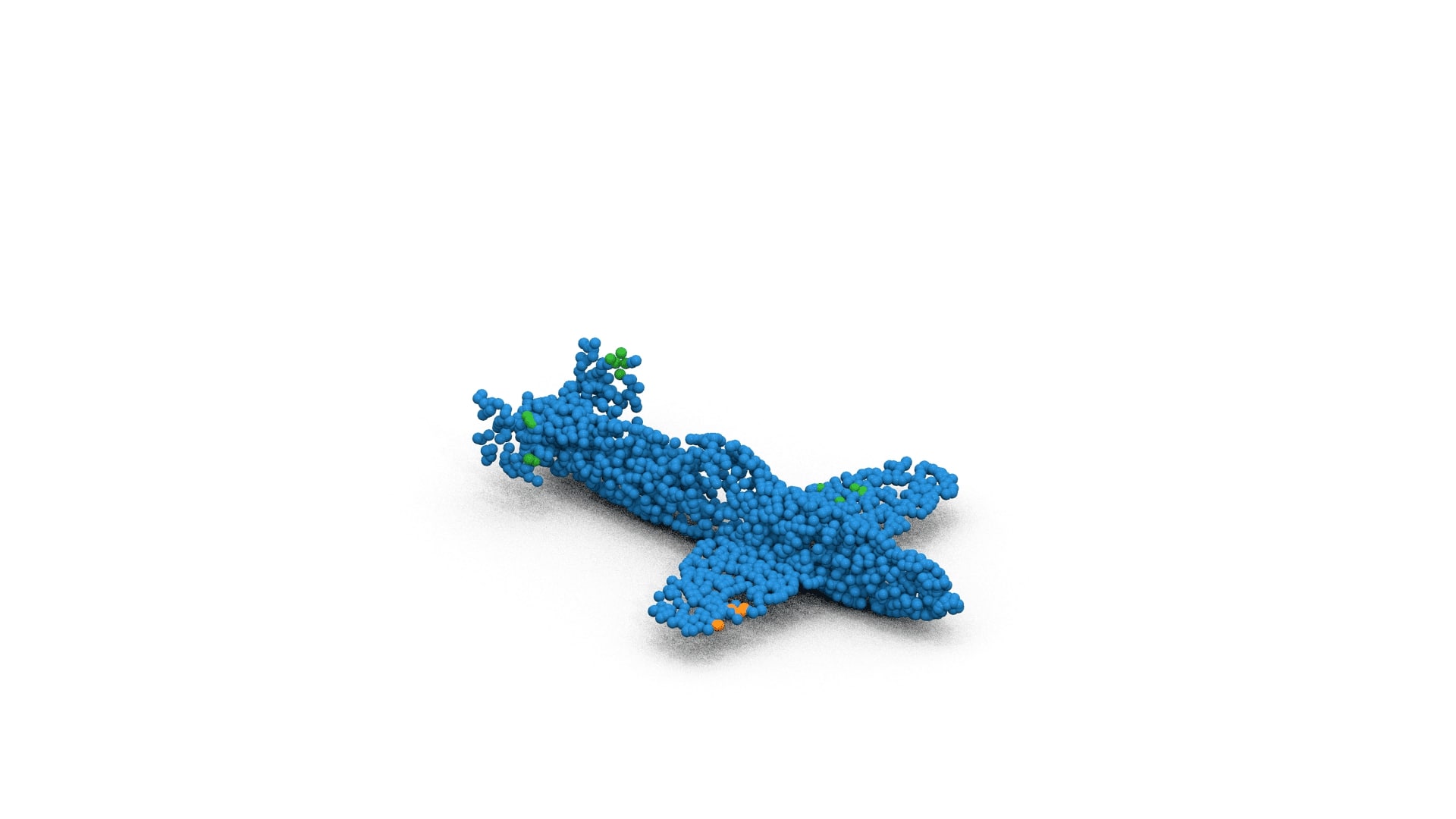}
        \put(-15,-10){\rotatebox{90}{\scriptsize PointCLIPv2}}
    \end{overpic}
\end{minipage}
\begin{minipage}{0.12\textwidth}
    \begin{overpic}[width=\textwidth,trim=680 220 645 465,clip]{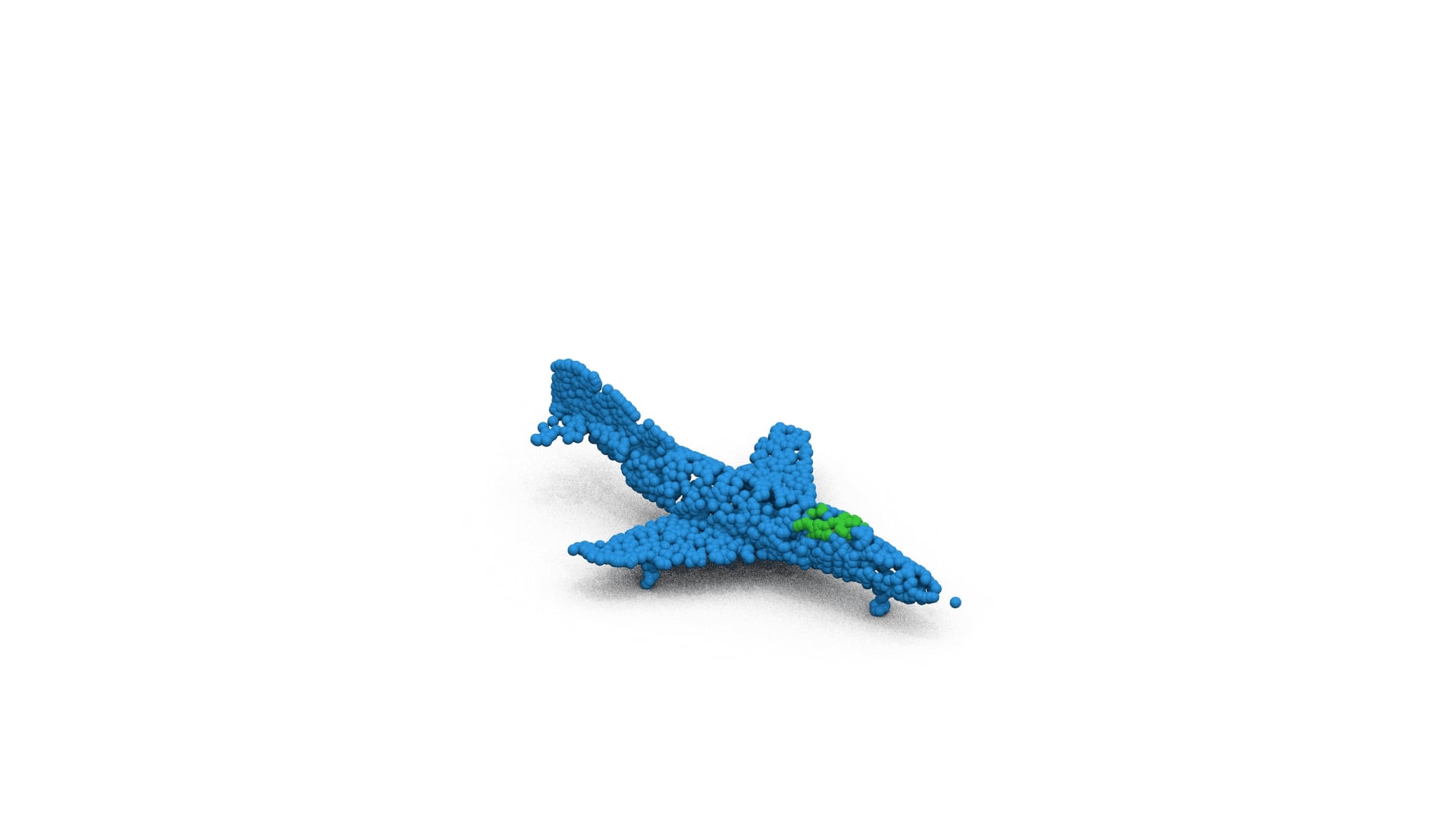}
    \end{overpic}
\end{minipage}
\begin{minipage}{0.12\textwidth}
    \begin{overpic}[width=\textwidth,trim=460 100 470 340,clip]{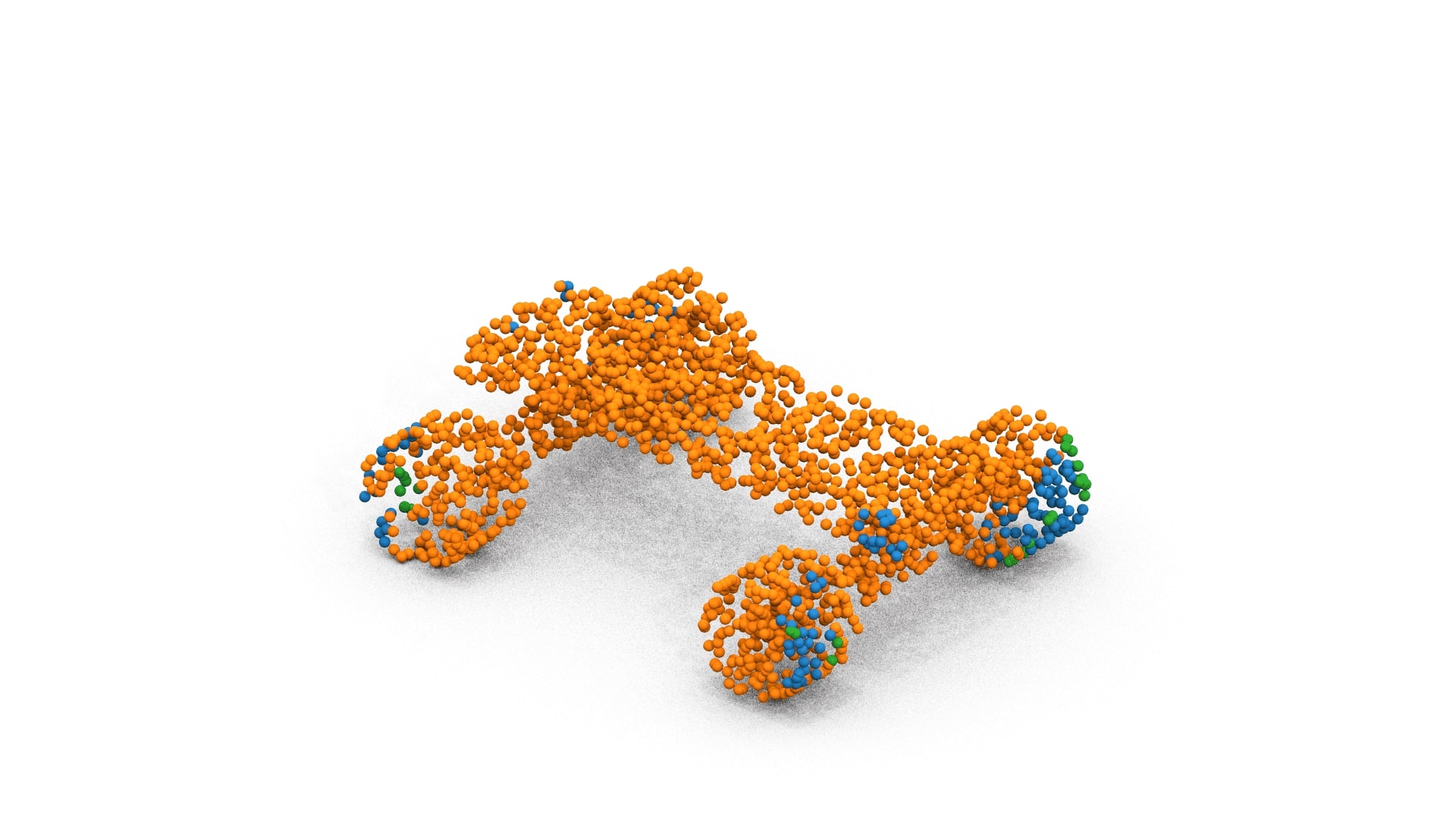}
    \end{overpic}
\end{minipage}
\begin{minipage}{0.12\textwidth}
    \begin{overpic}[width=\textwidth,trim=540 135 545 205,clip]{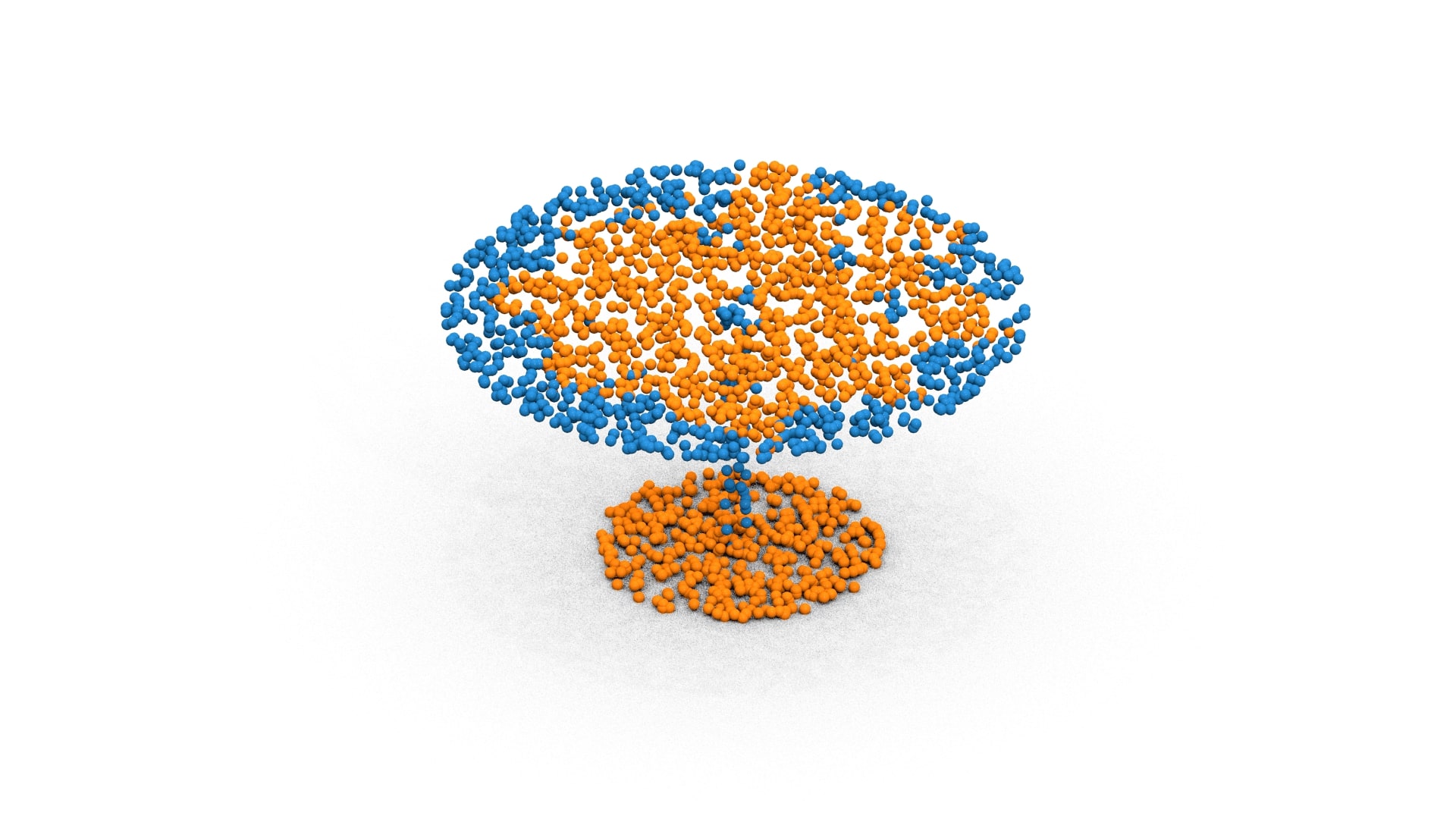}
    \end{overpic}
\end{minipage}
\begin{minipage}{0.12\textwidth}
    \begin{overpic}[width=\textwidth,trim=650 175 590 450,clip]{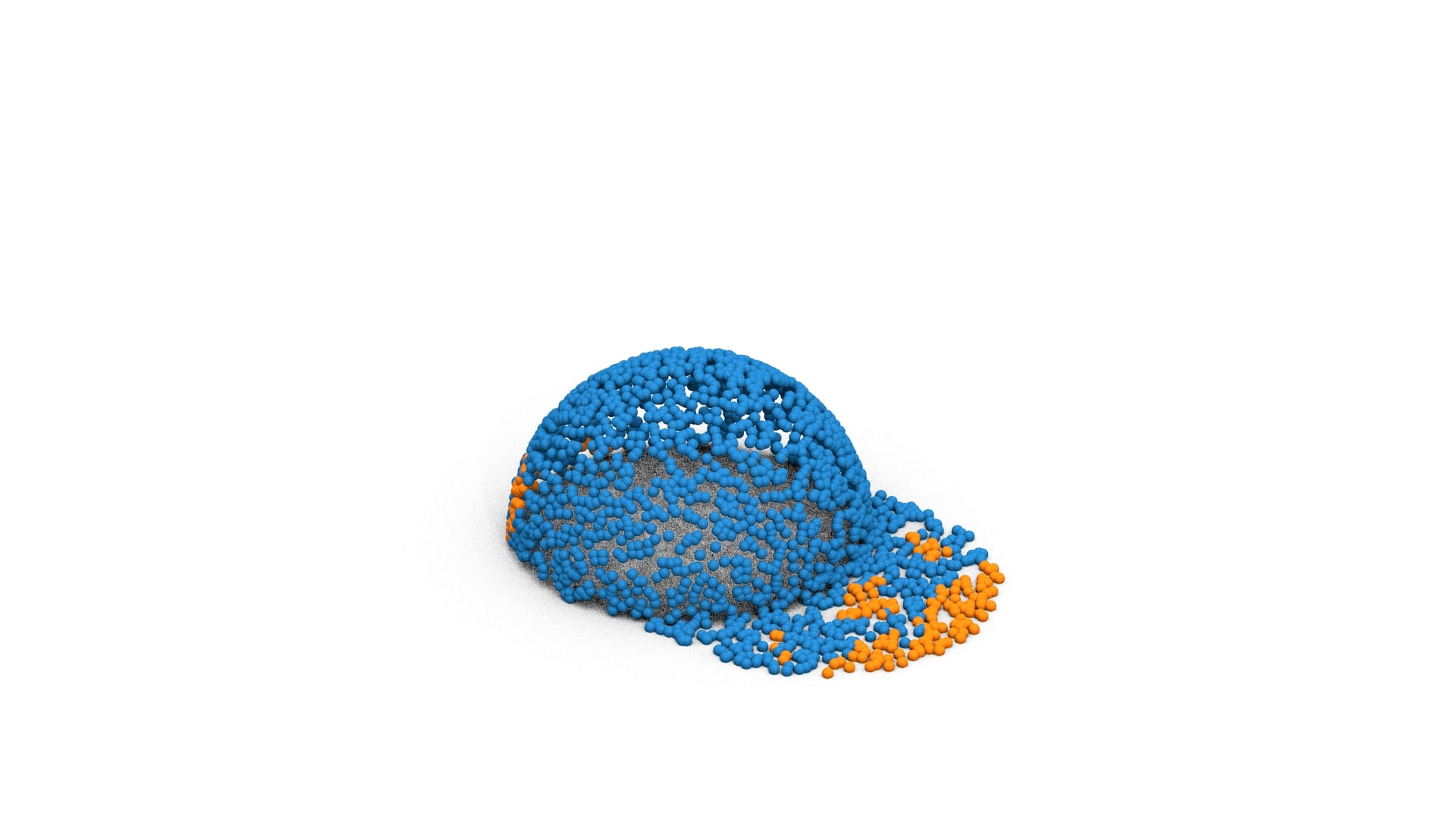}
    \end{overpic}
\end{minipage}
\begin{minipage}{0.12\textwidth}
    \begin{overpic}[width=\textwidth,trim=680 185 590 365,clip]{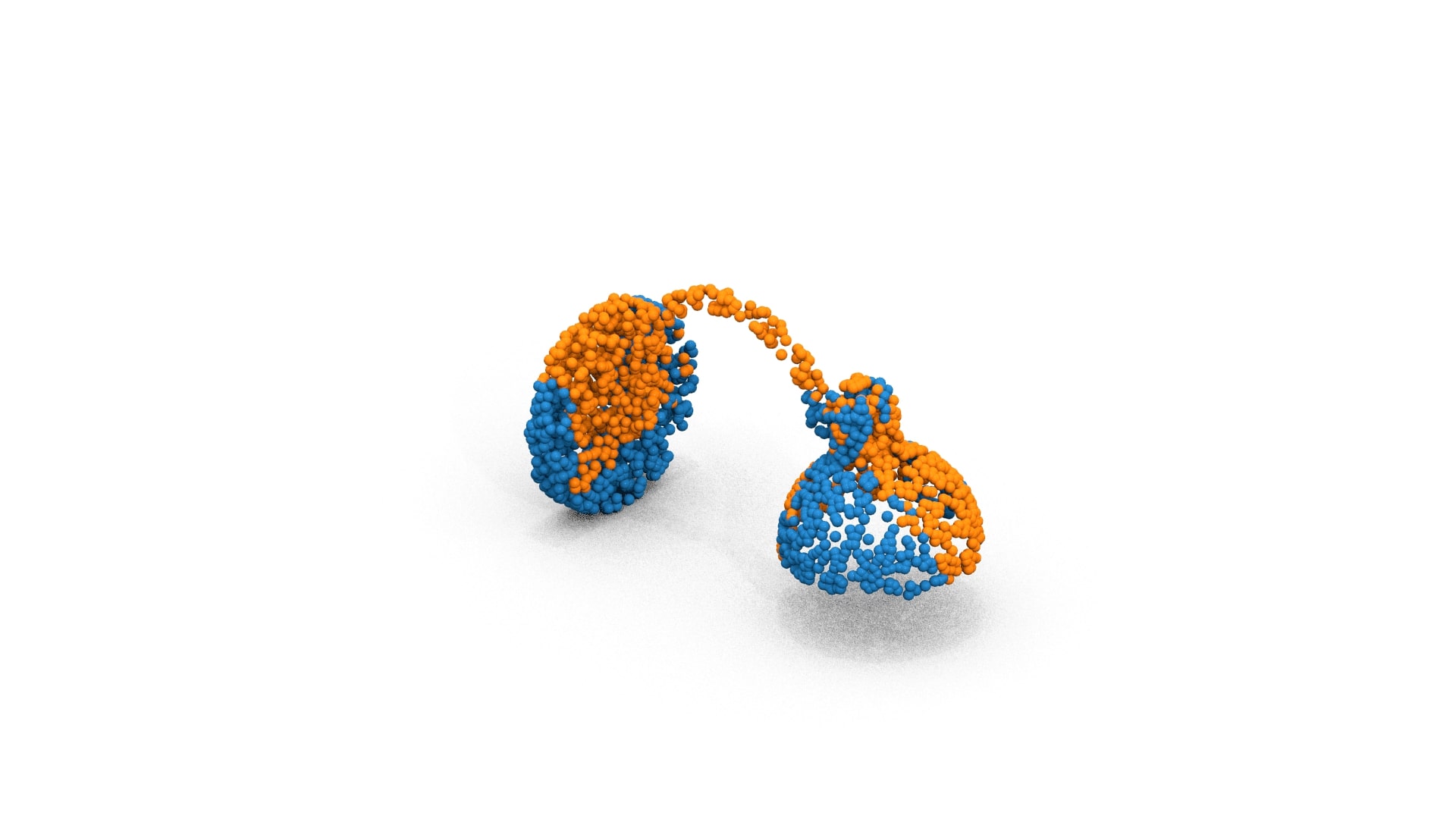}
    \end{overpic}
\end{minipage}
\begin{minipage}{0.12\textwidth}
    \begin{overpic}[width=\textwidth,trim=600 120 485 200,clip]{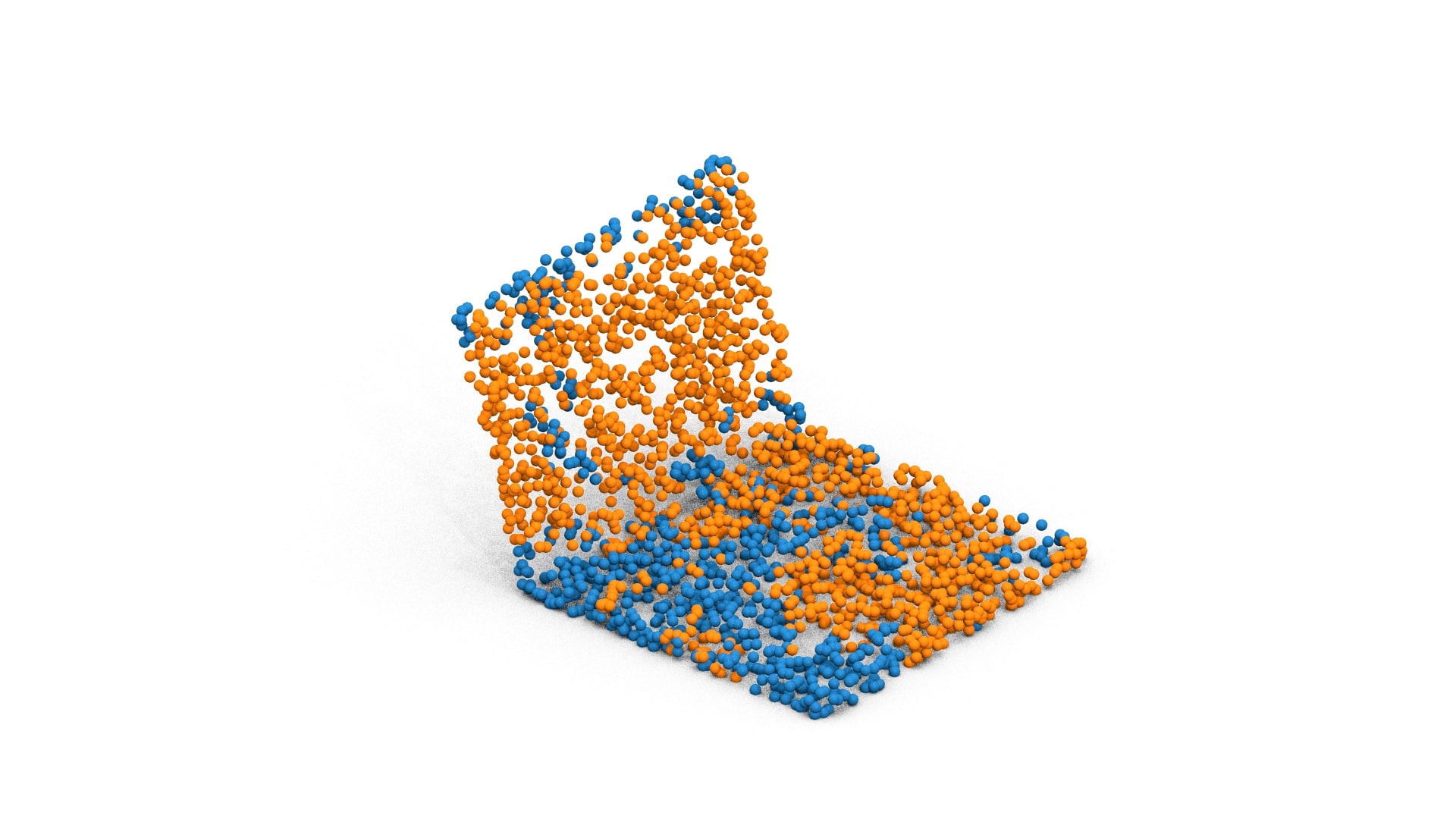}
    \end{overpic}
\end{minipage}\\
%
%
\begin{minipage}{0.12\textwidth}
    \begin{overpic}[width=\textwidth,trim=610 185 635 440,clip]{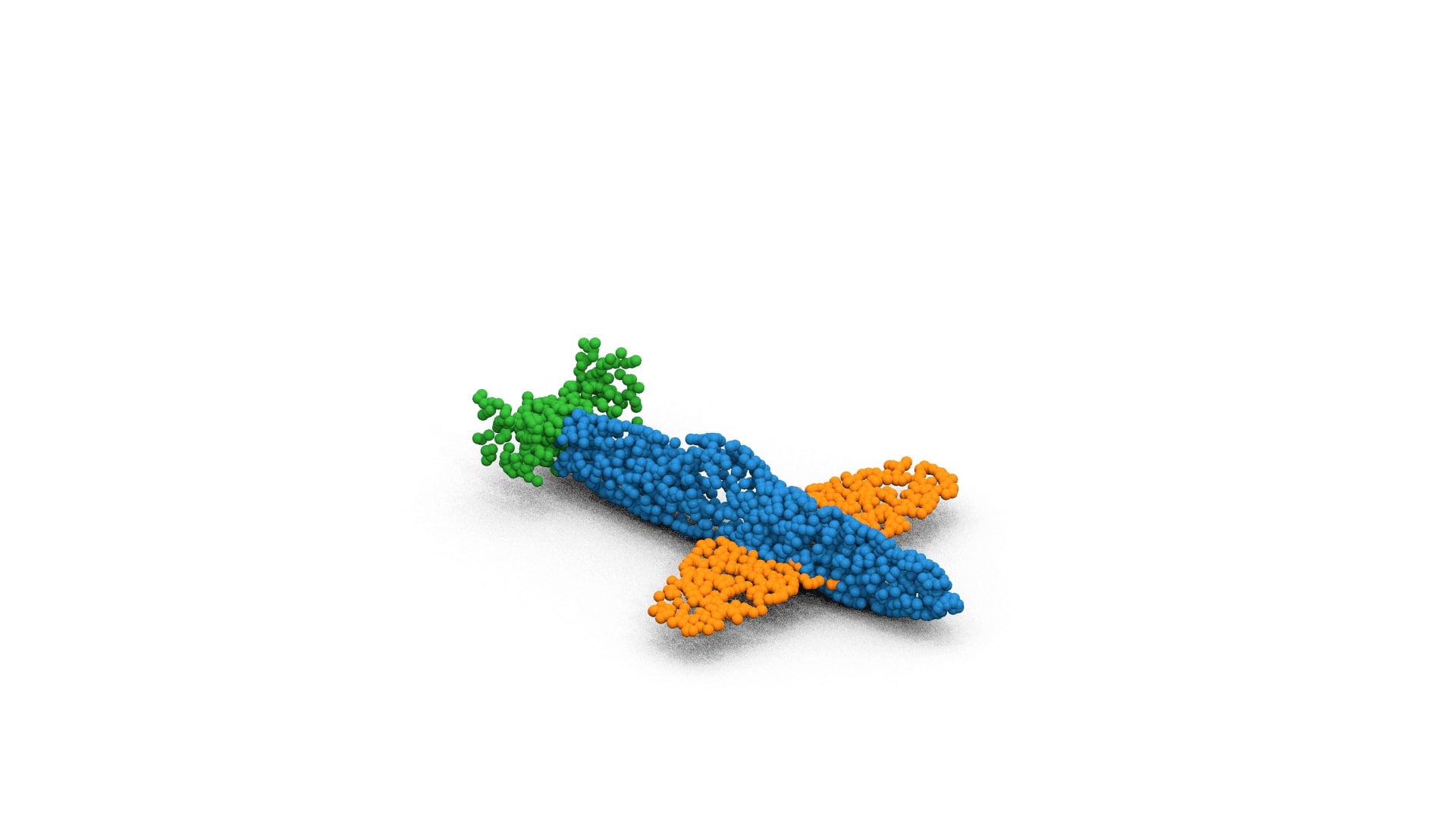}
        \put(-15,10){\rotatebox{90}{\scriptsize \acronym}}
    \end{overpic}
\end{minipage}
\begin{minipage}{0.12\textwidth}
    \begin{overpic}[width=\textwidth,trim=680 220 645 465,clip]{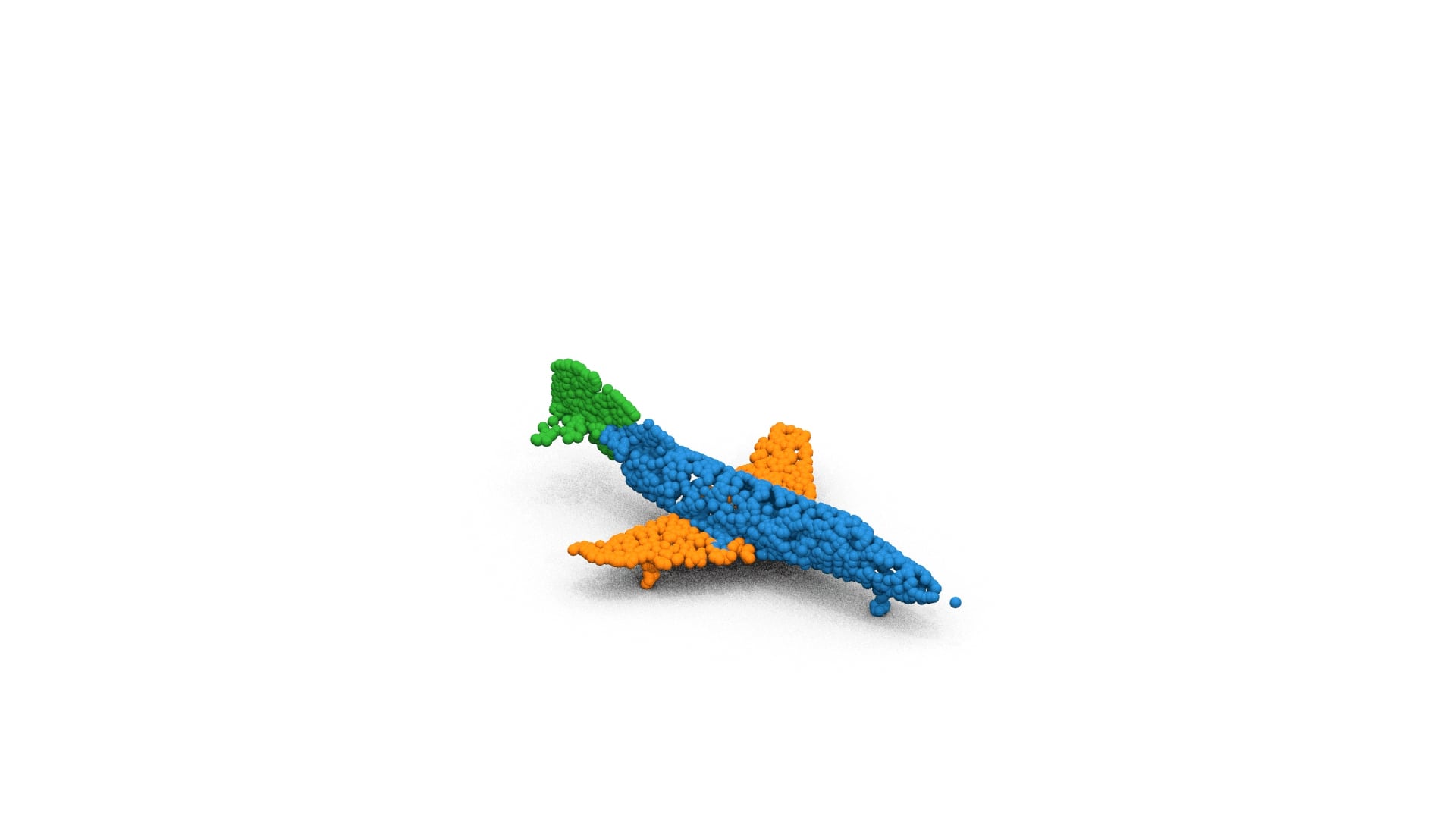}
    \end{overpic}
\end{minipage}
\begin{minipage}{0.12\textwidth}
    \begin{overpic}[width=\textwidth,trim=460 100 470 340,clip]{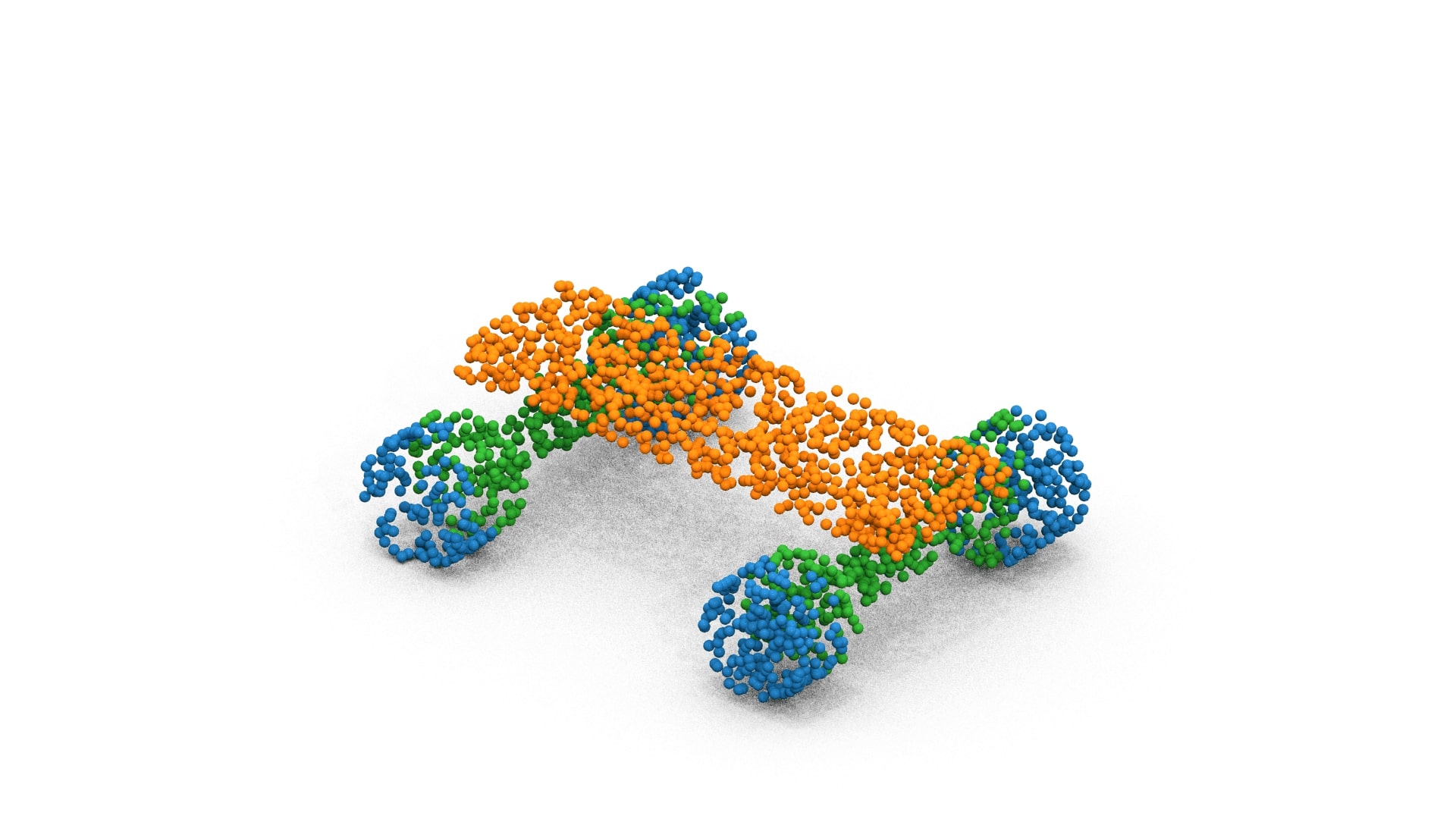}
    \end{overpic}
\end{minipage}
\begin{minipage}{0.12\textwidth}
    \begin{overpic}[width=\textwidth,trim=540 135 545 205,clip]{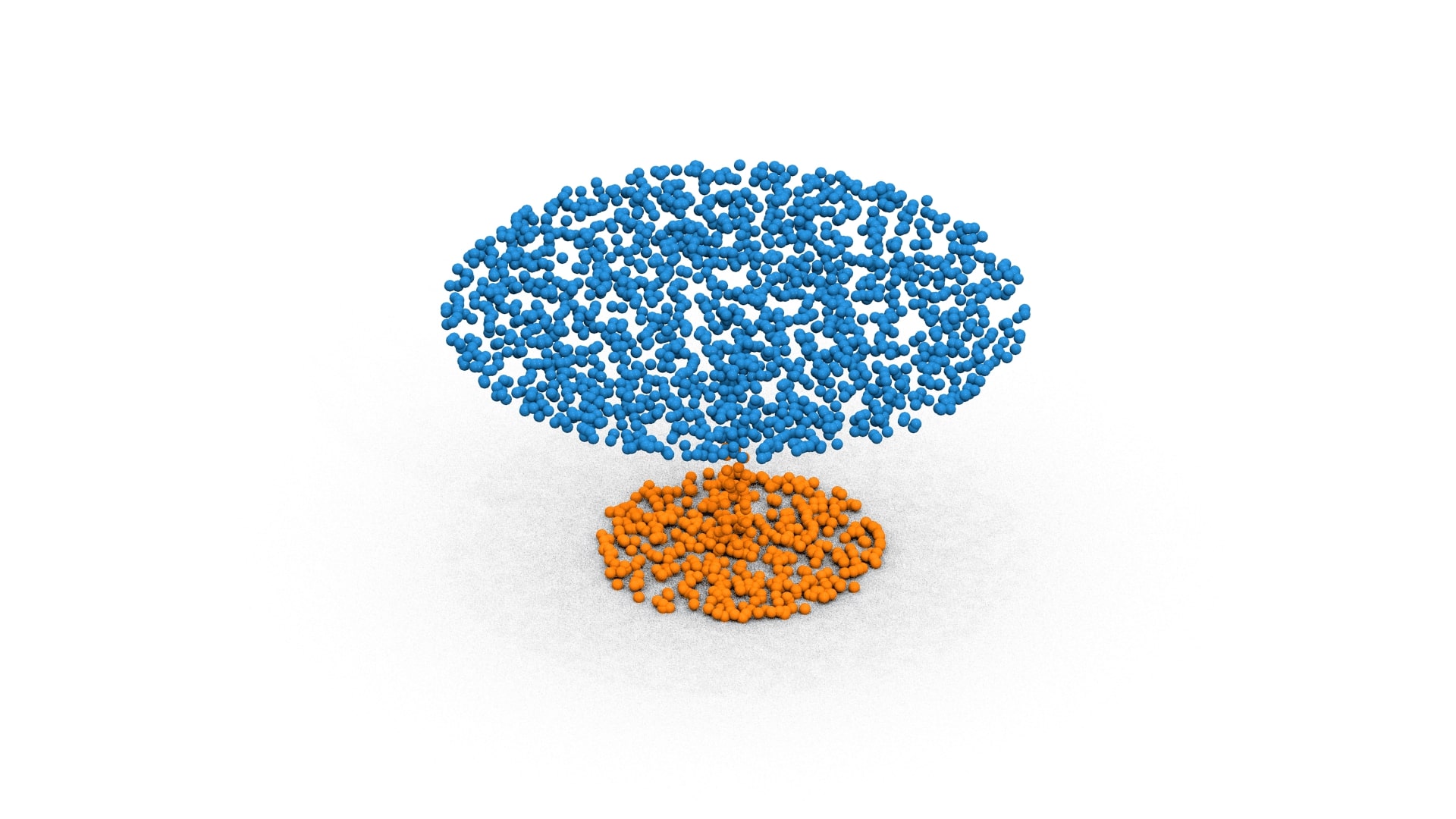}
    \end{overpic}
\end{minipage}
\begin{minipage}{0.12\textwidth}
    \begin{overpic}[width=\textwidth,trim=650 175 590 450,clip]{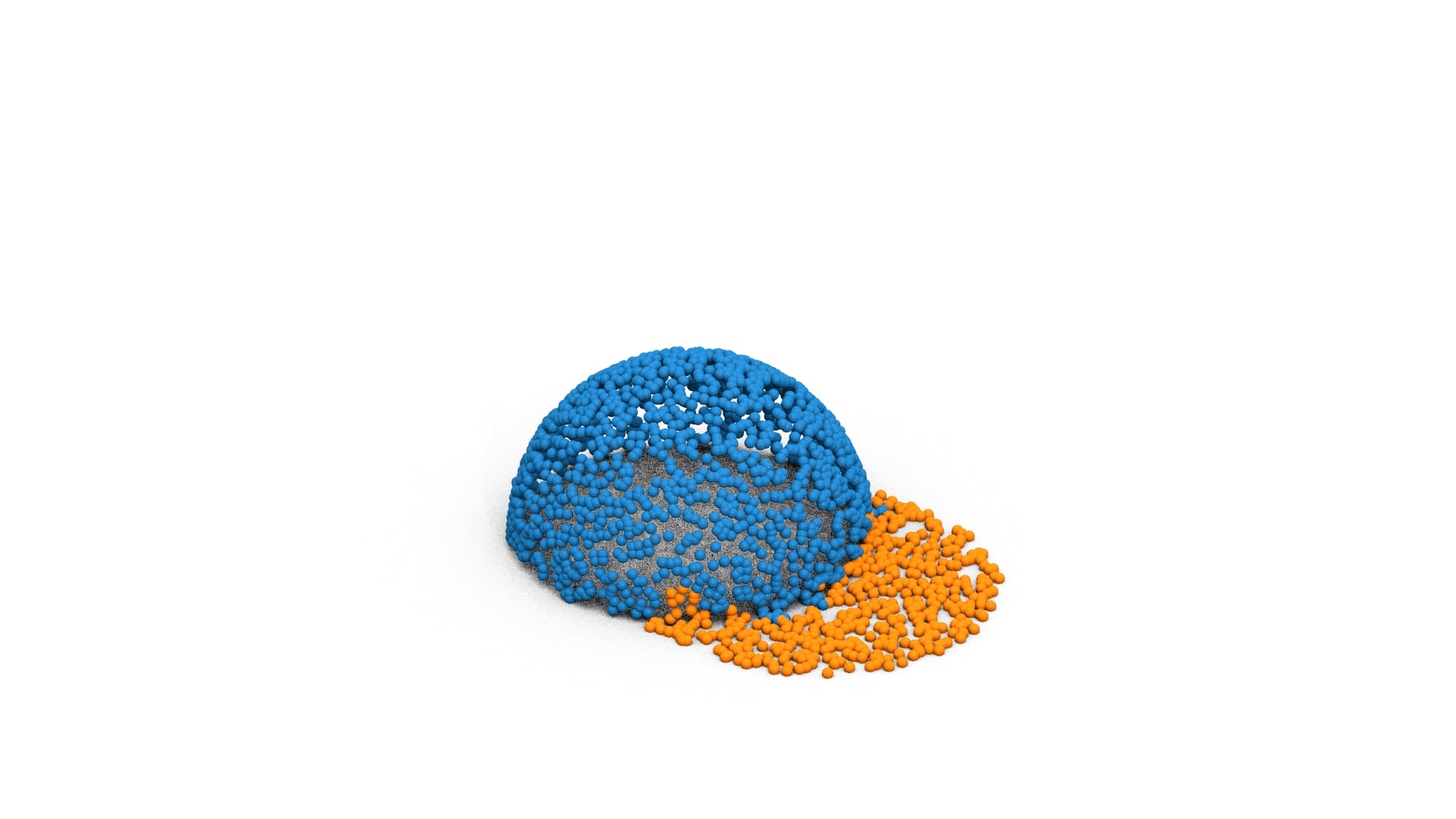}
    \end{overpic}
\end{minipage}
\begin{minipage}{0.12\textwidth}
    \begin{overpic}[width=\textwidth,trim=680 185 590 365,clip]{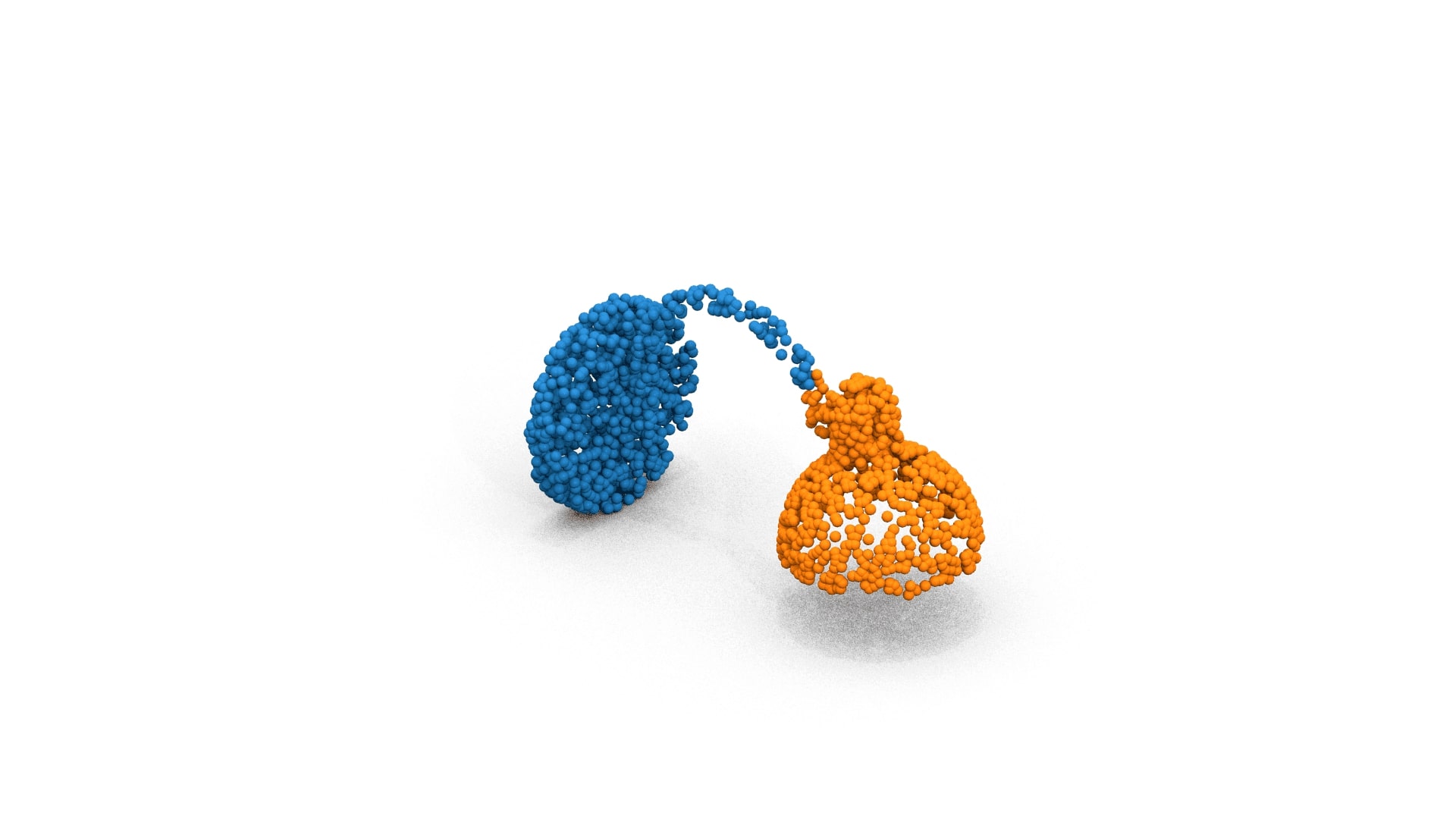}
    \end{overpic}
\end{minipage}
\begin{minipage}{0.12\textwidth}
    \begin{overpic}[width=\textwidth,trim=600 120 485 200,clip]{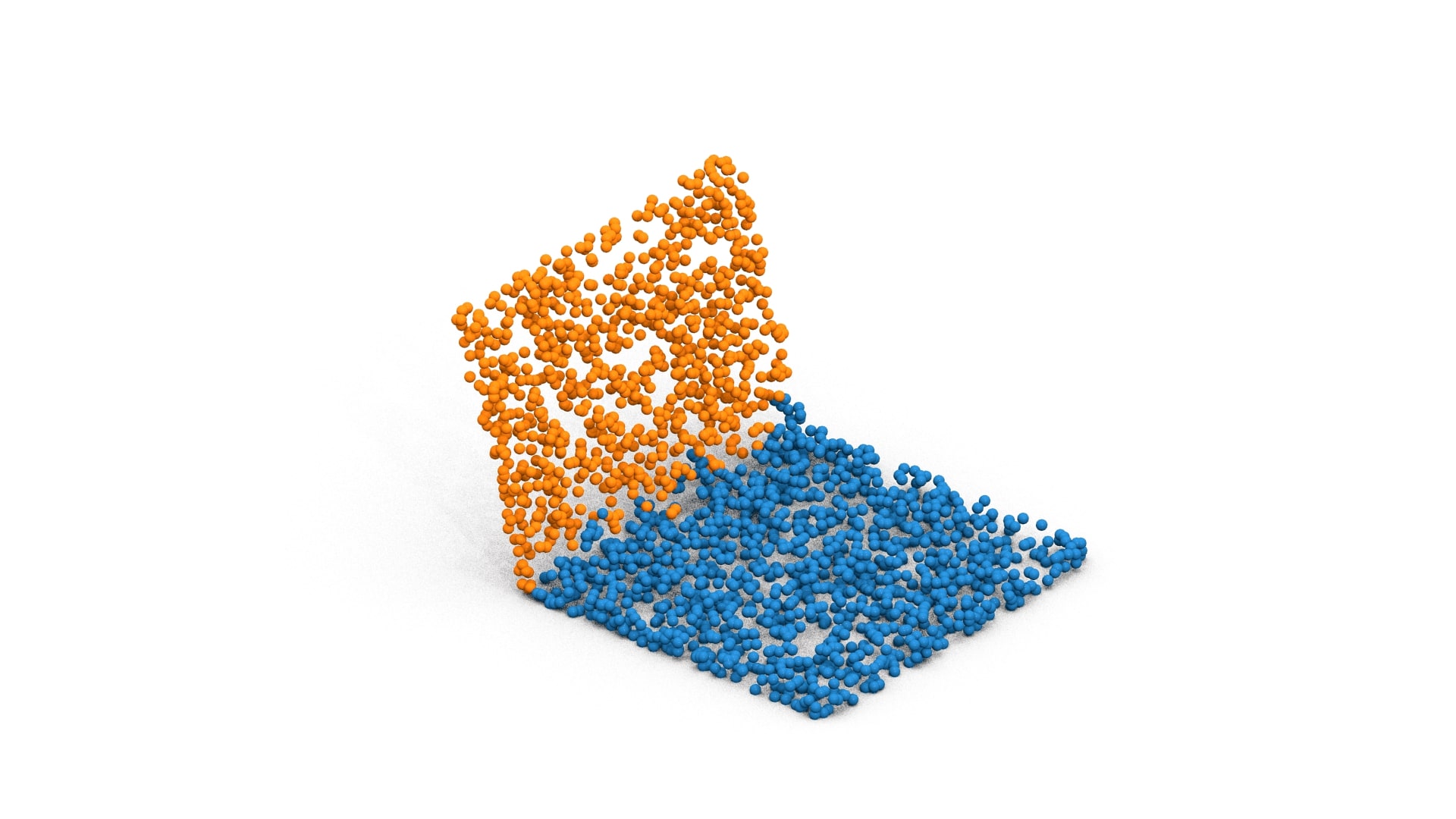}
    \end{overpic}
\end{minipage}\\
%
%
\begin{minipage}{0.12\textwidth}
    \begin{overpic}[width=\textwidth,trim=610 185 635 440,clip]{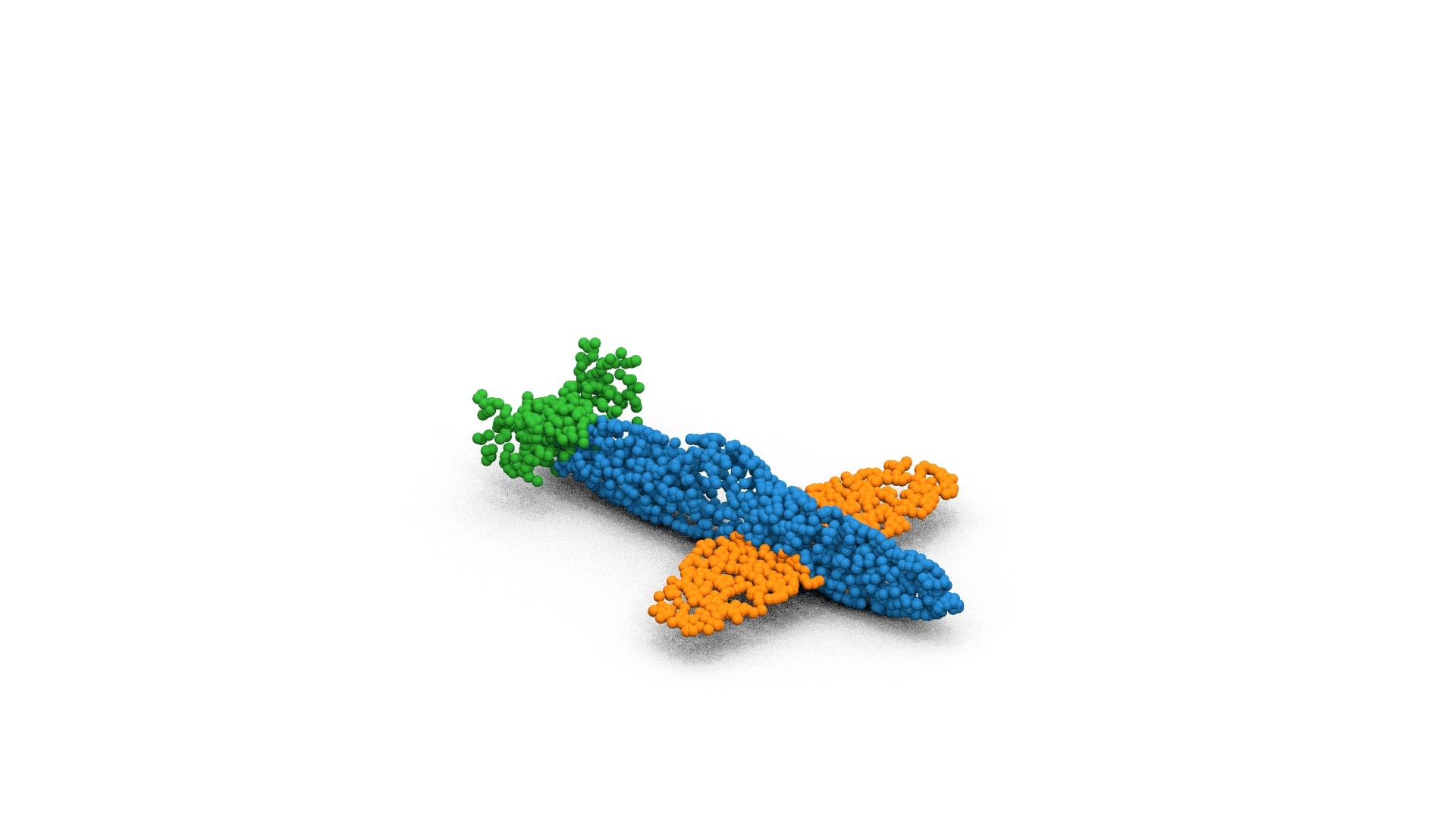}
        \put(-15,23){\rotatebox{90}{\scriptsize GT}}
    \end{overpic}
\end{minipage}
\begin{minipage}{0.12\textwidth}
    \begin{overpic}[width=\textwidth,trim=680 220 645 465,clip]{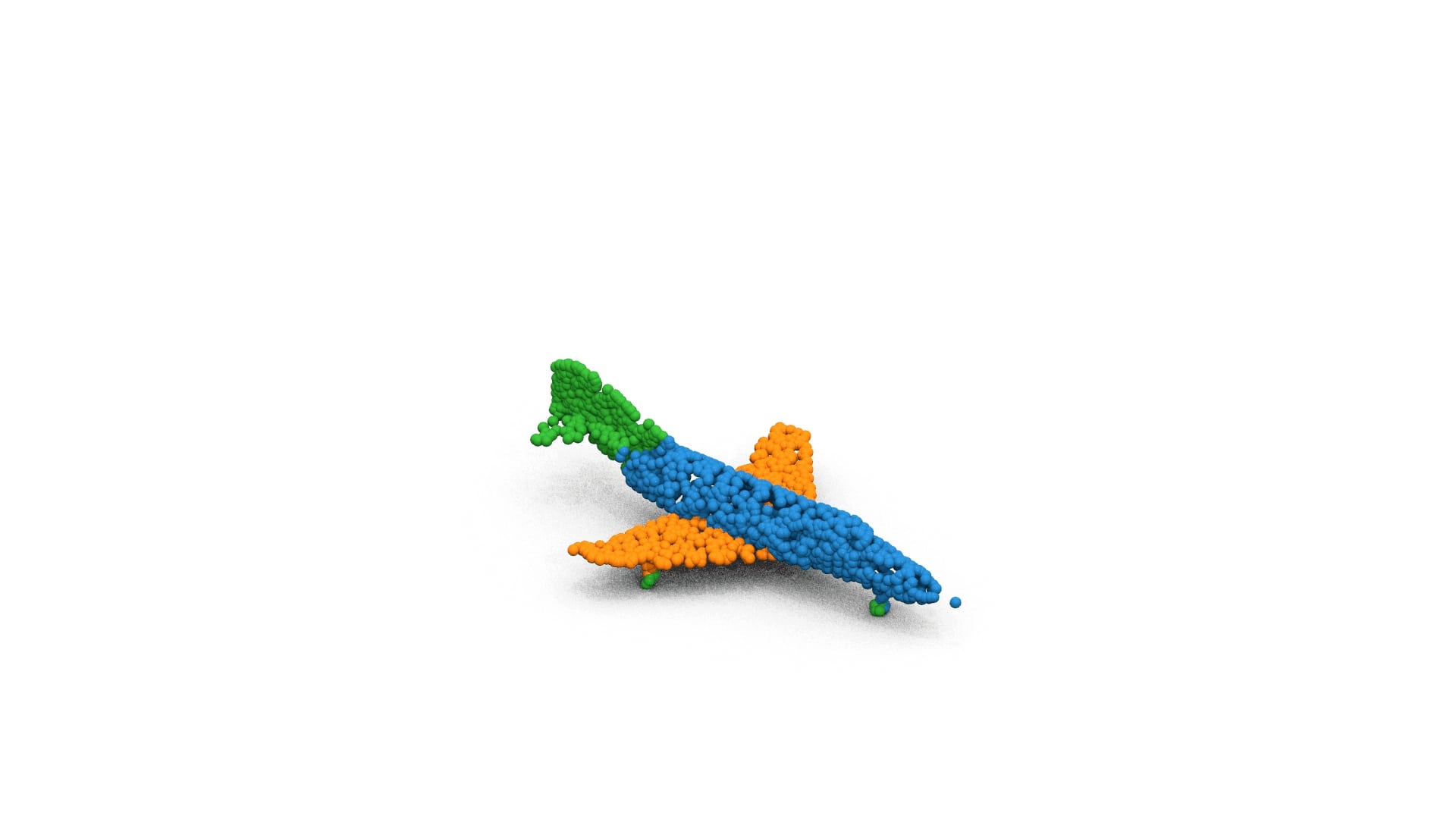}
    \end{overpic}
\end{minipage}
\begin{minipage}{0.12\textwidth}
    \begin{overpic}[width=\textwidth,trim=460 100 470 340,clip]{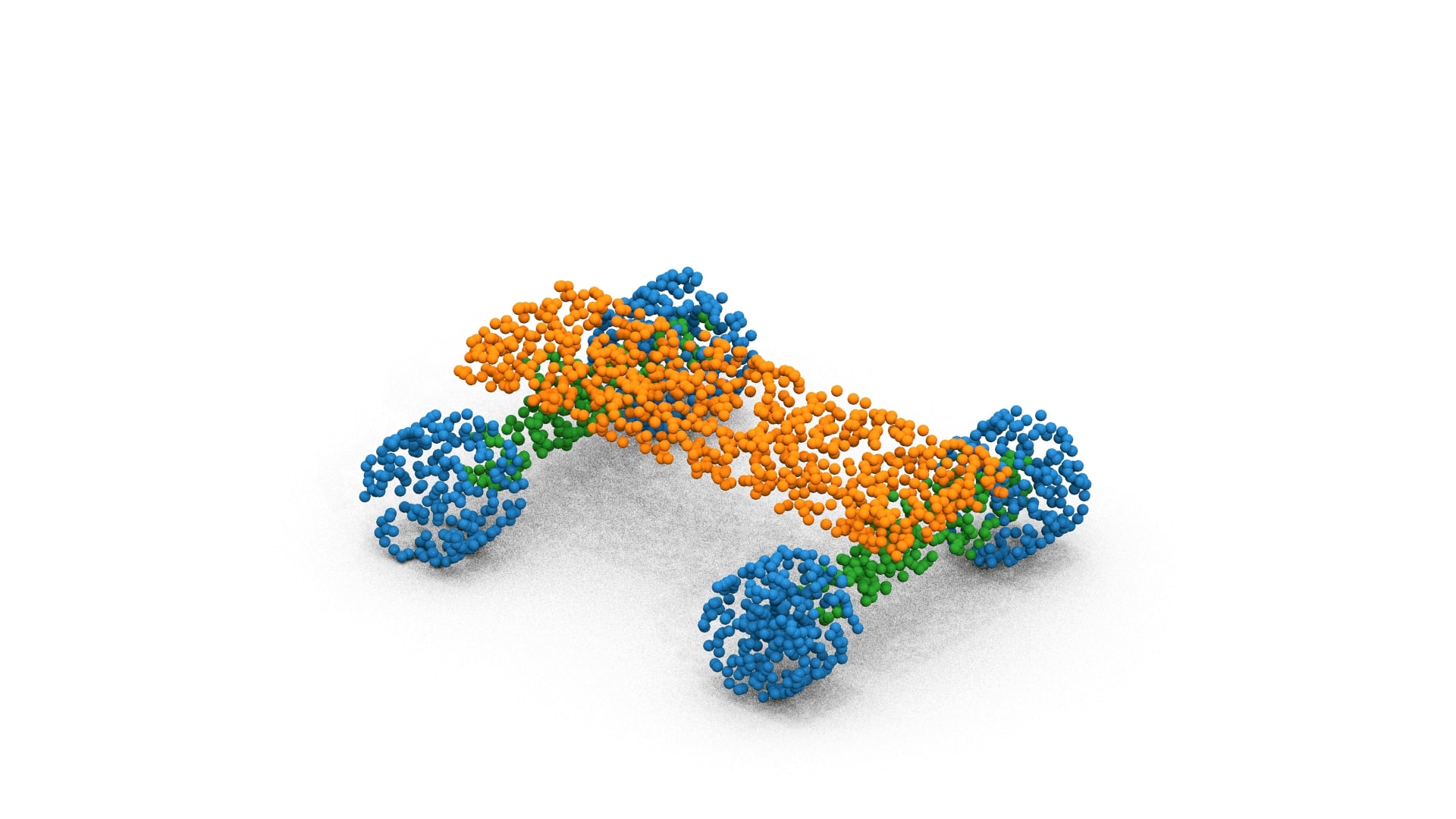}
    \end{overpic}
\end{minipage}
\begin{minipage}{0.12\textwidth}
    \begin{overpic}[width=\textwidth,trim=540 135 545 205,clip]{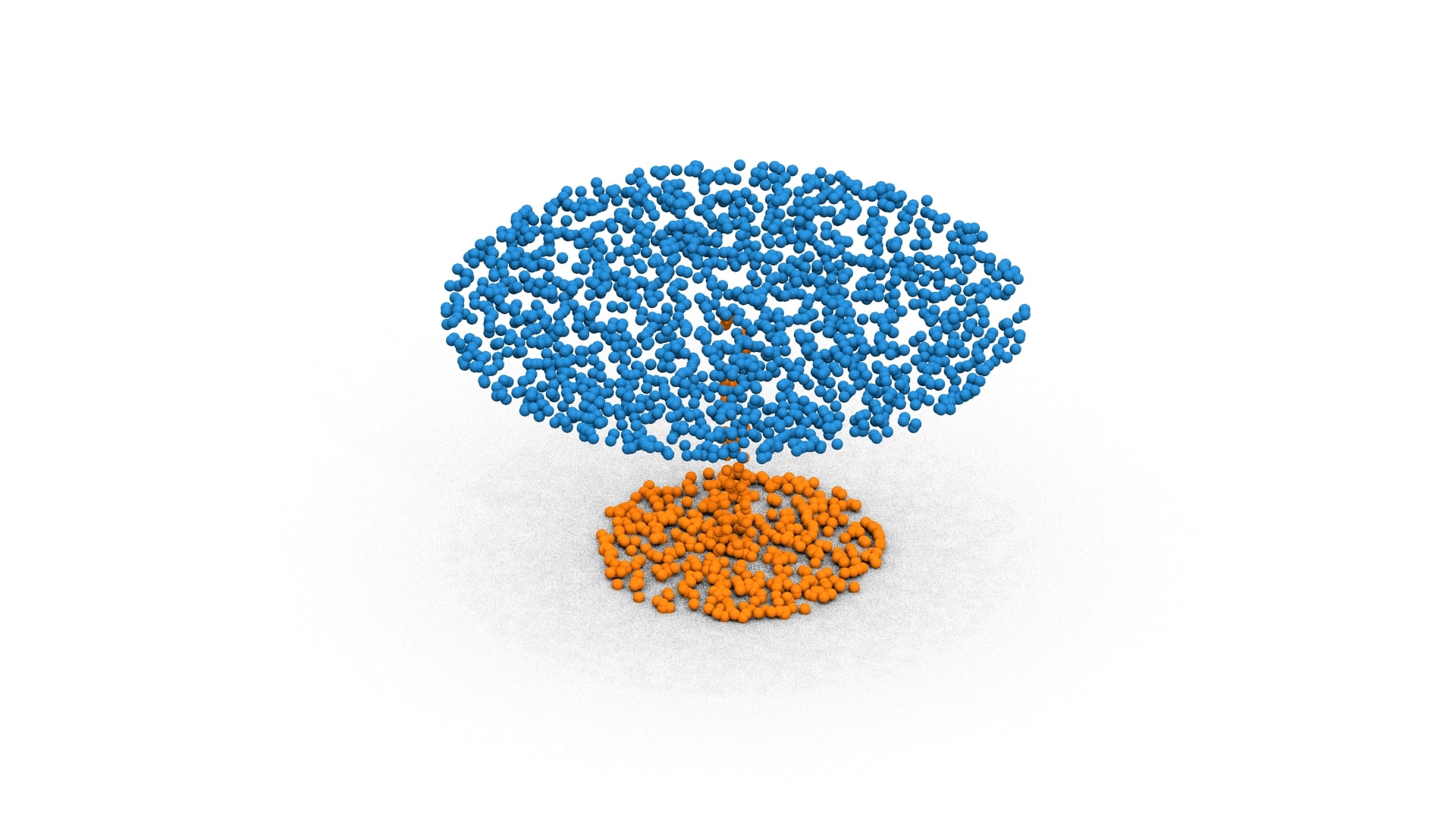}
    \end{overpic}
\end{minipage}
\begin{minipage}{0.12\textwidth}
    \begin{overpic}[width=\textwidth,trim=650 175 590 450,clip]{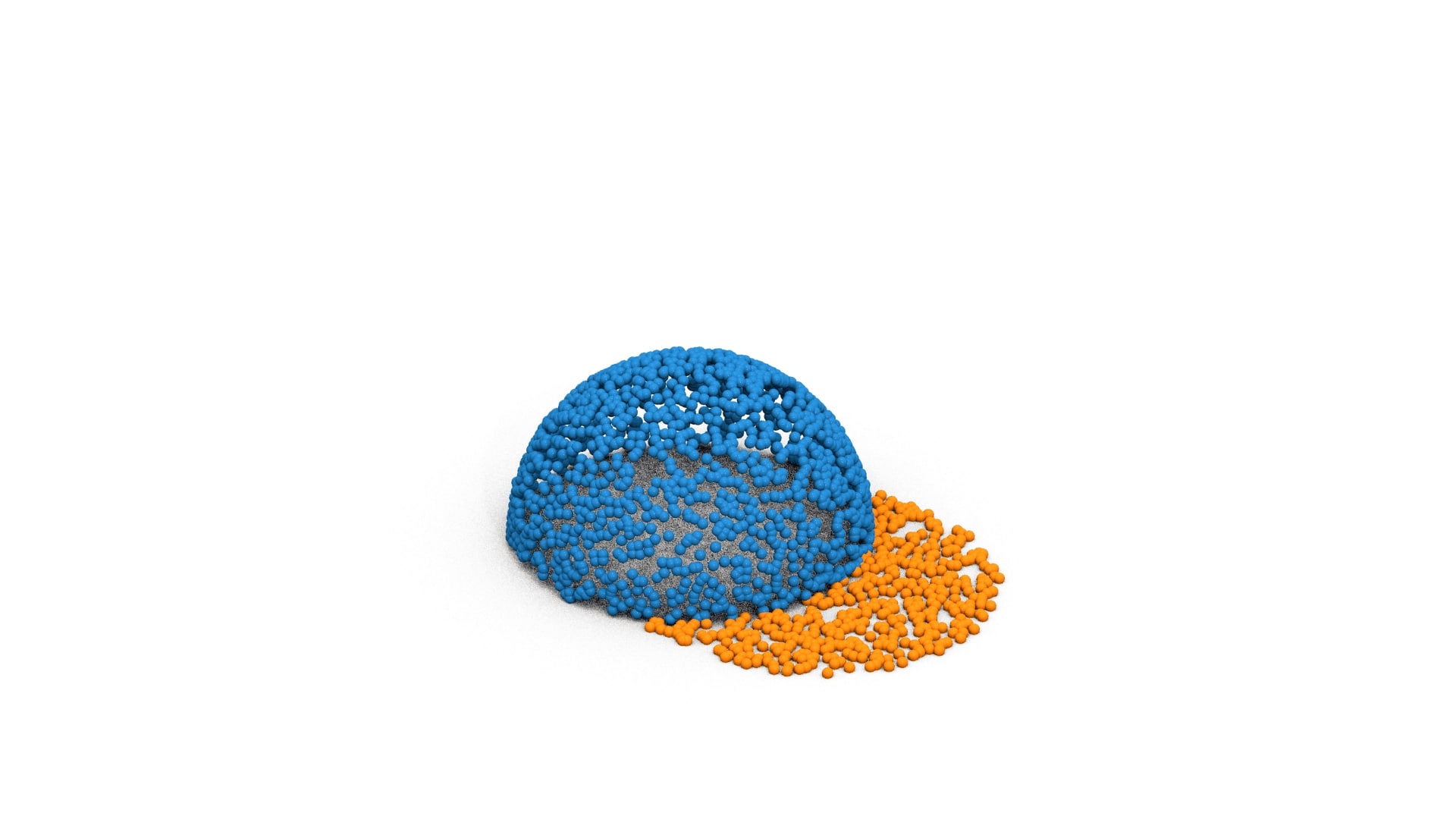}
    \end{overpic}
\end{minipage}
\begin{minipage}{0.12\textwidth}
    \begin{overpic}[width=\textwidth,trim=680 185 590 365,clip]{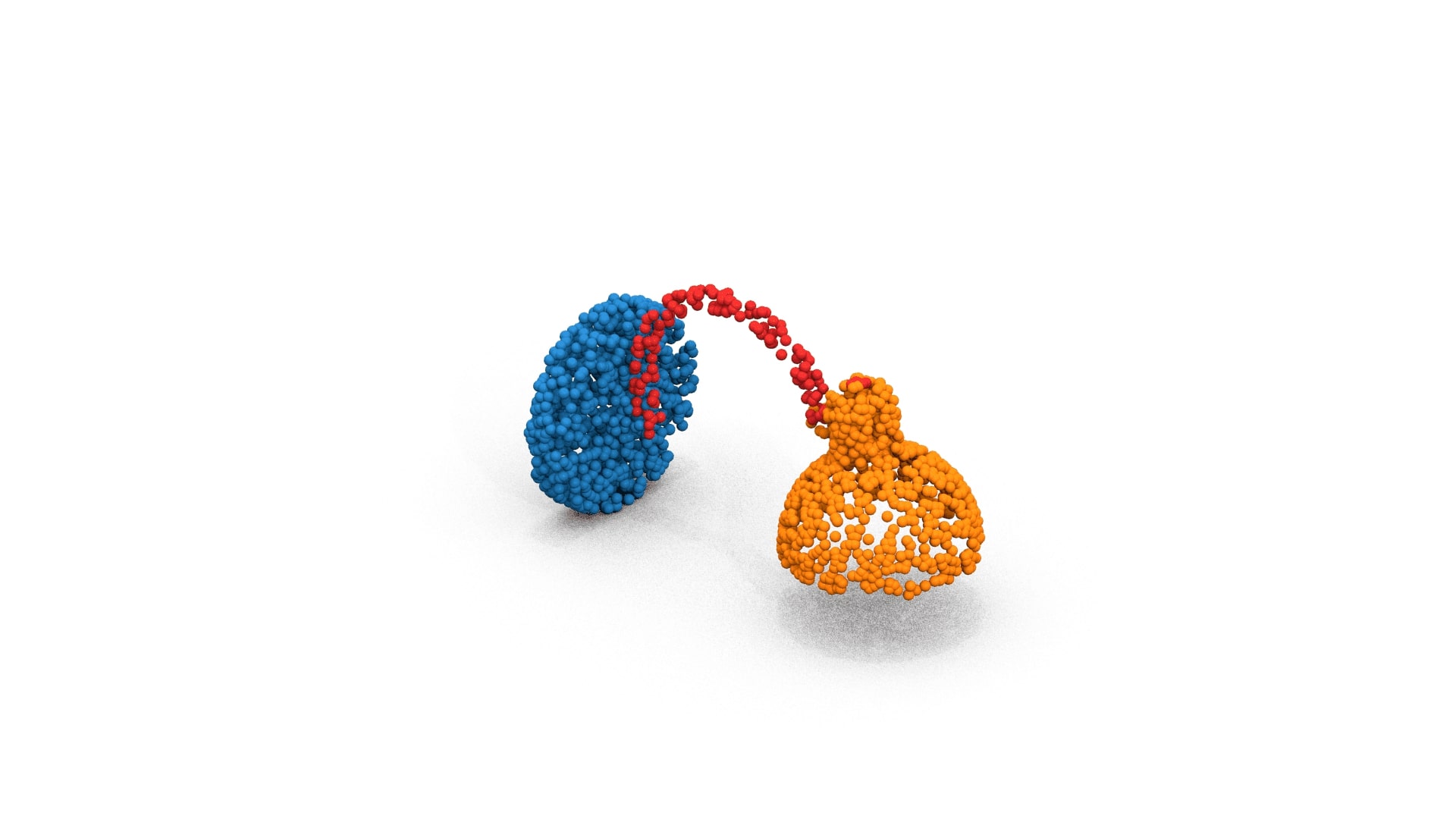}
    \end{overpic}
\end{minipage}
\begin{minipage}{0.12\textwidth}
    \begin{overpic}[width=\textwidth,trim=600 120 485 200,clip]{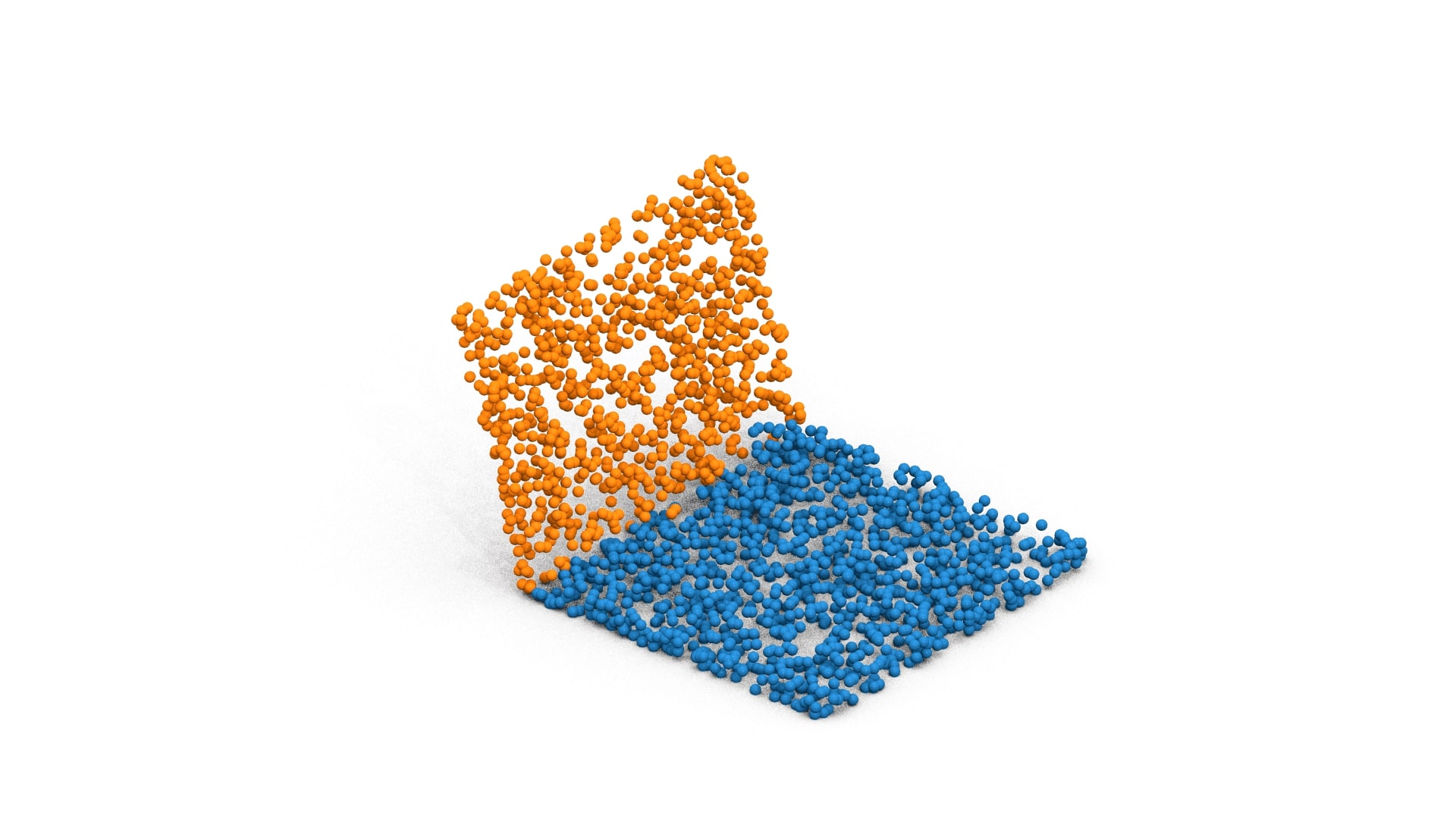}
    \end{overpic}
\end{minipage}\\

\vspace{-3mm}
\caption{
    Qualitative results on ShapeNetPart~\cite{yi2016shapenetpart}.
    Top to bottom: PointCLIPv2~\cite{zhu_pointclip_2023}, \acronym, ground-truth.
    These results show that PointCLIPv2 often struggles in describing and segmenting some parts, such as the wheels of the skateboard or the wings of the plane.
    \acronym instead produces a better segmentation, with more uniform part segments and sharper part boundaries.
}%
\label{fig:qual}
\end{figure*}

\begin{figure*}[!htb]
\centering

%
%
\resizebox{1.6\columnwidth}{!}{%
\begin{minipage}{0.15\textwidth}
    \begin{overpic}[width=\textwidth,trim=680 180 610 295,clip] {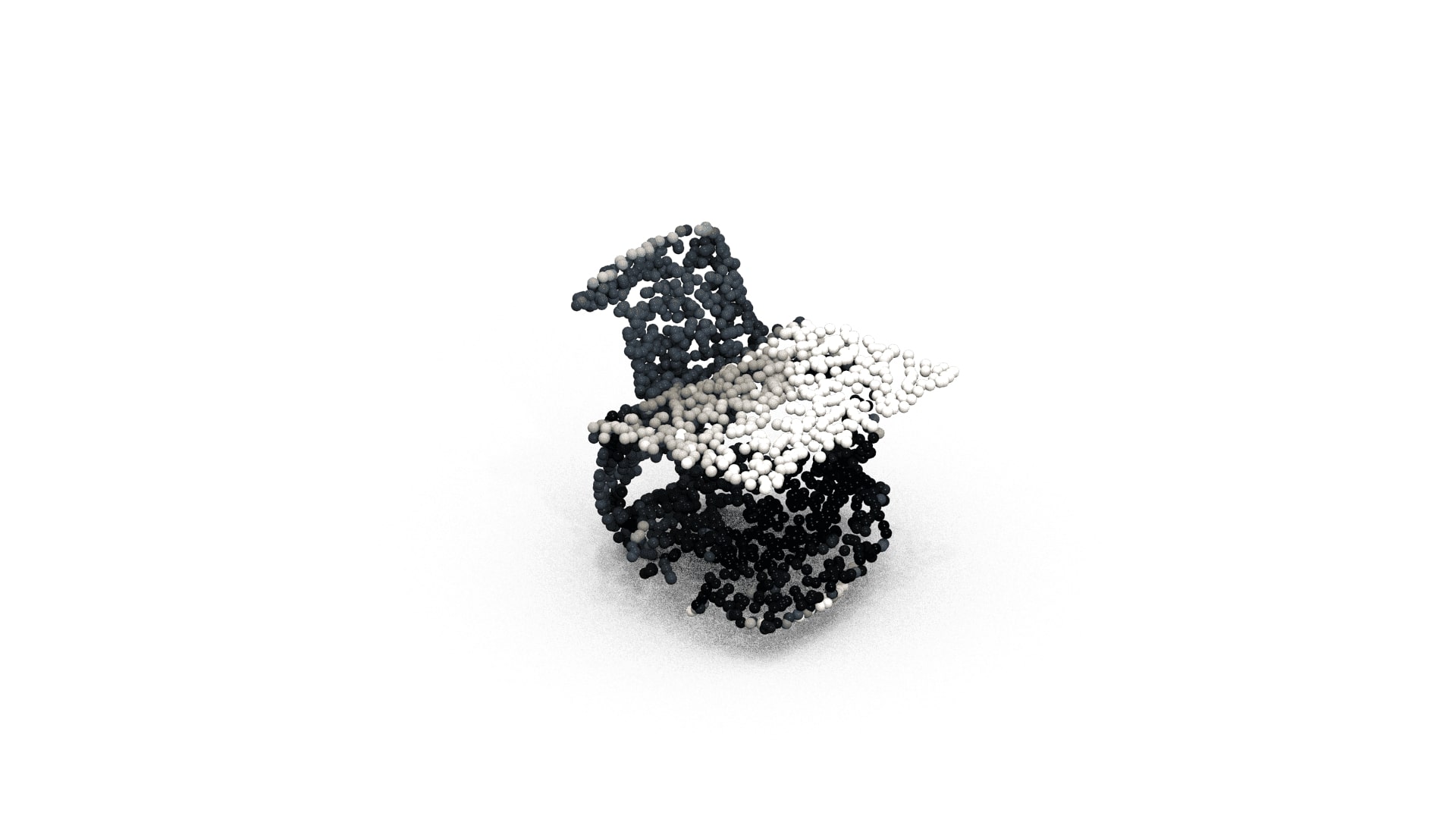}
    \put(-15,35){\rotatebox{90}{\scriptsize Input}}
    \end{overpic}
\end{minipage}
\begin{minipage}{0.18\textwidth} 
    \begin{overpic}[width=\textwidth,trim=555 170 530 340,clip]{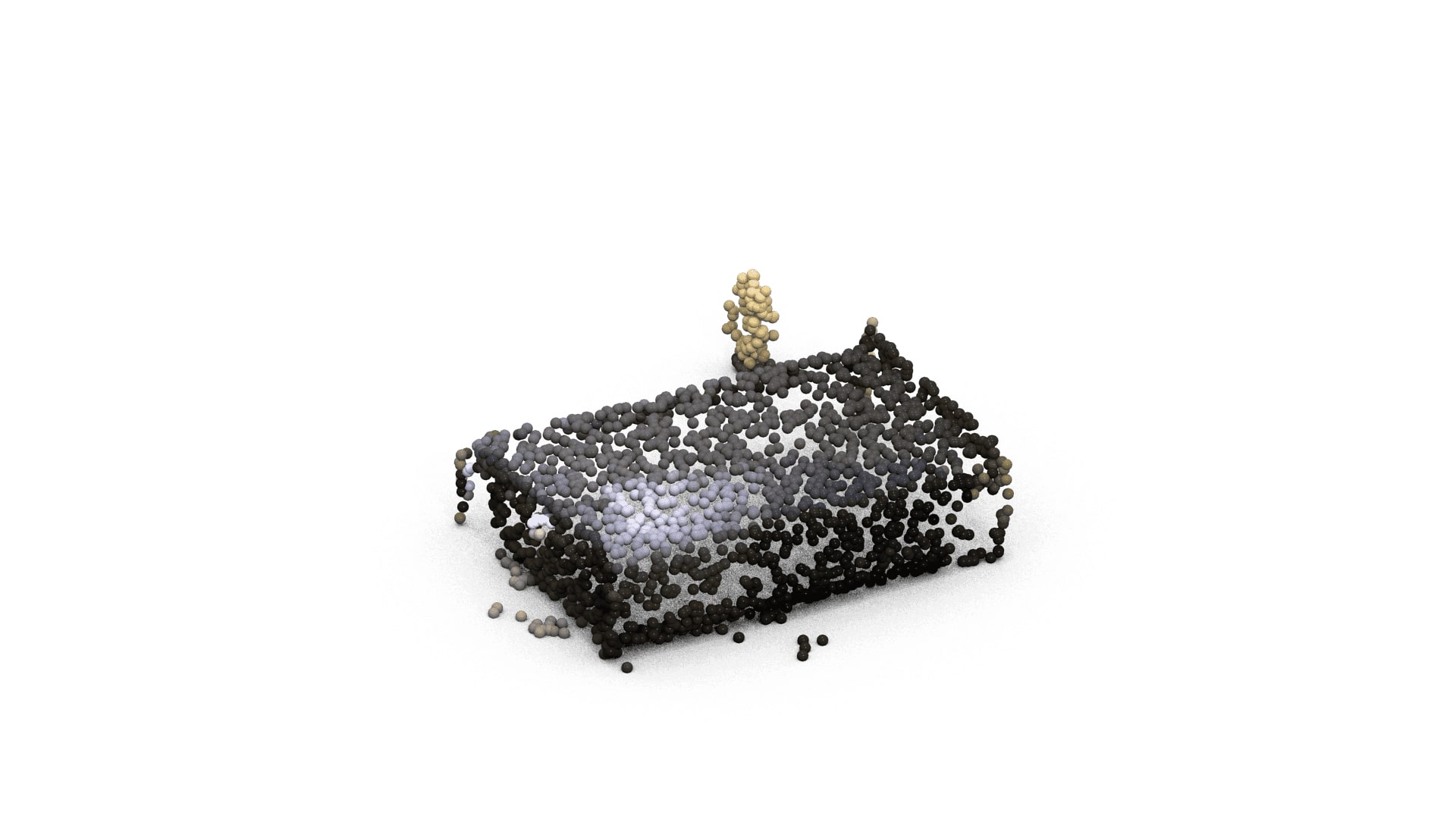}
    \end{overpic}
\end{minipage}
\begin{minipage}{0.18\textwidth} 
    \begin{overpic}[width=\textwidth,trim=570 145 540 285,clip]{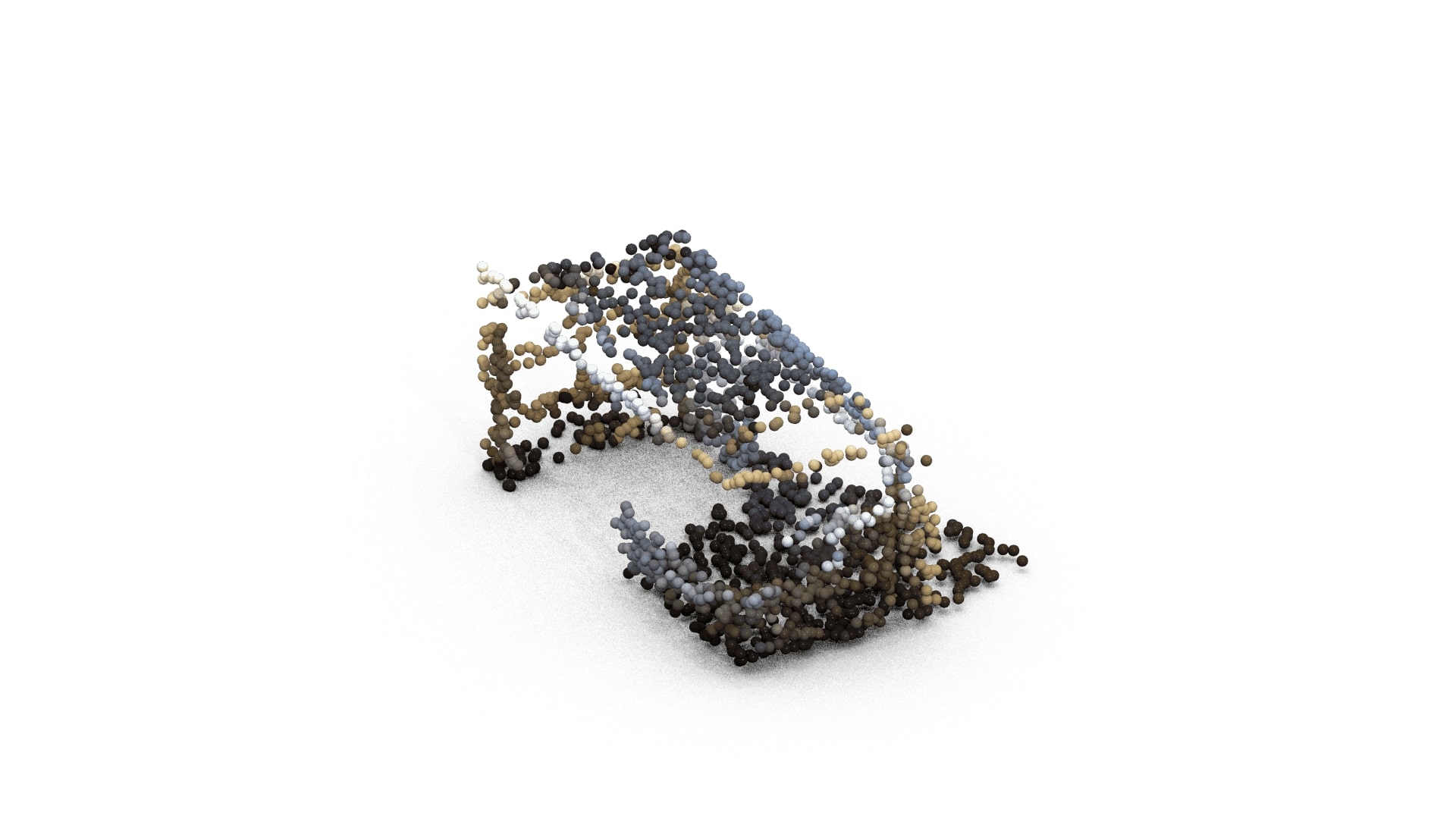}
    \end{overpic}
\end{minipage}
\begin{minipage}{0.18\textwidth} 
    \begin{overpic}[width=\textwidth,trim=610 110 500 385,clip]{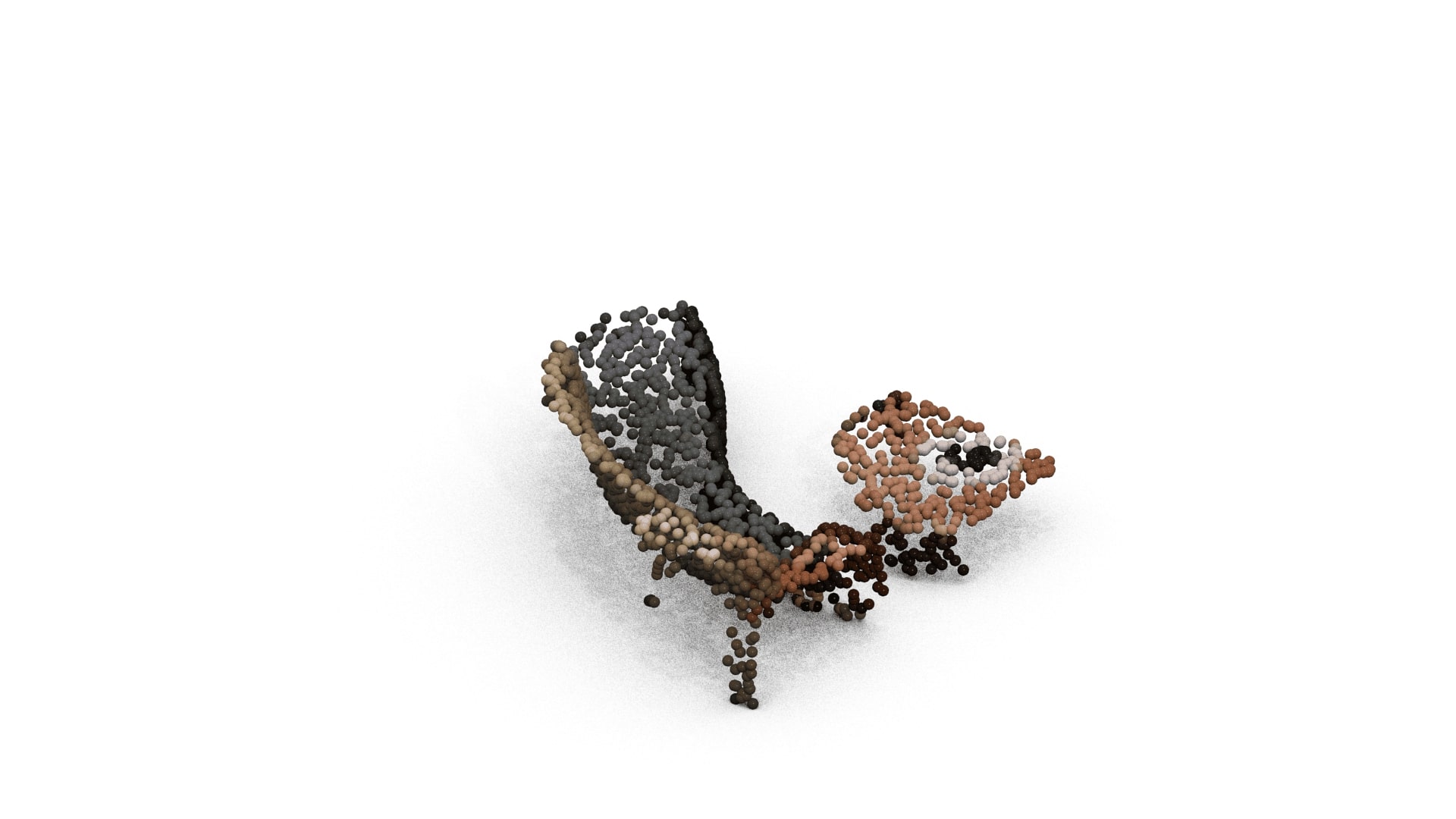}
    \end{overpic}
\end{minipage}
\begin{minipage}{0.14\textwidth}
    \centering
    \begin{overpic}[width=\textwidth,trim=710 230 720 300,clip]{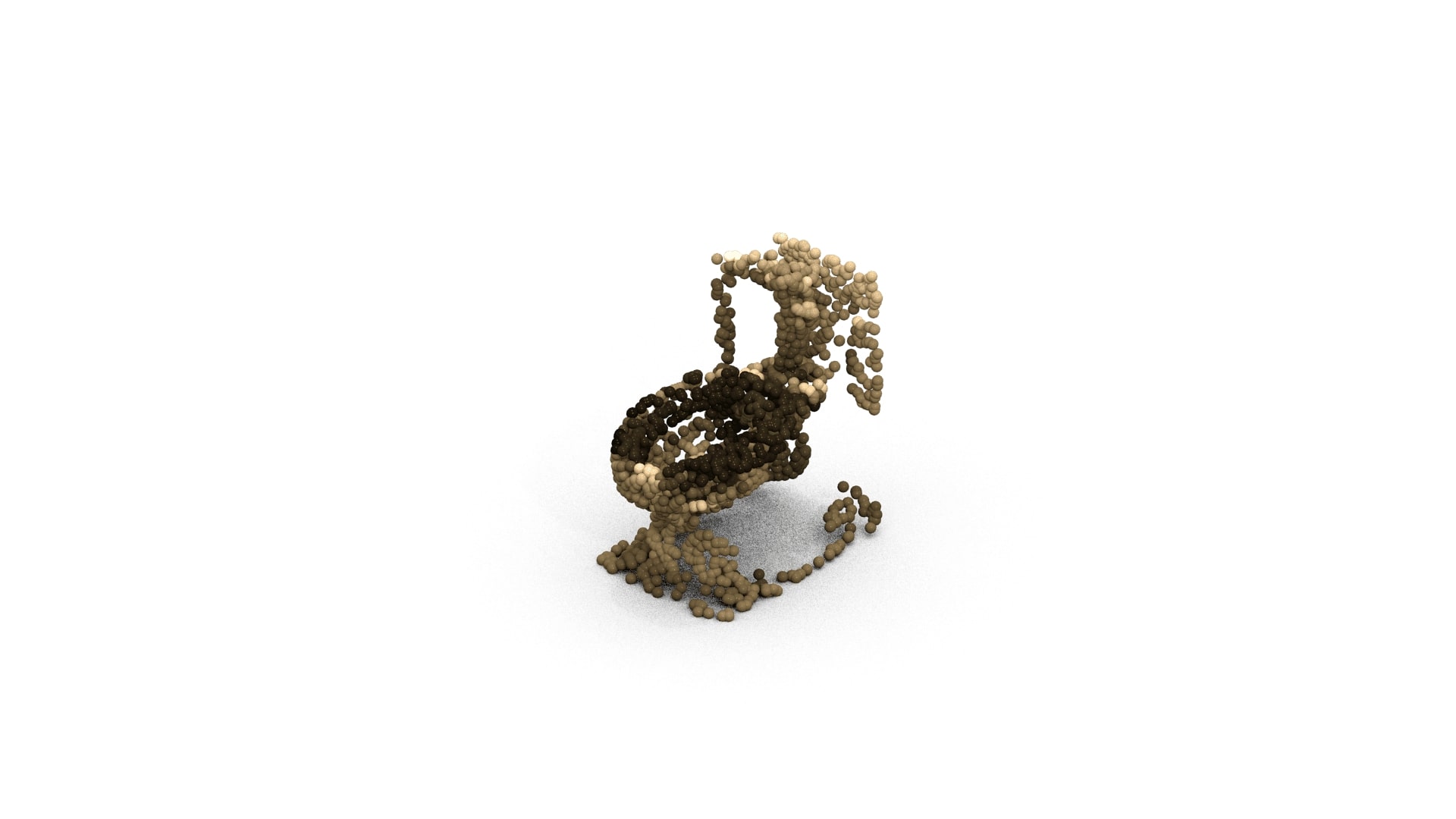}
    \end{overpic}
\end{minipage}
}\\
%
%
\resizebox{1.6\columnwidth}{!}{%
\begin{minipage}{0.15\textwidth}
    \begin{overpic}[width=\textwidth,trim=680 180 610 295,clip]{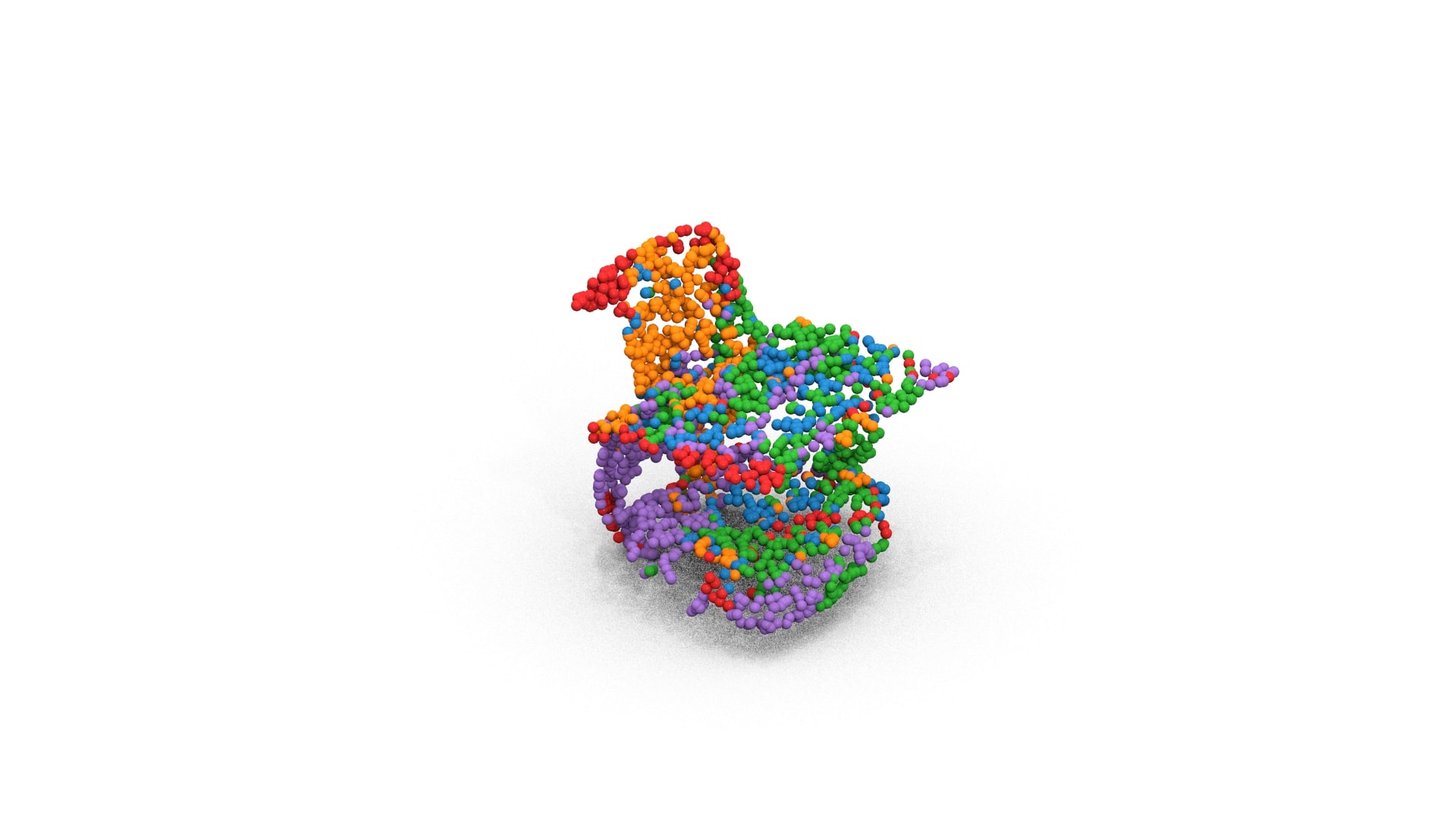}
    \put(-15,10){\rotatebox{90}{\scriptsize PointCLIPv2}}
    \end{overpic}
\end{minipage}
\begin{minipage}{0.18\textwidth}
    \begin{overpic}[width=\textwidth,trim=555 170 530 340,clip]{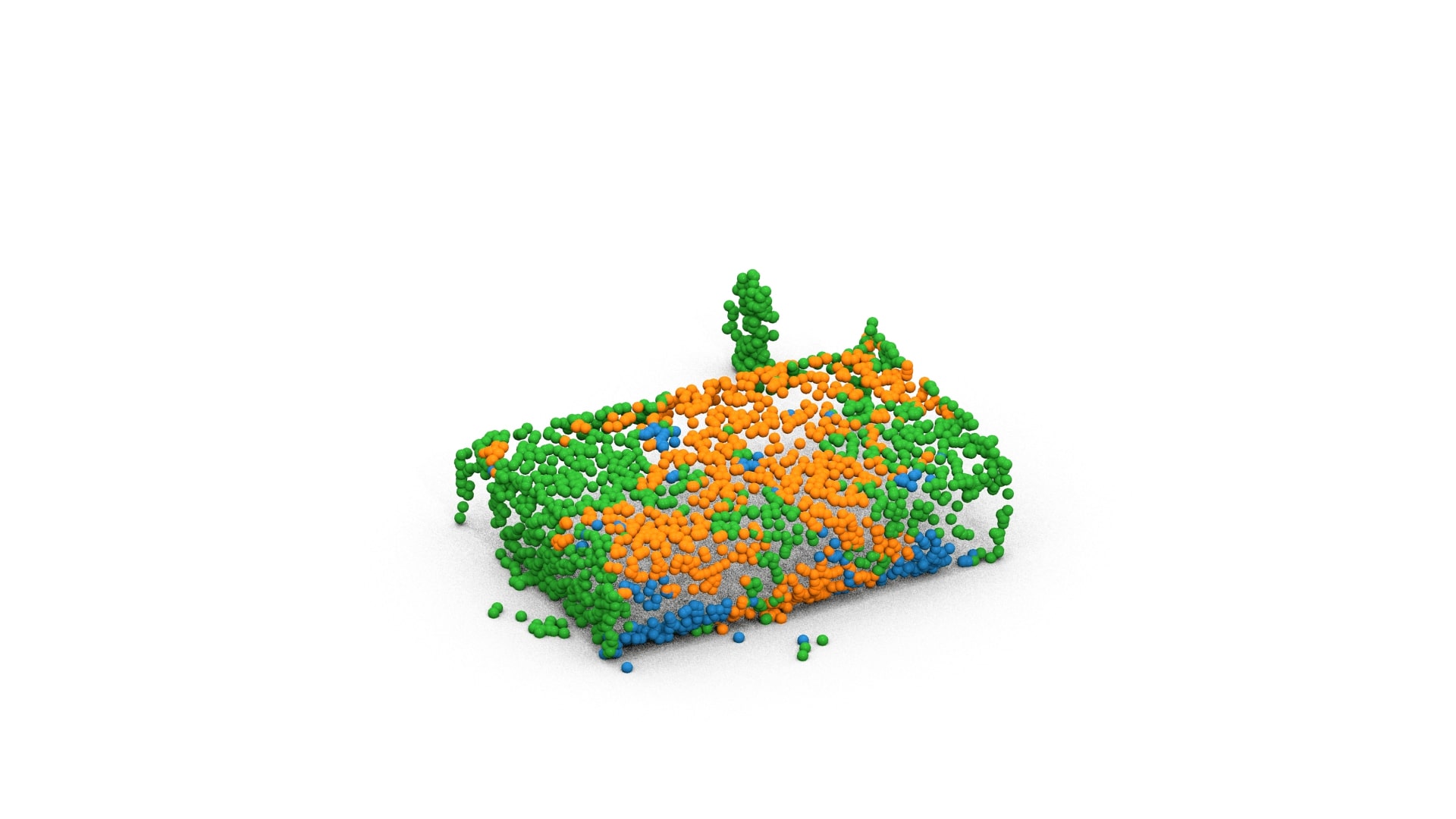}
    \end{overpic}
\end{minipage}
\begin{minipage}{0.18\textwidth}
    \begin{overpic}[width=\textwidth,trim=570 145 540 285,clip]{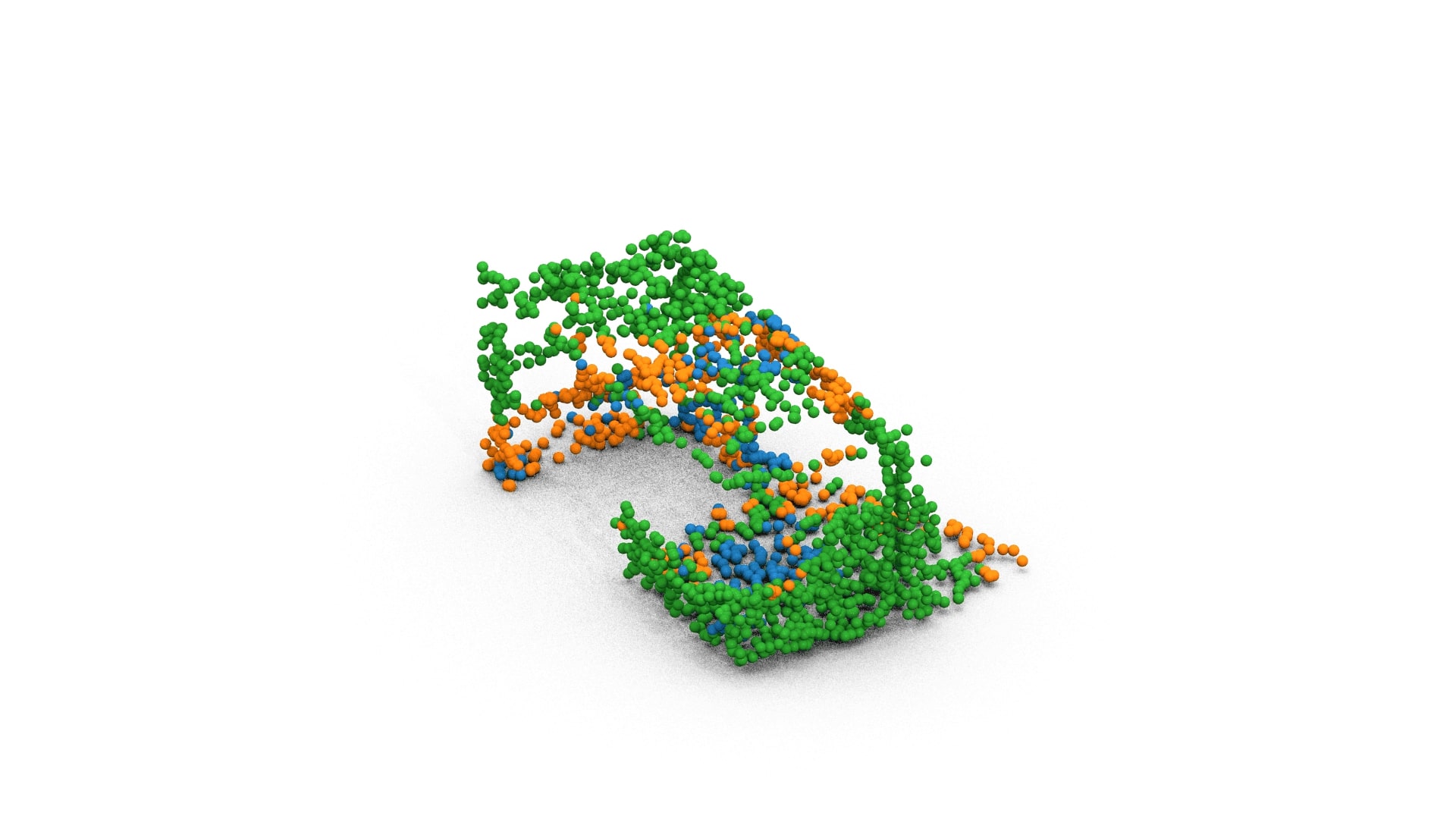}
    \end{overpic}
\end{minipage}
\begin{minipage}{0.18\textwidth}
    \begin{overpic}[width=\textwidth,trim=610 110 500 385,clip]{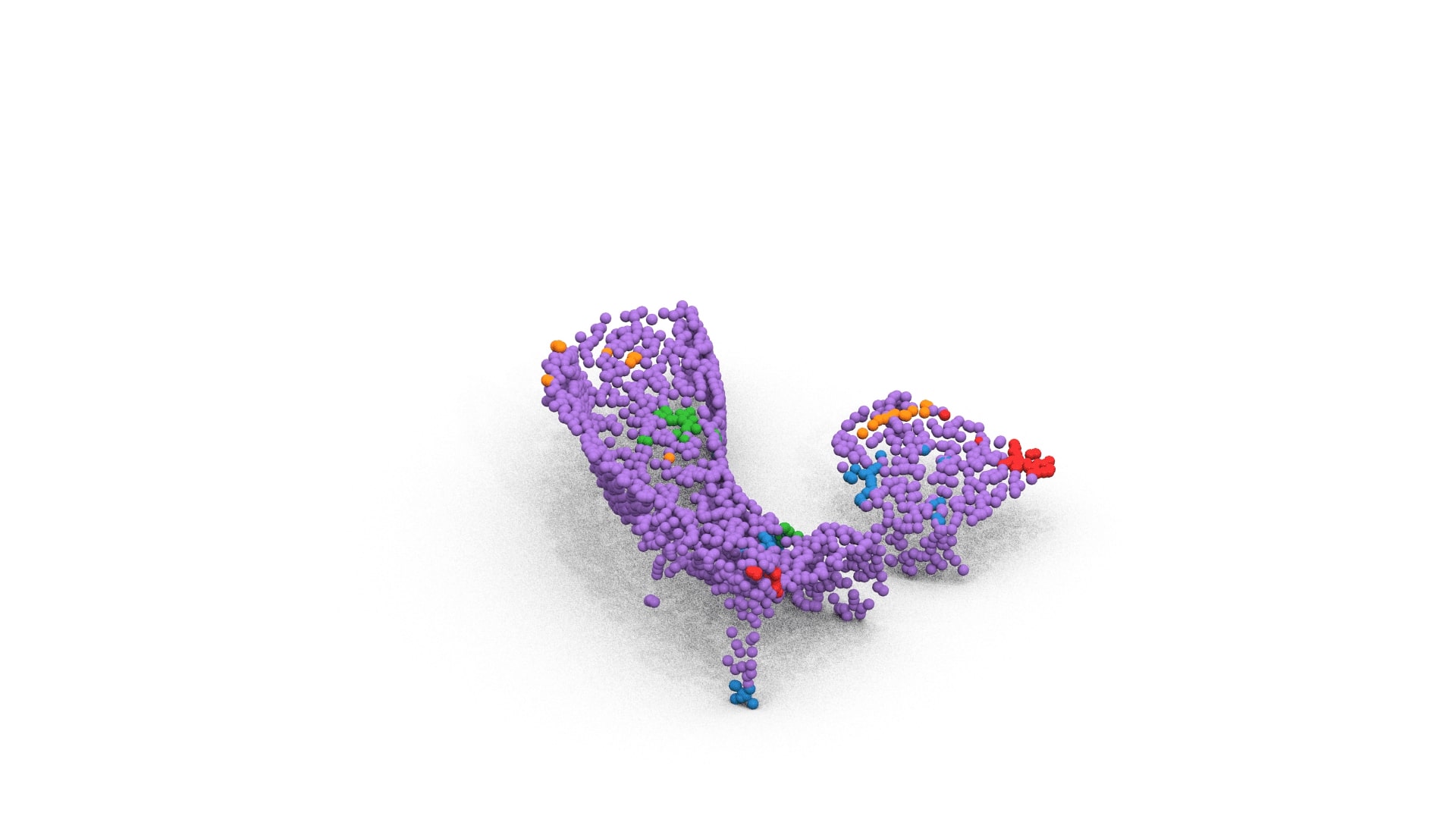}
    \end{overpic}
\end{minipage}
\begin{minipage}{0.14\textwidth}
    \centering
    \begin{overpic}[width=\textwidth,trim=710 230 720 300,clip]{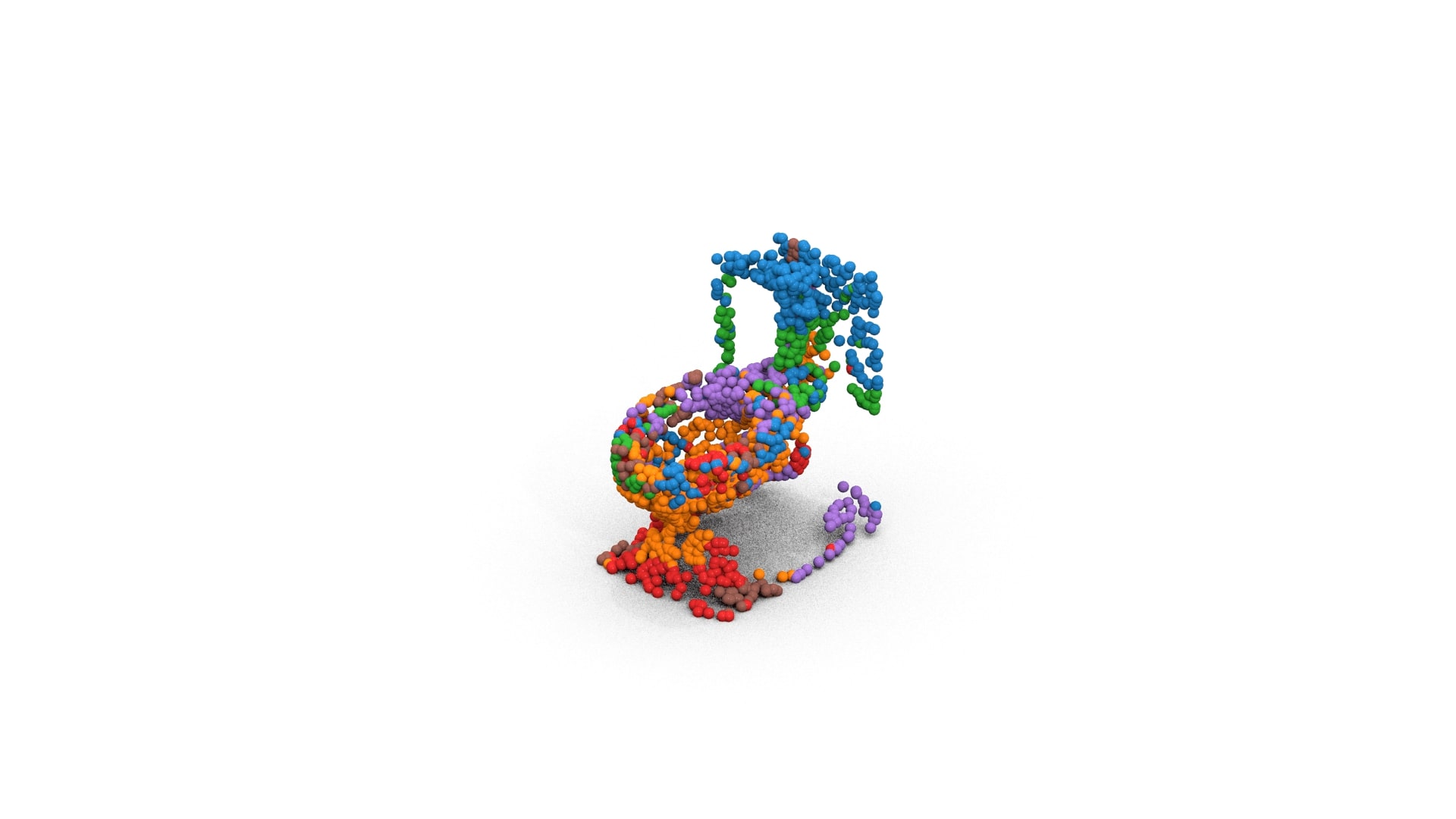}
    \end{overpic}
\end{minipage}
}\\
%
%
\resizebox{1.6\columnwidth}{!}{%
\begin{minipage}{0.15\textwidth}
    \begin{overpic}[width=\textwidth,trim=680 180 610 295,clip]{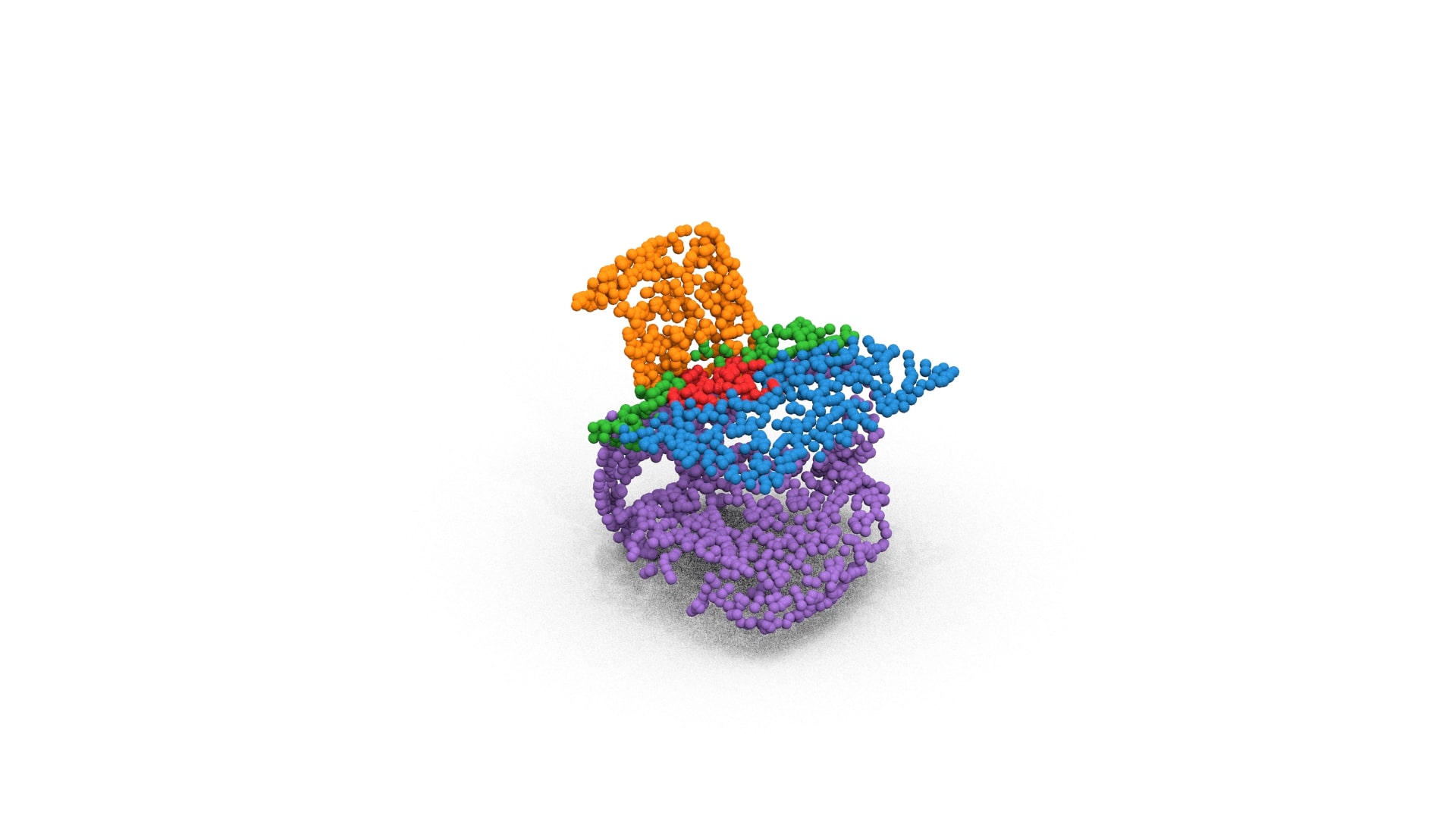}
    \put(-15,30){\rotatebox{90}{\scriptsize \acronym}}
    \end{overpic}
\end{minipage}
\begin{minipage}{0.18\textwidth}
    \begin{overpic}[width=\textwidth,trim=555 170 530 340,clip]{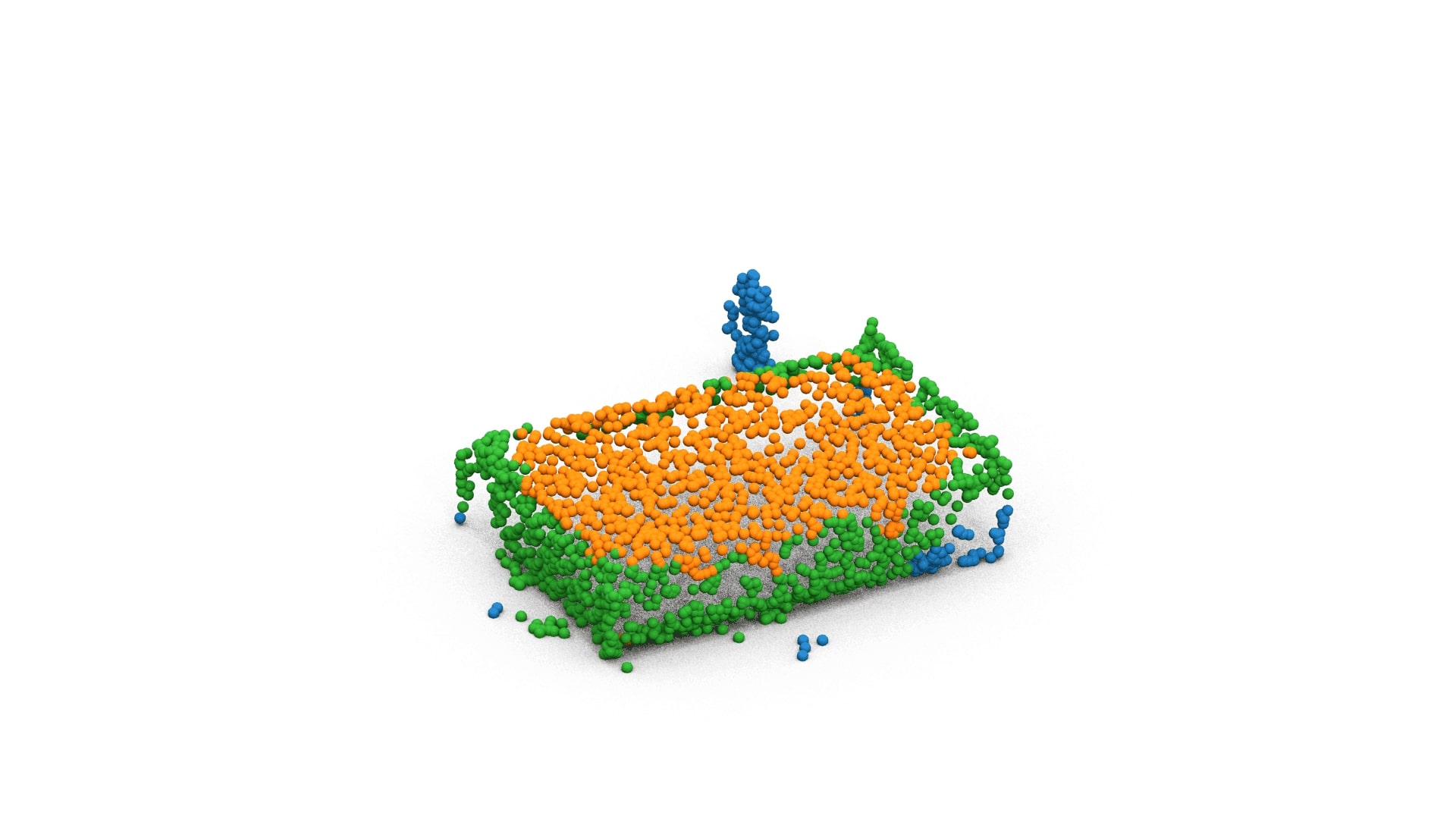}
    \end{overpic}
\end{minipage}
\begin{minipage}{0.18\textwidth}
    \begin{overpic}[width=\textwidth,trim=570 145 540 285,clip]{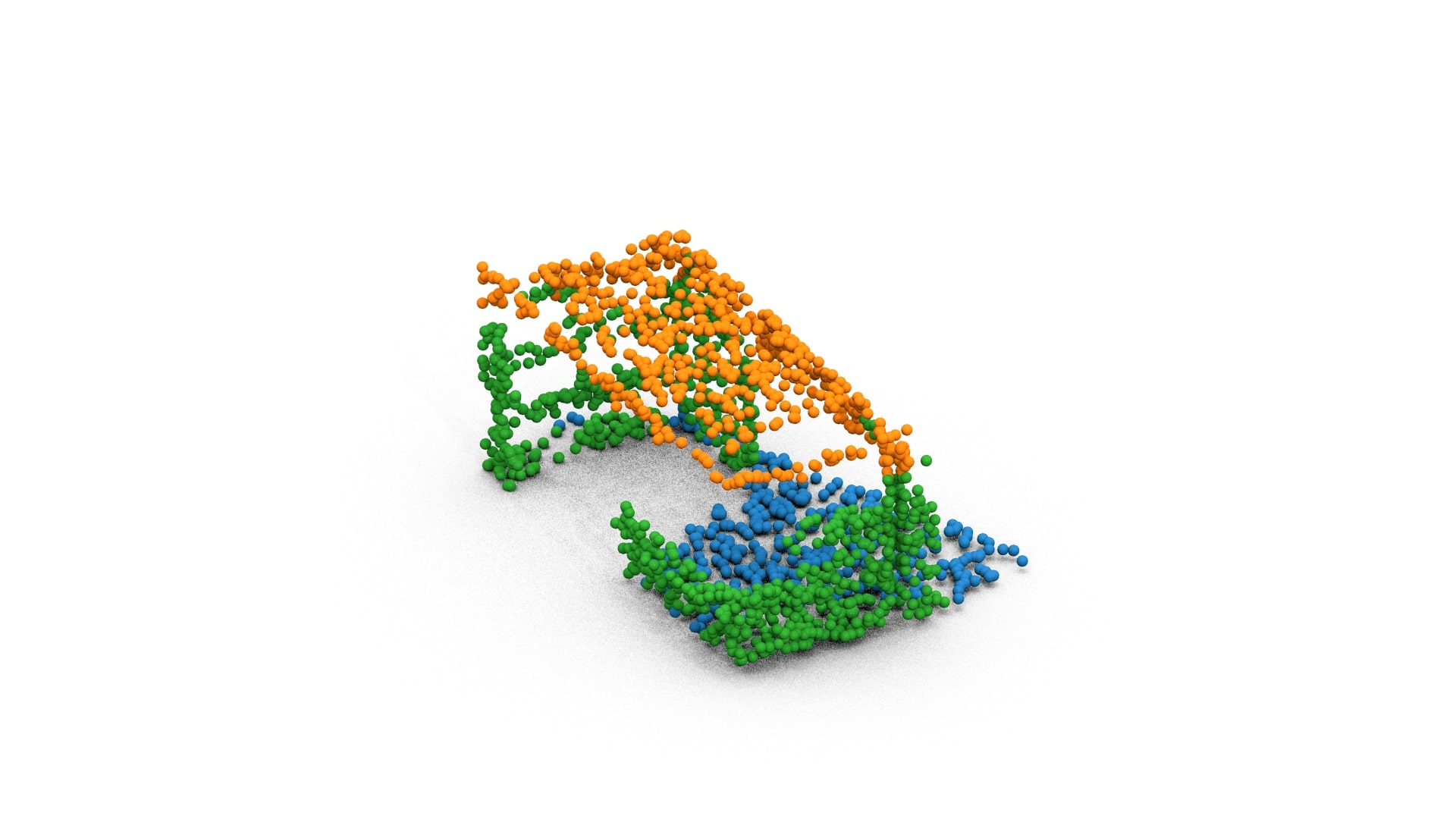}
    \end{overpic}
\end{minipage}
\begin{minipage}{0.18\textwidth}
    \begin{overpic}[width=\textwidth,trim=610 110 500 385,clip]{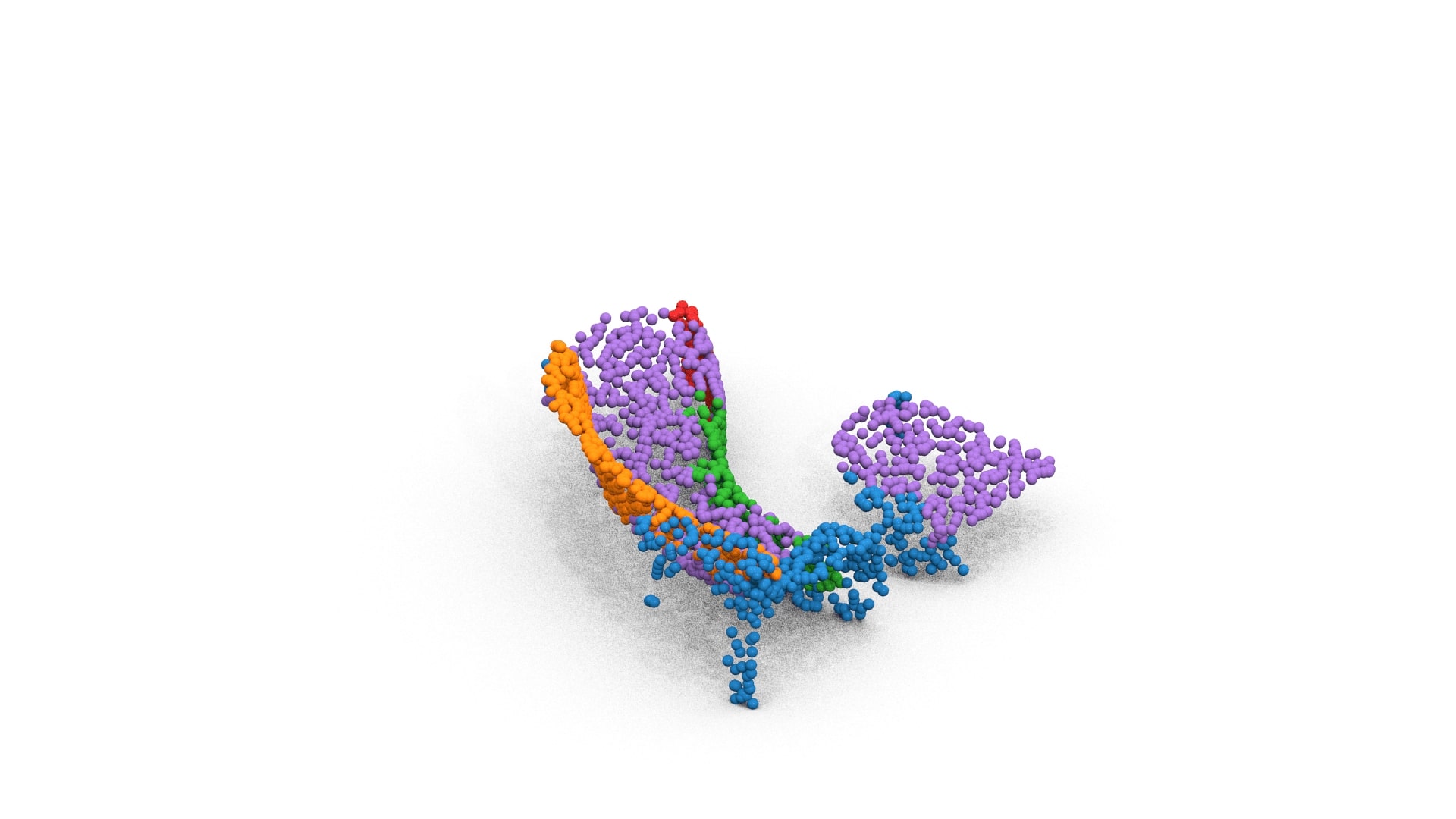}
    \end{overpic}
\end{minipage}
\begin{minipage}{0.14\textwidth} 
    \centering
    \begin{overpic}[width=\textwidth,trim=710 230 720 300,clip]{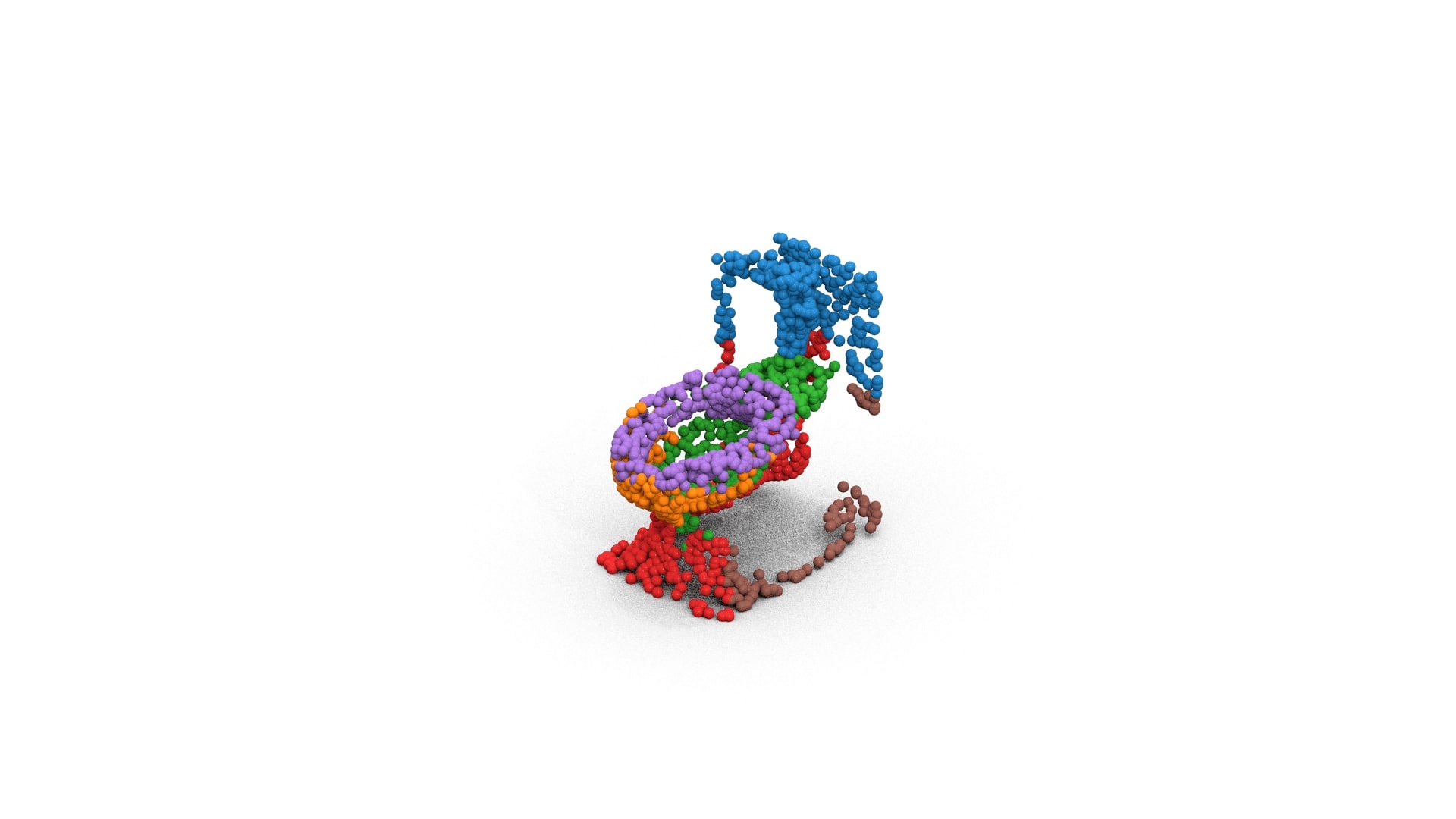}
    \end{overpic}
\end{minipage}
}\\
%
%
\resizebox{1.6\columnwidth}{!}{%
\begin{minipage}{0.15\textwidth}
    \begin{overpic}[width=\textwidth,trim=680 180 610 295,clip]{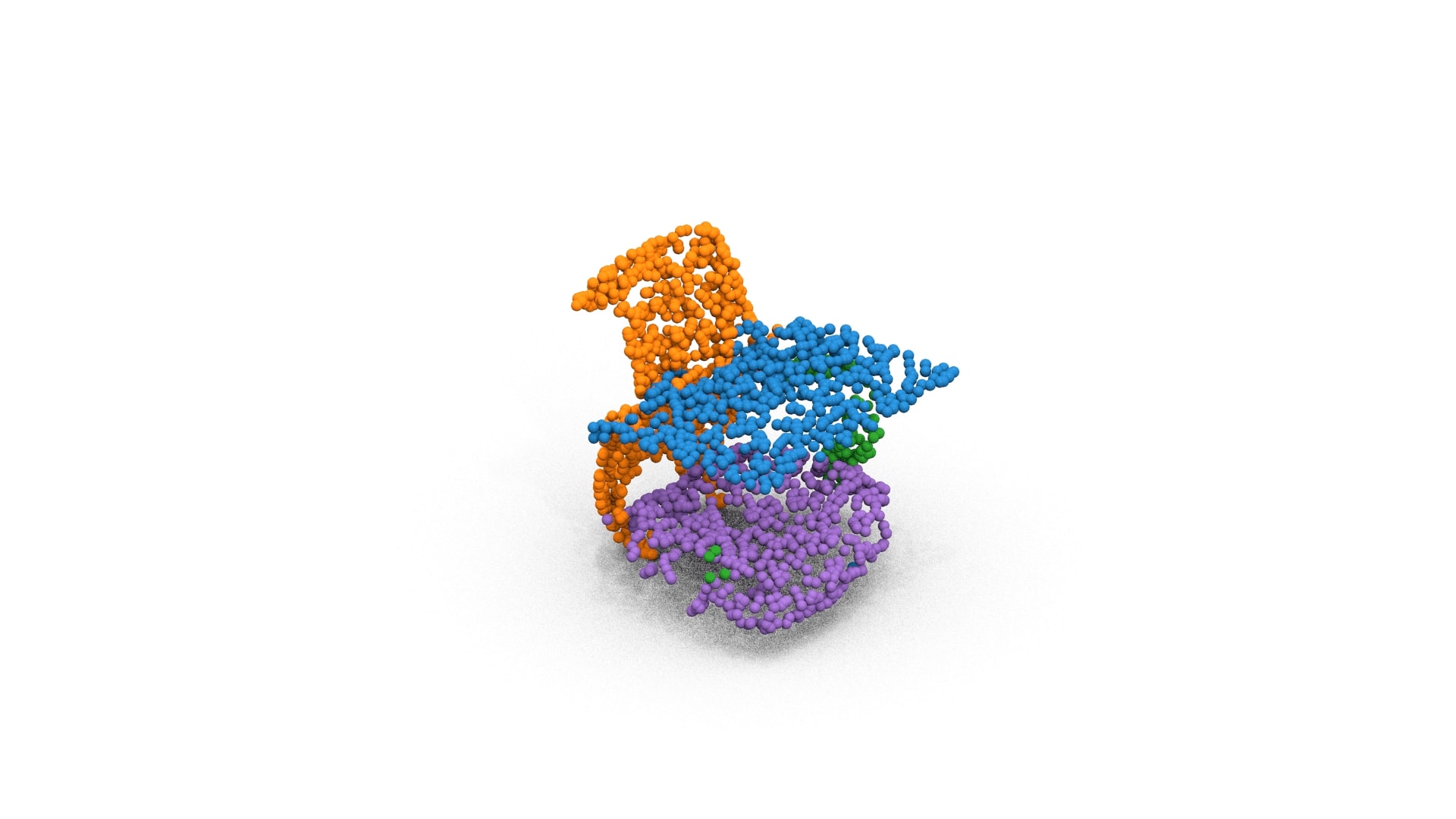}
    \put(-15,40){\rotatebox{90}{\scriptsize GT}}
    \end{overpic}
\end{minipage}
\begin{minipage}{0.18\textwidth}
    \begin{overpic}[width=\textwidth,trim=555 170 530 340,clip]{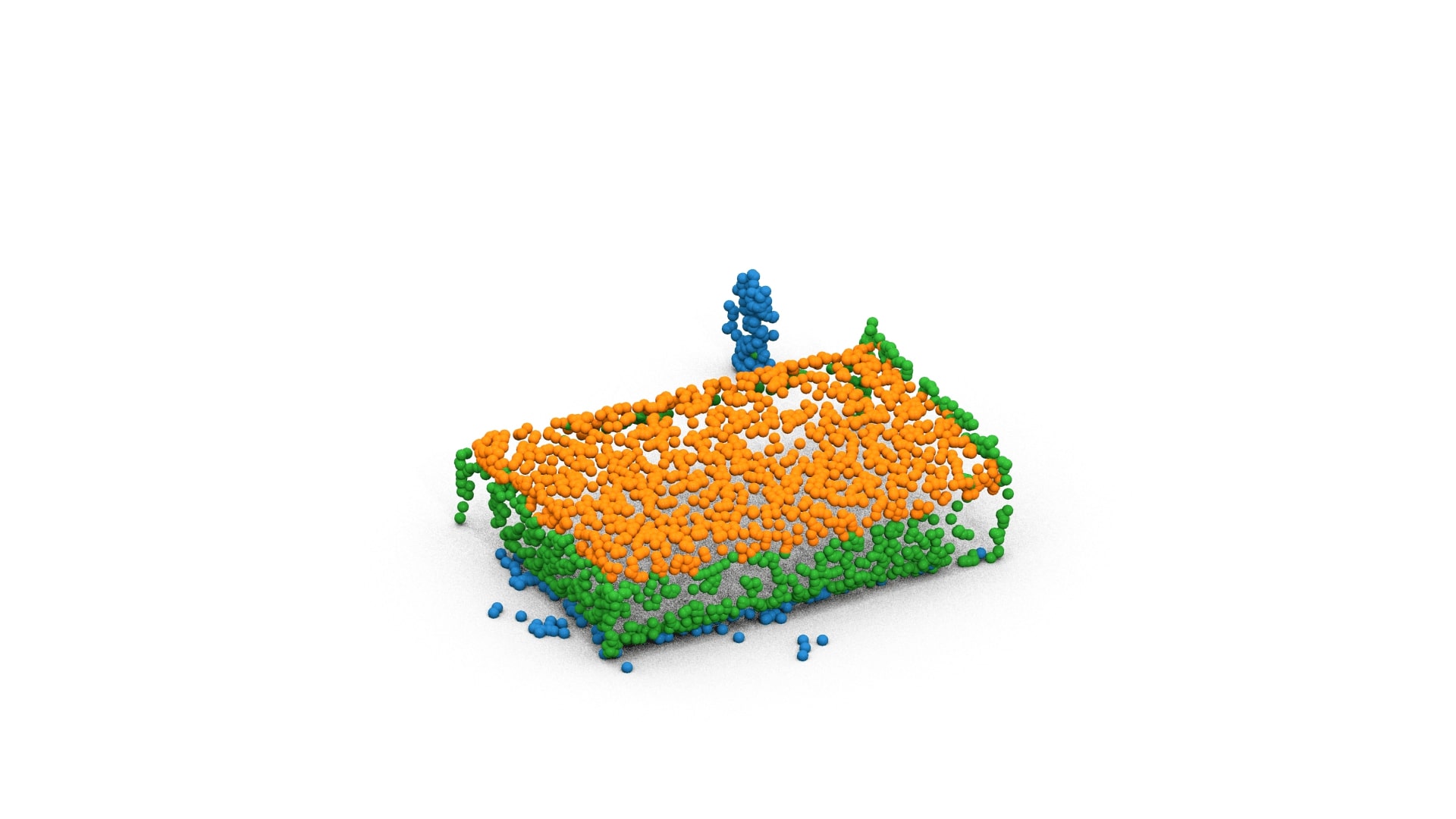}
    \end{overpic}
\end{minipage}
\begin{minipage}{0.18\textwidth}
    \begin{overpic}[width=\textwidth,trim=570 145 540 285,clip]{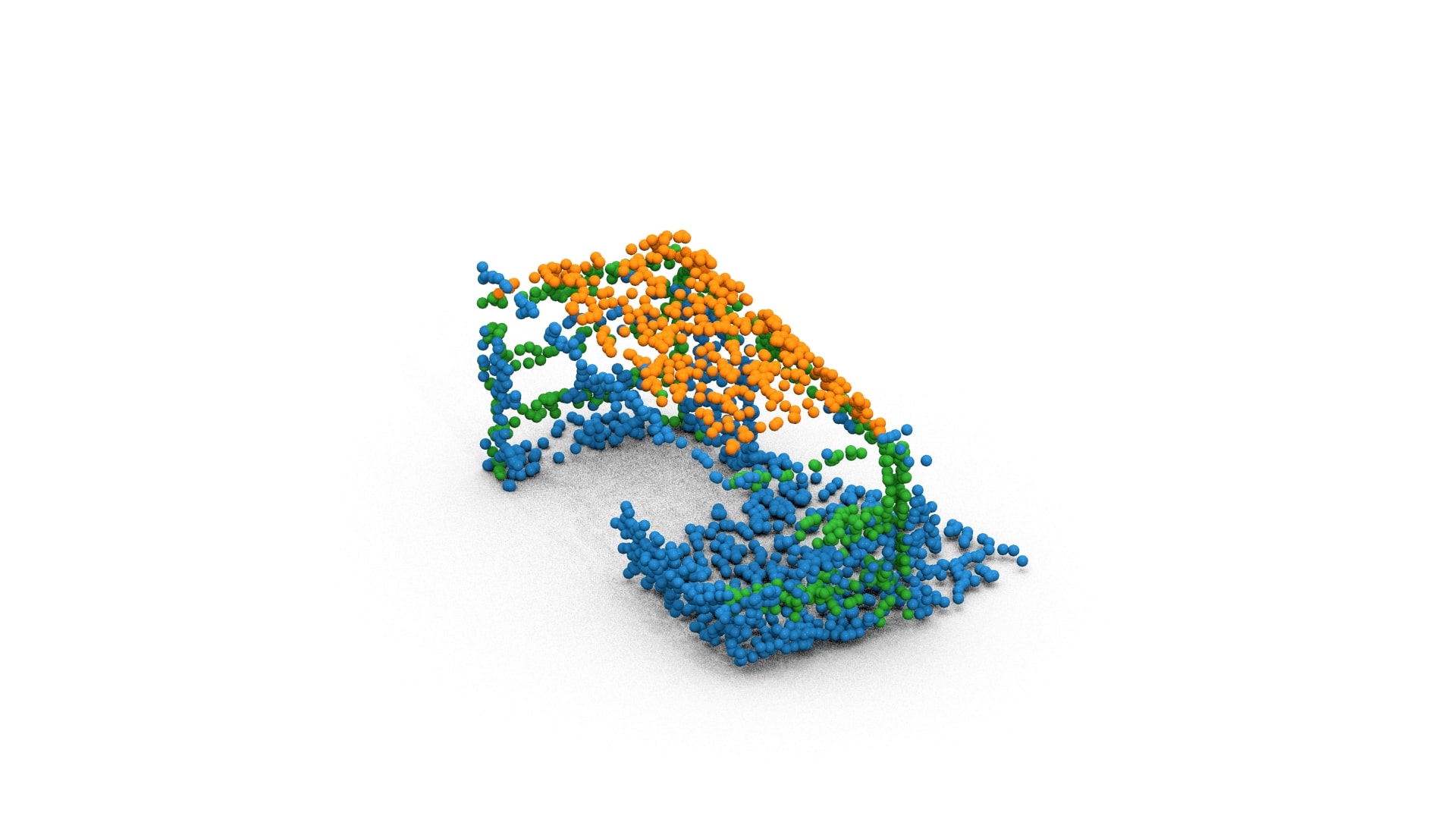}
    \end{overpic}
\end{minipage}
\begin{minipage}{0.18\textwidth}
    \begin{overpic}[width=\textwidth,trim=610 110 500 385,clip]{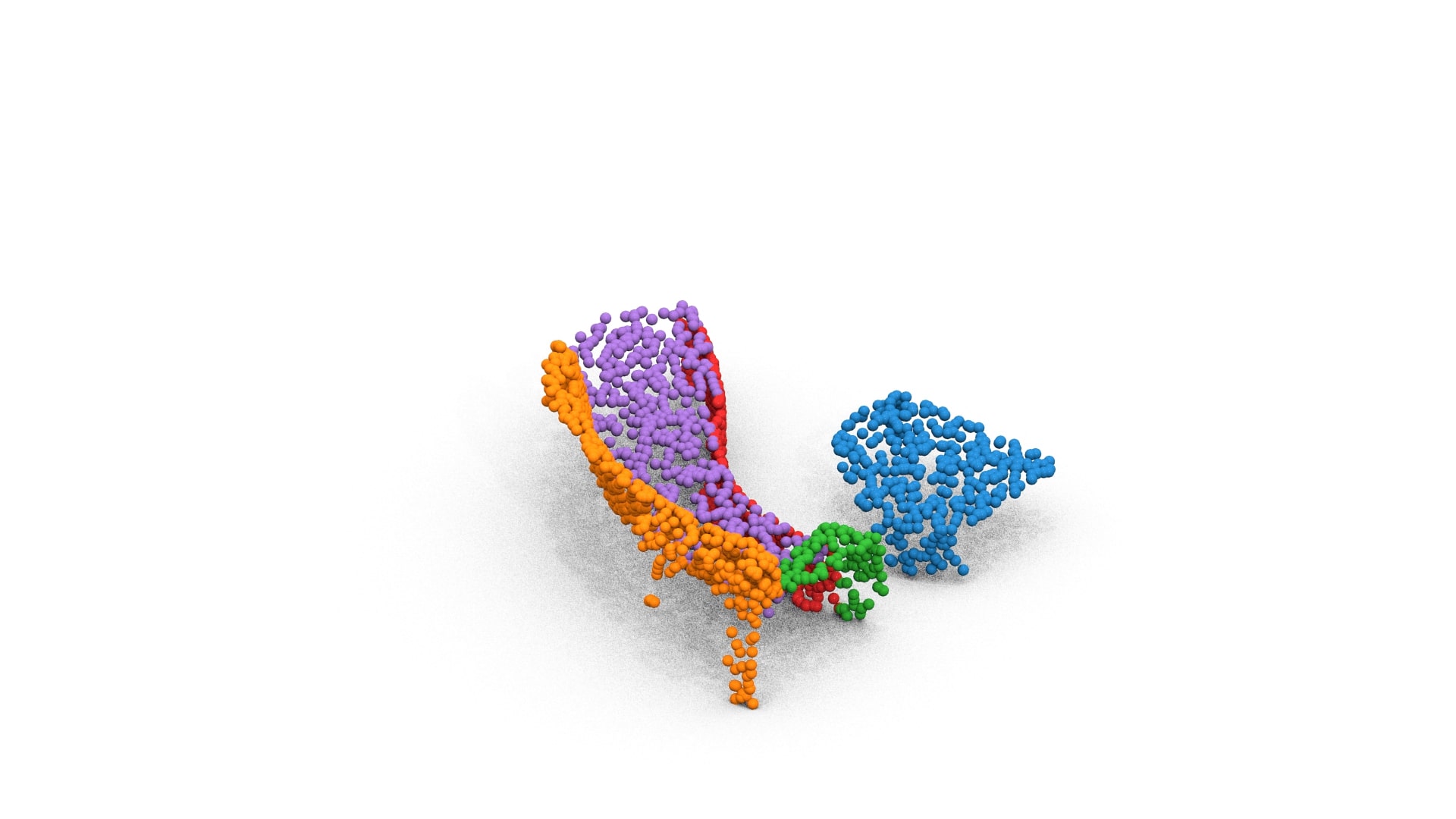}
    \end{overpic}
\end{minipage}
\begin{minipage}{0.14\textwidth}
    \centering
    \begin{overpic}[width=\textwidth,trim=710 230 720 300,clip]{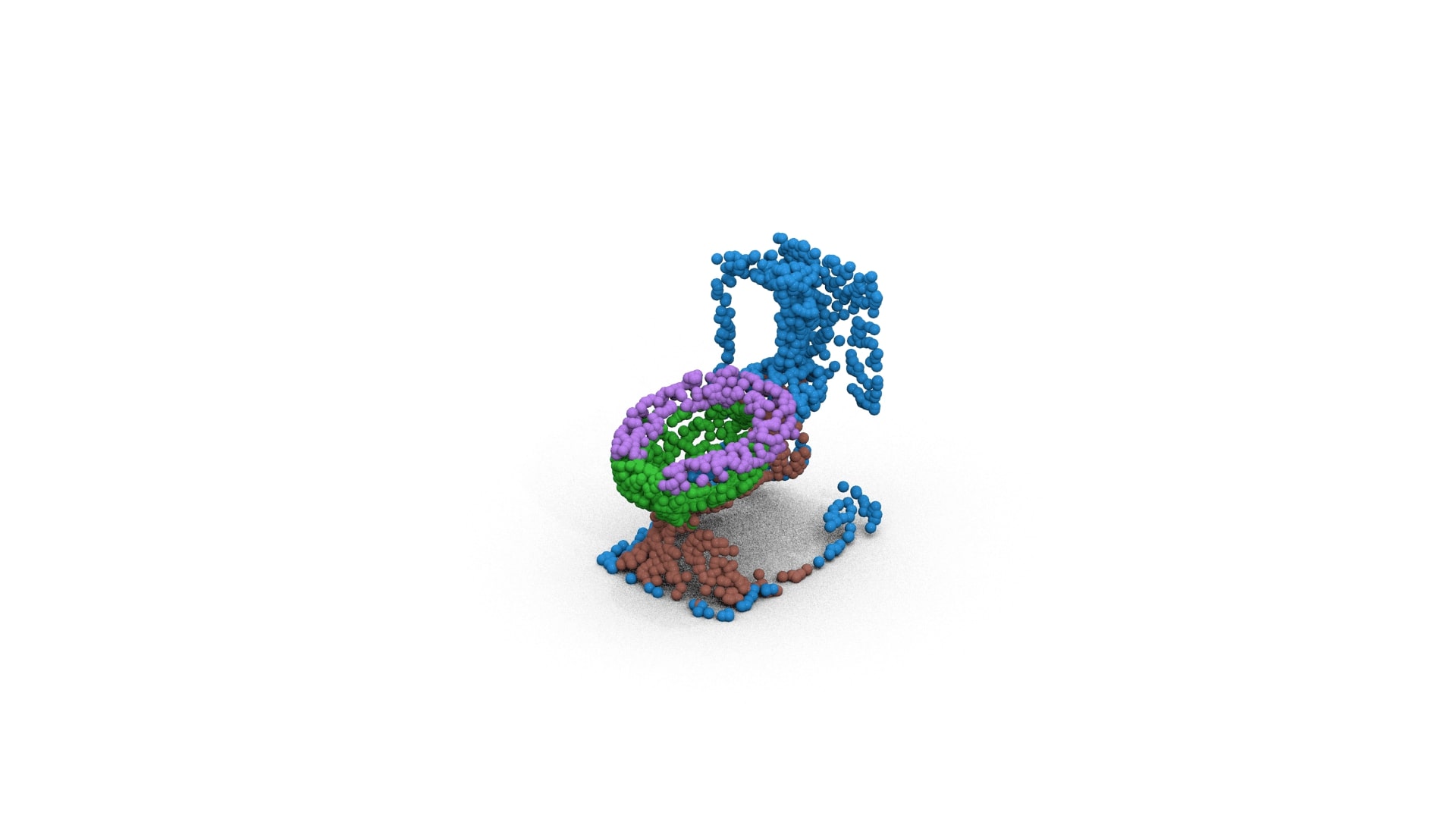}
    \end{overpic}
\end{minipage}
}

\vspace{-3mm}
\caption{
    Qualitative results on ScanObjectNN~\cite{uy2019scanobjectnn}. Top to bottom: input point cloud with color information; PointCLIPv2's prediction; \acronym's prediction; ground-truth segmentation.
    \acronym outputs better and sharper segmentations than PointCLIPv2.
}
\label{fig:qual_sonn}
\end{figure*}

\vspace{-8mm}
\cref{fig:qual} shows a qualitative comparison between our method and PointCLIPv2 on ShapeNetPart~\cite{yi2016shapenetpart} with seven shapes that cover various categories, \eg Airplane and Skateboard.
\cref{fig:qual_sonn} shows the comparison on five shapes from ScanObjectNN~\cite{uy2019scanobjectnn} \texttt{OBJ-BG} (the same as \cref{tab:zs_sonn}).
In both Figures, we observe the superior quality of segmentations produced by \acronym thanks to (i) the decoupling of part segmentation and semantic label assignment, and (ii) the GFA module that makes the VFM's features geometric-aware by introducing structural knowledge.
\vspace{10mm}

\subsection{Ablation study}%
\label{sec:ablation}

\begin{figure*}[t!]
\centering

\vspace{-1mm}
\begin{overpic}[width=0.95\textwidth,trim=7 0 7 0,clip]{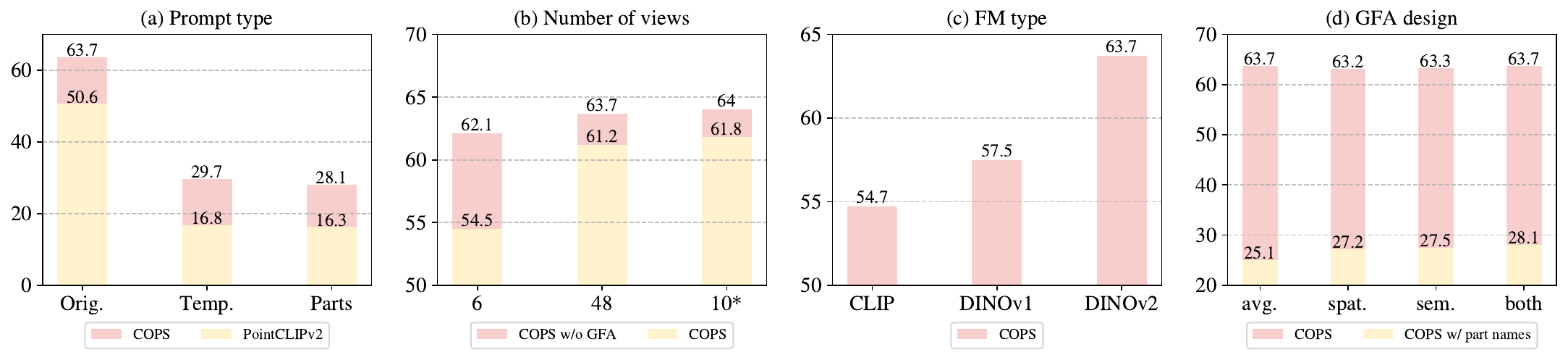}
\end{overpic}

\vspace{-3mm}
\caption{
    Ablation on ShapeNetPart~\cite{chang2015shapenet}.
    From left to right:
    (a) Different prompt types, comparing PointCLIPv2~\cite{zhu_pointclip_2023} and \acronym.
    (b) Varying the number of views during rendering, with and without our GFA module.
    (c) Changing the foundation model.
    (d) Ablating the GFA module.
}%
\label{fig:ablation}
\end{figure*}

In \cref{fig:ablation}, we conduct an ablation study of \acronym on ShapeNetPart~\cite{yi2016shapenetpart}.
We analyse:
(a) the influence of prompting techniques and
(b) rendered views on semantic label assignment,
(c) the effects of the type of VFM and
(d) the GFA module design on feature extraction.

\noindent\textbf{Prompt type.}
\cref{fig:ablation}(a) compares \acronym with PointCLIPv2 with three distinct prompting methodologies: (i) the prompts of PointCLIPv2~\cite{zhu_pointclip_2023} (`Orig.'), generated with GPT, (ii) template prompts containing part names (`Temp.') and (iii) utilising part names alone (`Parts').
\acronym always outperforms PointCLIPv2, with a margin of up to +13.1\%.
For `Orig.', PointCLIPv2 searches for the best-performing prompts for each part. Following this, our experiments suggest that these specifically chosen prompts lead to a performance boost compared to the other strategies.

\noindent\textbf{Number of rendered views.}
\cref{fig:ablation}(b) assesses the impact of the number of rendered views.
We test three different settings:
(i) 6 orthogonal views, following SimpleView~\cite{goyal2021revisiting};
(ii) 48 views, sampled around the object to ensure a good covering of all viewpoints;
(iii) 10 views, following PointCLIPv2~\cite{zhu_pointclip_2023}.
For each setting, we show the result of \acronym both with and without the GFA module, and we utilise the same parts' descriptions provided by PointCLIPv2.
We observe that increasing the number of views from 6 to 10 leads to a boost in performance of +7.3\% (+2.3\% with GFA), while further increasing it from 10 to 48 has a negative effect, with -0.6\% (-0.7\% with GFA).
We suggest that this may be due to a noise increment in the features extracted from \emph{unusual} views (\eg potentially out of distribution, or not capturing the object properly) when using the 48-view setting.
We also show the effectiveness of our full GFA module, which always increases performance and reduces the gap between different numbers of views.

\noindent\textbf{VFM type.}
\cref{fig:ablation}(c) investigates the contribution of the VFM $\Gamma_{\Theta}$ used in $\Phi_\mathcal{E}$.
Specifically, we compare the CLIP~\cite{radford2021learning} visual encoder and DINOv1~\cite{caron2021emerging} to our default choice, DINOv2~\cite{oquab2023dinov2}, using the same ViT-B~\cite{dosovitskiy2021image} backbone for all models to ensure a fair comparison.
We utilise the same prompts provided by PointCLIPv2 (`Orig.' from (a)) and the same 48 rendered views from (b).
When integrating CLIP within our pipeline, we observe an improvement of +4.1\% over PointCLIPv2, as evidenced by comparing against the first bin of (a).
This indicates the effectiveness of our two-step approach, which first decomposes a shape into parts before classifying them.
Next, DINOv1 further boosts performance by +2.8\%, suggesting that its self-supervised features are better suited for our task than CLIP's text-aligned visual features.
DINOv2 achieves the best results, with significant improvements of +6.2\% (\wrt DINOv1) and +9\%~(\wrt~CLIP).

\noindent\textbf{GFA design.}
\cref{fig:ablation}(d) analyses the contribution of our geometric feature aggregation (GFA) module in four distinct settings:
(i) `avg.', which simply performs average pooling;
(ii) `spat.', which performs spatially-consistent aggregation only;
(iii) `sem.', which performs semantically-consistent aggregation only;
(iv) `both', which is our final GFA design for \acronym, leveraging both spatially- and semantically-consistent aggregation.
For all settings, we measure performance both when using simple prompts (`Parts' in (a)), and when using PointCLIPv2's descriptions (`Orig.' in (a)). In both cases, we observe an increasing boost in performance due to GFA. Spatially-consistent aggregation introduces 3D structural knowledge: for instance, the points in an armrest \emph{exchange} information, thus increasing their feature similarity. Similarly, semantically-consistent aggregation makes features semantic-aware: for example, the two armrests of a chair will share information, getting closer to each other in feature space. 
Our final GFA design performs (ii) followed by (iii), improving performance of up to +3\%.

    \section{Conclusion}%
\label{sec:conclusion}
In this work, we presented \acronym, a zero-shot model for part segmentation that splits the task into part decomposition and semantic labelling. We introduced a geometric feature aggregation (GFA) module that fuses the features extracted by the vision foundation model on rendered views with geometric knowledge of the point cloud itself.
We conducted extensive evaluations on part segmentation datasets showing \acronym's strong performance in all these settings.

\noindent\textbf{Limitations.}
Thanks to the decoupling of part decomposition and semantic labelling, \acronym can segment parts regardless of natural language, as we show in ``$\uparrow$'' experiments in the tables.
However, as natural language is still needed when part semantic labels are required for the downstream task, low-quality prompts negatively affect part classification.

\noindent\textbf{Future work.}
\acronym may be extended to cover new tasks, such as indoor scene segmentation, where LiDAR or RGB-D acquisition sensors provide point clouds and images, thereby eliminating the need for rendering. Architecture-wise, a learning-based module may be introduced to dynamically weight the features across multiple views.

    \section*{Acknowledgements}
    This work was supported by the European Union's Horizon Europe research and innovation programme under grant agreement No.~101058589 (AI-PRISM).

    \clearpage
    {\small
    \bibliographystyle{wacv25toolkit/ieee_fullname}
    \bibliography{main}
    }

\end{document}


    \newcommand{\cmark}{\ding{51}}
\newcommand{\xmark}{\ding{55}}
\newcommand{\warning}[1]{\textbf{\color{red!90}{#1}}}
\newcommand{\myparagraph}[1]{\vspace{5pt}\noindent\textbf{{#1}}}

\definecolor{myblue}{rgb}{0.0000,0.7490,1.0000}
\definecolor{myyellow}{rgb}{1.0000,0.7530,0.0000}
\definecolor{mygray}{rgb}{0.9000,0.9000,0.9000}
\definecolor{myazure}{rgb}{0.8509,0.8980,0.9412}
\definecolor{mygreen}{rgb}{0.0,0.6,0.0}
\definecolor{mylightyellow}{rgb}{1.0000,0.9294,0.6902}

\definecolor{part_blue}{rgb}{0.12156862745098039, 0.4666666666666667, 0.7058823529411765}
\definecolor{part_orange}{rgb}{1.0, 0.4980392156862745, 0.054901960784313725}
\definecolor{part_green}{rgb}{0.17254901960784313, 0.6274509803921569, 0.17254901960784313}
\definecolor{part_red}{rgb}{0.8392156862745098, 0.15294117647058825, 0.1568627450980392}

\newcommand{\acronym}{COPS\xspace}
\newcommand{\spacing}{\vspace{0.8pt}}

    \def\eg{\emph{e.g}\onedot,\xspace} \def\Eg{\emph{E.g}\onedot,\xspace}
    \def\ie{\emph{i.e}\onedot,\xspace} \def\Ie{\emph{I.e}\onedot,\xspace}

    \title{Supplementary material for\\
    3D Part Segmentation via Geometric Aggregation of 2D Visual Features}

    \author{
        \begin{minipage}[t]{0.33\textwidth}
        \centering
        Marco Garosi\\
        University of Trento\\
        {\tt\small marco.garosi@unitn.it}
        \end{minipage}
        \begin{minipage}[t]{0.33\textwidth}
        \centering
        Riccardo Tedoldi\\
        University of Trento\\
        {\tt\small riccardo.tedoldi@unitn.it}
        \end{minipage}
        \begin{minipage}[t]{0.33\textwidth}
        \centering
        Davide Boscaini\\
        Fondazione Bruno Kessler\\
        {\tt\small dboscaini@fbk.eu}
        \end{minipage}
        \and
        \begin{minipage}[t]{0.33\textwidth}
        \centering
        Massimiliano Mancini\\
        University of Trento\\
        {\tt\small massimiliano.mancini@unitn.it}
        \end{minipage}
        \begin{minipage}[t]{0.33\textwidth}
        \centering
        Nicu Sebe\\
        University of Trento\\
        {\tt\small sebe@disi.unitn.it}
        \end{minipage}
        \begin{minipage}[t]{0.33\textwidth}
        \centering
        Fabio Poiesi\\
        Fondazione Bruno Kessler\\
        {\tt\small poiesi@fbk.eu}
        \end{minipage}
    }

    \maketitle
    \appendix

    \section{Introduction}\label{sec:intro}

We provide additional material in support of our main paper.
This document is organised as follows:
\begin{itemize}[noitemsep,nolistsep,leftmargin=*]
    \item In \cref{sec:pipeline_viz}, we describe the steps involved in \acronym and we show them qualitatively on some point clouds.
    \item In \cref{sec:implem}, we provide implementation details of \acronym and specify the hyper-parameter used in our experiments to facilitate reproducibility.
    \item In \cref{sec:abl_layers}, we ablate the role of the specific layer of the DINOv2-Base architecture from which we perform feature extraction.
    \item In \cref{sec:abl_gfa}, we investigate the role of the number of spatial and semantic nearest neighbours used in the Geometric Feature Aggregation (GFA) module.
    \item In \cref{sec:qual}, we provide additional qualitative results on ScanObjectNN~\cite{uy2019scanobjectnn} and FAUST~\cite{Bogo:CVPR:2014} datasets.
    \item In \cref{sec:hardware}, we provide details on the computational resources utilised.
\end{itemize}

    \section{Pipeline visualisation}%
\label{sec:pipeline_viz}

\begin{figure*}[!htb]
\centering

\begin{overpic}[width=1.0\textwidth]{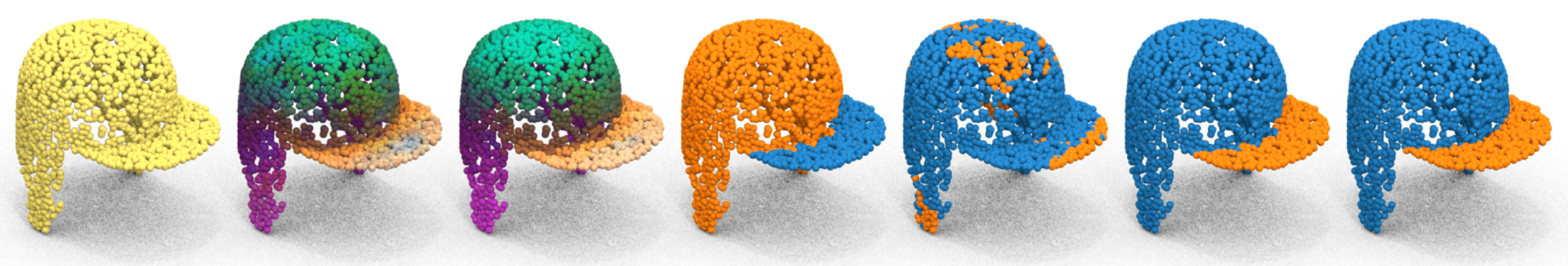}
\end{overpic}
\begin{overpic}[width=1.0\textwidth]{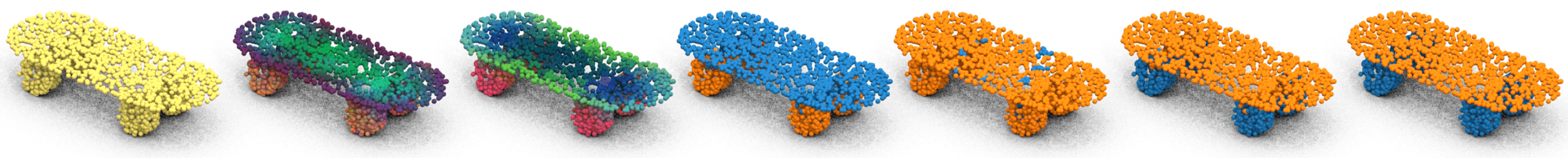}
\end{overpic}
\begin{overpic}[width=1.0\textwidth]{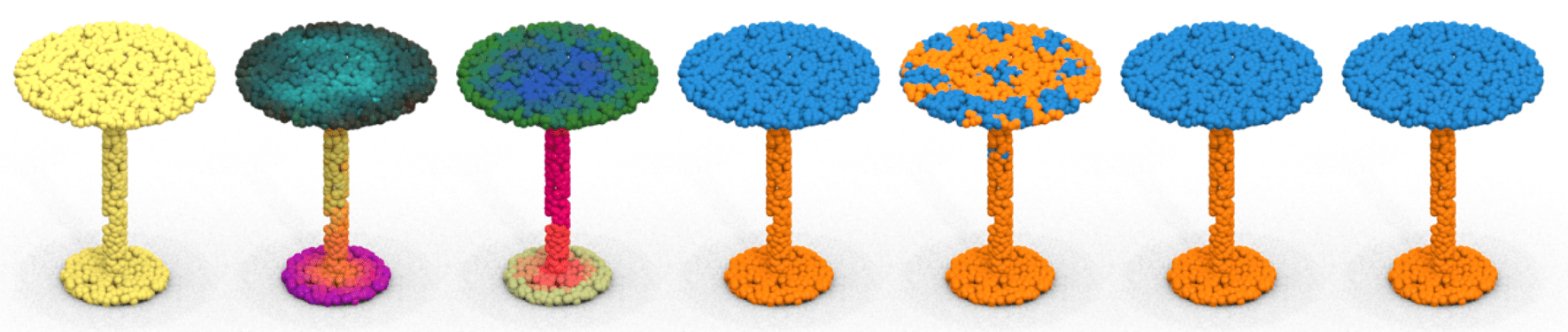}
\end{overpic}
\begin{overpic}[width=1.0\textwidth]{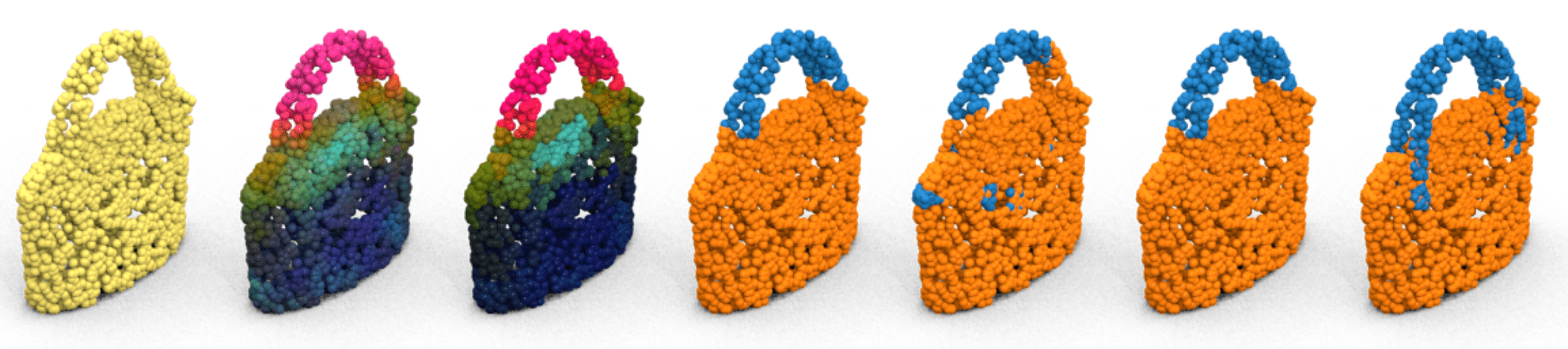}
    \put(4,-1){\footnotesize Input}
    \put(14,-1){\footnotesize Before GFA}
    \put(29,-1){\footnotesize After GFA}
    \put(45,-1){\footnotesize Clusters}
    \put(56,-1){\footnotesize PointCLIPv2}
    \put(75,-1){\footnotesize \acronym}
    \put(91,-1){\footnotesize GT}
\end{overpic}

\vspace{1.7mm}
\hspace{2mm}
\begin{overpic}[width=0.9\textwidth]{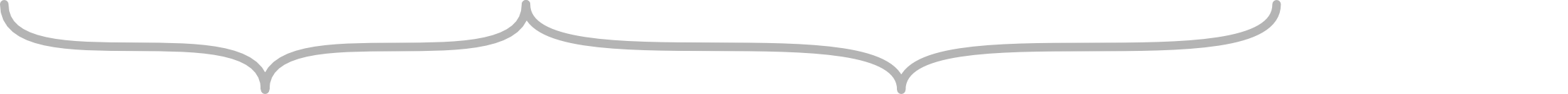}
    \put(6,-2){\footnotesize Feature extractor $\Phi$}
    \put(51,-2){\footnotesize Segmenter $\Psi$}
\end{overpic}

\vspace{1mm}
\caption{
    Detailed visualisation of the steps required by \acronym.
    From left to right:
    input point cloud,
    intermediate features obtained by 3D-lifting DINOv2 features,
    final features obtained with GFA,
    part decomposition obtained via feature clustering (colours are not informative because cluster labels are not semantic),
    PointCLIPv2 predictions,
    \acronym predictions,
    and ground-truth segmentation.
    By disentangling part decomposition (fourth column) from semantic label assignment, \acronym can leverage noisy PointCLIPv2 predictions (fifth column) to produce accurate segmentations (sixth column).
}%
\label{fig:steps}
\end{figure*}

In \cref{fig:steps}, we provide a detailed visualisation of the different steps required by \acronym on four point clouds from ShapeNetPart~\cite{yi2016shapenetpart}.
The first three columns illustrate the feature extractor $\Phi$, which processes the input point cloud depicted in the first column, extracts features using DINOv2~\cite{oquab2023dinov2} (second column), and modifies them using the Geometric Feature Aggregation (GFA) module (third column).
The next three columns illustrate the segmenter $\Psi$, which decomposes the object into parts via feature clustering (fourth column, colours are not informative) and assigns each part a semantic label (sixth column) by leveraging PointCLIPv2~\cite{zhu_pointclip_2023} predictions (fifth column) as semantic anchors.
The last column displays the ground-truth segmentation.
The minimal difference between the last two columns suggests that \acronym produces very accurate segmentations, despite the low quality of the PointCLIPv2 predictions shown in the fifth column.

    \section{Implementation details}%
\label{sec:implem}

In this section, we discuss implementation details and the hyper-parameters of \acronym.

\noindent\textbf{Point cloud processing.}
We perform rendering at the original point cloud resolution to retain finer details.
Then, we randomly sample 10,000 points from each object, we update the pixel-to-point mappings utilised to back-project features to 3D, and we project them back.
Next, GFA performs farthest point sampling (FPS) to find super points for feature aggregation.
Subsequently, we randomly sample 2,048 points to obtain semantic labels via PointCLIPv2.
Lastly, we perform clustering on these sampled points and we assign each cluster a semantic label via Hungarian with PointCLIPv2's predictions.

\begin{figure*}[t!]
\centering

\begin{minipage}{0.32\textwidth}
    \begin{overpic}[width=\textwidth,trim=10 130 10 150,clip]{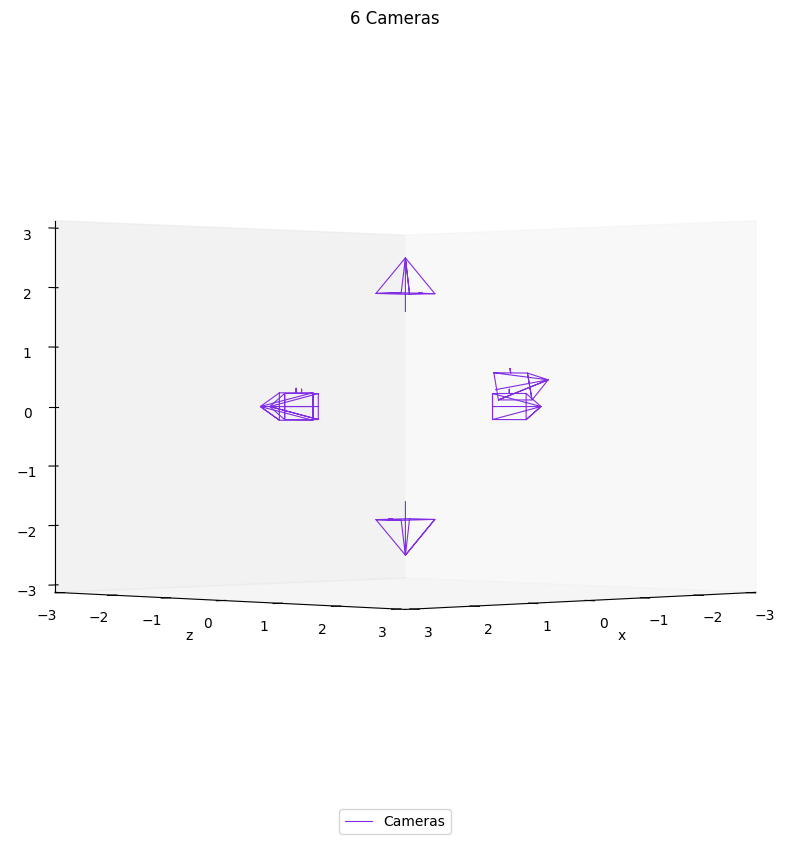}
    \put(35,-5){\scriptsize (a) 6 views}
    \end{overpic}
\end{minipage}
\begin{minipage}{0.32\textwidth}
    \begin{overpic}[width=\textwidth,trim=10 130 10 150,clip]{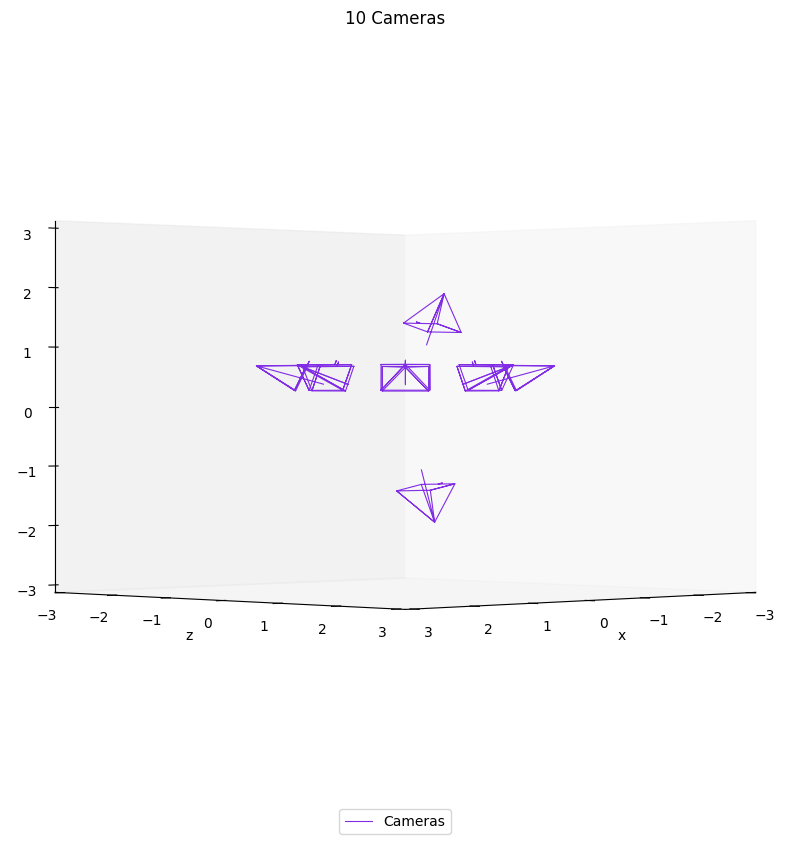}
    \put(32,-5){\scriptsize (b) 10 views}
    \end{overpic}
\end{minipage}
\begin{minipage}{0.32\textwidth}
    \begin{overpic}[width=\textwidth,trim=10 130 10 150,clip]{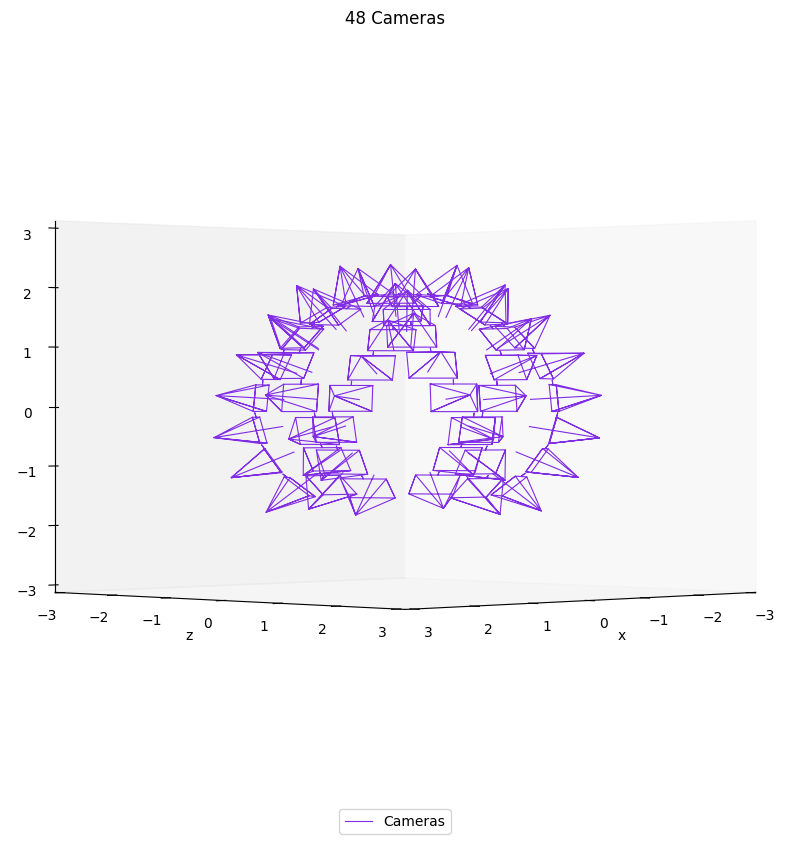}
    \put(32,-5){\scriptsize (c) 48 views}
    \end{overpic}
\end{minipage}

\caption{
    Cameras visualised in 3D space.
    (a) shows the 6-camera setting, where the cameras are orthogonal to one another.
    (b) shows the 10 cameras adapted from PointCLIPv2~\cite{zhu_pointclip_2023}.
    They capture the object from more sides, while the top and bottom views are similar to those in (a).
    (c) shows the largest setting, with 48 cameras sampled bottom to top on eight arches around the object.
    While this configuration provides more views than (a) and (b), there are many redundant views which contribute little to the overall performance, as we have shown in the corresponding ablation study.
}%
\label{fig:viewpoints}
\end{figure*}

\noindent\textbf{Rendering.}
We utilise PyTorch3D~\cite{ravi2020pytorch3d} for rendering. Notably, we set: (i) camera orientations, (ii) point size, and (iii) rendering canvas size.
We have defined three camera settings: 6 orthogonal cameras, facing front, back, left, right, top, and bottom in \cref{fig:viewpoints}(a); 10 cameras, following PointCLIPv2~\cite{zhu_pointclip_2023} in \cref{fig:viewpoints}(b); 48 cameras in \cref{fig:viewpoints}(c).
We set the point size to small values for datasets whose point clouds are \textit{dense}, \ie containing many points. We enlarge the point size for \textit{sparse} datasets, such as ShapeNetPart, to obtain smooth renders.
Lastly, we set the canvas size to the input size of DINOv2 of $224 \times 224$ pixels, ensuring no scaling and/or cropping is required.
When photometric (RGB) information is not available, we render depth maps.
We utilise both the depth maps produced by PyTorch3D, where light pixels correspond to close points, and the depth maps in PointCLIPv2's style, where the dark pixels correspond to points close to the camera.
We found \acronym to perform the best with PyTorch3D's depth maps.

\noindent\textbf{Feature extraction.}
DINOv2 is based on the vision transformer architecture~\cite{vaswani2023attention, dosovitskiy2021image}. It splits the image into patches of $14 \times 14$ pixels and outputs a feature vector for each patch.
However, we need pixel-level, or dense, features to perform lifting to 3D. 
Therefore, we upsample the feature maps to the input image size of $224 \times 224$ pixels using bicubic interpolation.

\noindent\textbf{Geometric feature aggregation (GFA).}
GFA works in three steps:
(i) it samples super points,
(ii) it aggregates features via either spatial or semantic attention, and
(iii) it upsamples super point features to the whole point cloud.
GFA has two hyper-parameters: the number of super points and the number of neighbours considered in the aggregation step. For the first, too many super points may lead to reduced spatial/semantic consistency, while too few can make the features collapse, not accounting for the local variability of the point cloud.
By default, we sample 256 super points.
For the second, the more the points and the larger the context window used to compute the super point's feature.
By default, we set it to 10 for spatial attention and to 90 for semantic attention.
In \cref{sec:abl_gfa}, we conduct an ablation study on these hyper-parameters.

    \section{Ablation study on DINOv2-Base layers}%
\label{sec:abl_layers}

Following FoundPose~\cite{ornek2023foundpose}, in \cref{tab:abl_layer} we evaluate \acronym with different patch descriptors. We utilise DINOv2-Base (ViT-B) and sample patch-level features at different levels, showing how performance changes.
We observe an increment in performance as we utilise increasingly higher-level patch descriptors, which encode more abstract semantic information, \eg about parts~\cite{oquab2023dinov2}.
We note that performance improves in certain categories, such as ``airplane'', as we move towards higher-level representations. However, we find that for other categories, such as ``mug'' or ``knife'', lower-level representations may provide the optimal level of abstraction for achieving accurate part segmentation results.
This highlights the need for further exploration in future work, assessing how to consider representations at different scales to achieve consistent improvements.

\begin{table*}[t]
\centering
\tabcolsep 3pt

\vspace{-3mm}
\resizebox{2\columnwidth}{!}{%
\begin{tabular}{rl|cc|cccccccccccccccc}
    \toprule
    & ViT-B layer & \rotatebox{90}{$\textrm{mIoU}_\textrm{I}$} & \rotatebox{90}{$\textrm{mIoU}_\textrm{C}$} & \rotatebox{90}{Airplane} & \rotatebox{90}{Bag} & \rotatebox{90}{Cap} & \rotatebox{90}{Car} & \rotatebox{90}{Chair} & \rotatebox{90}{Earphone} & \rotatebox{90}{Guitar} & \rotatebox{90}{Knife} & \rotatebox{90}{Lamp} & \rotatebox{90}{Laptop} & \rotatebox{90}{Motorbike} & \rotatebox{90}{Mug} & \rotatebox{90}{Pistol} & \rotatebox{90}{Rocket} & \rotatebox{90}{Skate} & \rotatebox{90}{Table} \\
    \toprule
    \color{gray} 1 & 0 & 49.7 & 46.4 & 31.1 & 53.0 & 45.7 & 27.9 & 47.5 & 50.8 & 60.7 & 66.6 & 45.5 & 71.3 & 21.6 & 42.9 & 35.6 & 31.0 & 48.8 & 61.8\\
    \color{gray} 2 & 1 & 55.9 & 52.1 & 31.5 & 72.9 & 65.2 & 29.9 & 60.6 & 56.0 & 47.7 & 62.9 & 45.9 & 90.0 & 22.6 & 58.4 & 39.8 & 29.5 & 50.0 & 70.3 \\
    \color{gray} 3 & 2 & 58.4 & 51.6 & 32.3 & 65.2 & 49.2 & 32.2 & 66.0 & 55.7 & 50.9 & 68.9 & 46.8 & 90.6 & 23.9 & 43.9 & 42.7 & 31.4 & 52.5 & 72.7 \\
    \color{gray} 4 & 3 & 58.8 & 52.5 & 33.1 & 64.4 & 54.5 & 32.9 & 66.4 & 55.3 & 48.7 & 72.6 & 47.2 & 92.0 & 23.7 & 53.4 & 40.1 & 31.2 & 52.1 & 72.7\\
    \color{gray} 5 & 6 & 60.5 & 55.2 & 36.8 & 53.8 & 65.4 & 32.2 & 68.0 & 60.8 & 66.6 & 78.5 & 48.4 & 91.1 & 24.6 & 36.5 & 48.9 & 45.7 & 53.3 & 71.8\\
    \color{gray} 6 & 7 & 61.0 & 56.4 & 41.1 & 59.9 & 64.4 & 31.8 & 68.2 & 63.5 & 67.2 & 80.1 & 48.4 & 88.9 & 24.8 & 41.1 & 49.6 & 47.0 & 55.6 & 71.0 \\
    \color{gray} 7 & 8 & 61.4 & 56.3 & 43.2 & 60.7 & 61.6 & 31.8 & 68.4 & 64.7 & 67.1 & 77.5 & 48.2 & 90.1 & 25.1 & 40.3 & 49.8 & 45.8 & 55.1 & 71.7\\
    \color{gray} 8 & 10 & 62.6 & 59.0 & 50.7 & 64.7 & 71.1 & 30.5 & 69.0 & 66.2 & 70.2 & 81.6 & 48.7 & 77.9 & 25.3 & 60.1 & 52.2 & 47.6 & 56.3 & 71.2 \\
    \color{gray} 9 & 11 & 63.3 & 60.1 & 50.6 & 70.7 & 69.8 & 29.7 & 70.0 & 66.0 & 66.2 & 79.7 & 49.0 & 86.8 & 25.9 & 68.2 & 53.5 & 46.0 & 57.5 & 72.1 \\
    \color{gray} 10 & 12 (w/o norm.) & 64.0 & 61.1 & 50.9 & 75.8 & 69.5 & 29.7 & 71.2 & 66.4 & 68.8 & 76.5 & 48.8 & 89.2 & 25.1 & 72.4 & 56.4 & 47.6 & 56.8 & 72.8 \\
    \rowcolor{myazure} \color{gray} 11 & 12 (w/ norm.) & 64.4 & 60.9 & 51.3 & 71.0 & 69.7 & 29.9 & 71.7 & 66.4 & 72.6 & 80.2 & 48.6 & 91.1 & 26.1 & 62.5 & 54.3 & 46.7 & 59.4 & 72.9 \\
    \bottomrule
\end{tabular}
}

\caption{
    Part segmentation performance on ShapeNetPart~\cite{yi2016shapenetpart} obtained when extracting DINOv2 base features from different layers, \ie at different depths.
    DINOv2 base is based on ViT-B, and has 13 layers.
    The \colorbox{myazure}{highlighted} row corresponds to the results shown in the main paper.
}%
\label{tab:abl_layer}

\end{table*}

    \section{Ablation study on GFA}%
\label{sec:abl_gfa}

In \cref{tab:abl_gfa}(a), we conduct an additional ablation study on the GFA module.
We evaluate 13 distinct configurations, including the standard GFA employed in the main paper.
Specifically, we vary the number of sampled super points, the number of neighbours taken into account during the attention operation, \ie the ``context window'', and weighting by distance.
In \cref{tab:abl_gfa}(b) we repeat these experiments by swapping the order of spatially- and semantically-consistent feature aggregation.
The results show that our default configuration performs the best.
We observe that increasing the number of neighbours reduces performance, thus suggesting that a smaller ``context window'' allows GFA to aggregate only the most relevant features.
Raising the number of super points leads to a decrease in performance because it limits the effect of GFA.
If all the points are kept as super points, GFA has no effect, while if too few super points are sampled, features can collapse.
Lastly, performing spatially- before semantically-consistent feature aggregation makes GFA capture geometric knowledge better, thus achieving higher performance.

\begin{table*}[t]
\centering
\tabcolsep 3pt

\resizebox{2\columnwidth}{!}{%
    \subfloat[Spatially-consistent aggregation followed by semantically-consistent aggregation]{
    \begin{tabular}{crcc|cc|cc|cccccccccccccccc}
        \toprule
        & & \rotatebox{90}{Sup. spat.} & \rotatebox{90}{Sup. sem.} & \rotatebox{90}{Nei. spat.} & \rotatebox{90}{Nei. sem.} & \rotatebox{90}{$\textrm{mIoU}_\textrm{I}$} & \rotatebox{90}{$\textrm{mIoU}_\textrm{C}$} & \rotatebox{90}{Airplane} & \rotatebox{90}{Bag} & \rotatebox{90}{Cap} & \rotatebox{90}{Car} & \rotatebox{90}{Chair} & \rotatebox{90}{Earphone} & \rotatebox{90}{Guitar} & \rotatebox{90}{Knife} & \rotatebox{90}{Lamp} & \rotatebox{90}{Laptop} & \rotatebox{90}{Motorbike} & \rotatebox{90}{Mug} & \rotatebox{90}{Pistol} & \rotatebox{90}{Rocket} & \rotatebox{90}{Skate} & \rotatebox{90}{Table} \\
        \toprule
        \color{gray} \multirow{6}{*}{\rotatebox{90}{Superpoints}} & \color{gray} 1 & 512 & 256 & \multirow{6}{*}{\rotatebox{0}{10}} & \multirow{6}{*}{\rotatebox{0}{90}} & 64.2 & 60.0 & 50.6 & 68.3 & 69.2 & 30.1 & 71.6 & 65.4 & 74.2 & 78.6 & 48.9 & 90.9 & 26.3 & 58.1 & 52.9 & 45.5 & 57.3 & 72.7 \\
        &\color{gray} 2 & 256 & 512 & & & 64.3 & 60.7 & 51.0 & 70.6 & 70.0 & 29.7 & 71.5 & 64.8 & 72.0 & 79.0 & 48.9 & 91.6 & 26.1 & 63.3 & 53.8 & 46.4 & 59.5 & 73.0 \\
        & \color{gray} 3 & 512 & 512 & & & 64.3 & 60.8 & 50.9 & 71.4 & 70.1 & 29.7 & 71.2 & 65.8 & 73.7 & 78.6 & 48.5 & 91.2 & 26.2 & 62.9 & 52.3 & 48.0 & 59.6 & 73.1 \\
        & \color{gray} 4 & 128 & 256 & & & 64.3 & 61.0 & 51.0 & 68.9 & 72.1 & 30.0 & 71.7 & 66.0 & 70.5 & 79.0 & 48.9 & 91.4 & 25.5 & 64.3 & 55.9 & 49.0 & 59.4 & 72.8 \\
        & \color{gray} 5 & 256 & 128 & & & 64.2 & 60.3 & 51.4 & 66.0 & 72.0 & 29.9 & 71.7 & 65.2 & 70.9 & 78.9 & 48.9 & 91.0 & 25.8 & 62.9 & 53.1 & 48.1 & 57.2 & 72.6 \\
        & \color{gray} 6 & 128 & 128 & & & 64.3 & 60.5 & 50.7 & 73.3 & 67.0 & 29.6 & 71.5 & 66.5 & 72.1 & 80.3 & 49.1 & 91.0 & 25.8 & 55.7 & 56.2 & 47.7 & 59.4 & 73.0 \\
        \midrule
        \rowcolor{myazure} \color{gray} \multirow{4}{*}{\rotatebox{90}{Neighbours}} & \color{gray} 7 & \multirow{4}{*}{\rotatebox{0}{256}} & \multirow{4}{*}{\rotatebox{0}{256}} & 10 & 90 & 64.4 & 60.9 & 51.3 & 71.0 & 69.7 & 29.9 & 71.7 & 66.4 & 72.6 & 80.2 & 48.6 & 91.1 & 26.1 & 62.5 & 54.3 & 46.7 & 59.4 & 72.9 \\
        & \color{gray} 8 & & & 90 & 90 & 61.9 & 58.6 & 49.6 & 62.8 & 68.9 & 28.0 & 67.8 & 64.0 & 73.3 & 80.3 & 48.5 & 89.0 & 24.9 & 58.9 & 54.9 & 42.1 & 55.0 & 69.5 \\
        & \color{gray} 9 & & & 170 & 256 & 58.6 & 54.9 & 49.0 & 59.5 & 69.3 & 25.3 & 62.8 & 60.4 & 70.5 & 79.5 & 48.3 & 86.6 & 23.9 & 32.1 & 53.5 & 46.4 & 46.3 & 65.6\\
        & \color{gray} 10 & & & 90 & 10 & 61.2 & 58.5 & 49.1 & 68.0 & 66.7 & 27.3 & 68.2 & 64.7 & 69.6 & 79.3 & 48.4 & 90.7 & 24.6 & 54.5 & 53.8 & 46.5 & 55.0 & 69.2 \\
        \bottomrule
    \end{tabular}
    }
}

\bigskip

\resizebox{2\columnwidth}{!}{%
    \subfloat[Semantically-consistent aggregation followed by spatially-consistent aggregation]{
    \begin{tabular}{crcc|cc|cc|cccccccccccccccc}
        \toprule
        & & \rotatebox{90}{Sup. sem.} & \rotatebox{90}{Sup. spat.} & \rotatebox{90}{Nei. sem.} & \rotatebox{90}{Nei. spat.}  & \rotatebox{90}{$\textrm{mIoU}_\textrm{I}$} & \rotatebox{90}{$\textrm{mIoU}_\textrm{C}$} & \rotatebox{90}{Airplane} & \rotatebox{90}{Bag} & \rotatebox{90}{Cap} & \rotatebox{90}{Car} & \rotatebox{90}{Chair} & \rotatebox{90}{Earphone} & \rotatebox{90}{Guitar} & \rotatebox{90}{Knife} & \rotatebox{90}{Lamp} & \rotatebox{90}{Laptop} & \rotatebox{90}{Motorbike} & \rotatebox{90}{Mug} & \rotatebox{90}{Pistol} & \rotatebox{90}{Rocket} & \rotatebox{90}{Skate} & \rotatebox{90}{Table} \\
        \toprule
        \color{gray} \multirow{6}{*}{\rotatebox{90}{Superpoints}} & \color{gray} 1 & 256 & 512 & \multirow{6}{*}{\rotatebox{0}{90}} & \multirow{6}{*}{\rotatebox{0}{10}} &  63.9 & 60.3 & 50.6 & 74.1 & 68.3 & 29.2 & 71.3 & 65.5 & 71.4 & 78.4 & 48.5 & 90.9 & 26.3 & 61.0 & 51.2 & 46.4 & 58.9 & 72.6 \\
        &\color{gray} 2 & 512 & 256 & & & 64.2 & 60.3 & 50.8 & 70.5 & 68.2 & 29.9 & 71.1 & 65.6 & 74.1 & 80.1 & 48.7 & 91.3 & 26.4 & 54.9 & 53.6 & 48.6 & 58.6 & 72.9 \\
        & \color{gray} 3 & 512 & 512 &  & & 63.9 & 60.1 & 50.2 & 71.5 & 71.0 & 29.2 & 71.0 & 65.6 & 72.0 & 79.2 & 48.6 & 91.0 & 26.4 & 55.1 & 53.7 & 46.7 & 57.9 & 73.0 \\
        & \color{gray} 4 & 256 & 128 &  & & 64.2 & 60.4 & 50.6 & 69.7 & 66.7 & 30.1 & 71.6 & 65.5 & 72.5 & 80.4 & 48.6 & 91.1 & 25.8 & 63.3 & 54.9 & 46.4 & 56.1 & 72.6 \\
        & \color{gray} 5 & 128 & 256 &  & & 64.1 & 60.3 & 50.7 & 71.4 & 68.1 & 29.6 & 71.3 & 65.3 & 72.1 & 79.7 & 48.6 & 91.0 & 26.2 & 59.3 & 53.9 & 46.7 & 58.6 & 72.7 \\
        & \color{gray} 6 & 128 & 128 &  & & 64.0 & 59.9 & 50.2 & 67.5 & 65.8 & 29.6 & 71.7 & 65.2 & 71.8 & 80.4 & 48.9 & 90.8 & 26.8 & 60.1 & 53.5 & 44.7 & 58.7 & 72.5 \\
        \midrule
        \color{gray} \multirow{4}{*}{\rotatebox{90}{Neighbours}} & \color{gray} 7 & \multirow{4}{*}{\rotatebox{0}{256}} & \multirow{4}{*}{\rotatebox{0}{256}} & 90 & 10  & 64.1 & 60.4 & 50.7 & 64.7 & 68.2 & 30.0 & 71.4 & 65.2 & 72.2 & 79.2 & 48.4 & 91.1 & 26.0 & 67.0 & 54.0 & 47.8 & 58.3 & 72.7 \\
        & \color{gray} 8 & & & 90 & 90 &  61.8 & 58.8 & 49.3 & 69.7 & 69.2 & 27.4 & 67.9 & 63.9 & 74.4 & 79.7 & 48.5 & 89.9 & 25.0 & 52.6 & 52.2 & 47.9 & 53.7 & 69.4 \\
        & \color{gray} 9 & & & 256 & 170 &  59.0 & 55.3 & 47.6 & 59.8 & 67.3 & 25.1 & 64.8 & 60.9 & 69.2 & 80.3 & 48.2 & 89.2 & 24.0 & 36.5 & 51.4 & 46.7 & 48.8 & 65.7 \\
        & \color{gray} 10 & & & 10 & 90  & 61.6 & 58.1 & 49.3 & 66.4 & 66.9 & 27.1 & 68.3 & 64.1 & 72.1 & 79.7 & 48.3 & 90.5 & 26.3 & 46.5 & 53.1 & 48.7 & 54.2 & 68.9 \\
        \bottomrule
    \end{tabular}
    }
}

\caption{
    Ablation on the geometric feature extractor (GFA) module on ShapeNetPart~\cite{yi2016shapenetpart}.
    Results were obtained using 10 rendered (depth only) views.
    All experiments use a two-stage GFA. Panels (a) and (b) show how performance changes when swapping the two stages.
    Each of the first two groups of columns reports the configuration for the two stages.
    ``Sup.'' is the number of super points sampled from the input point cloud.
    ``Nei.'' is the number of neighbours considered during the aggregation phase.
    The \colorbox{myazure}{highlighted} row corresponds to the results shown in the main paper. 
}%
\label{tab:abl_gfa}

\end{table*}

    \section{Additional qualitative results}%
\label{sec:qual}

\begin{figure*}[!htb]
\centering

\begin{overpic}[width=1.0\textwidth]{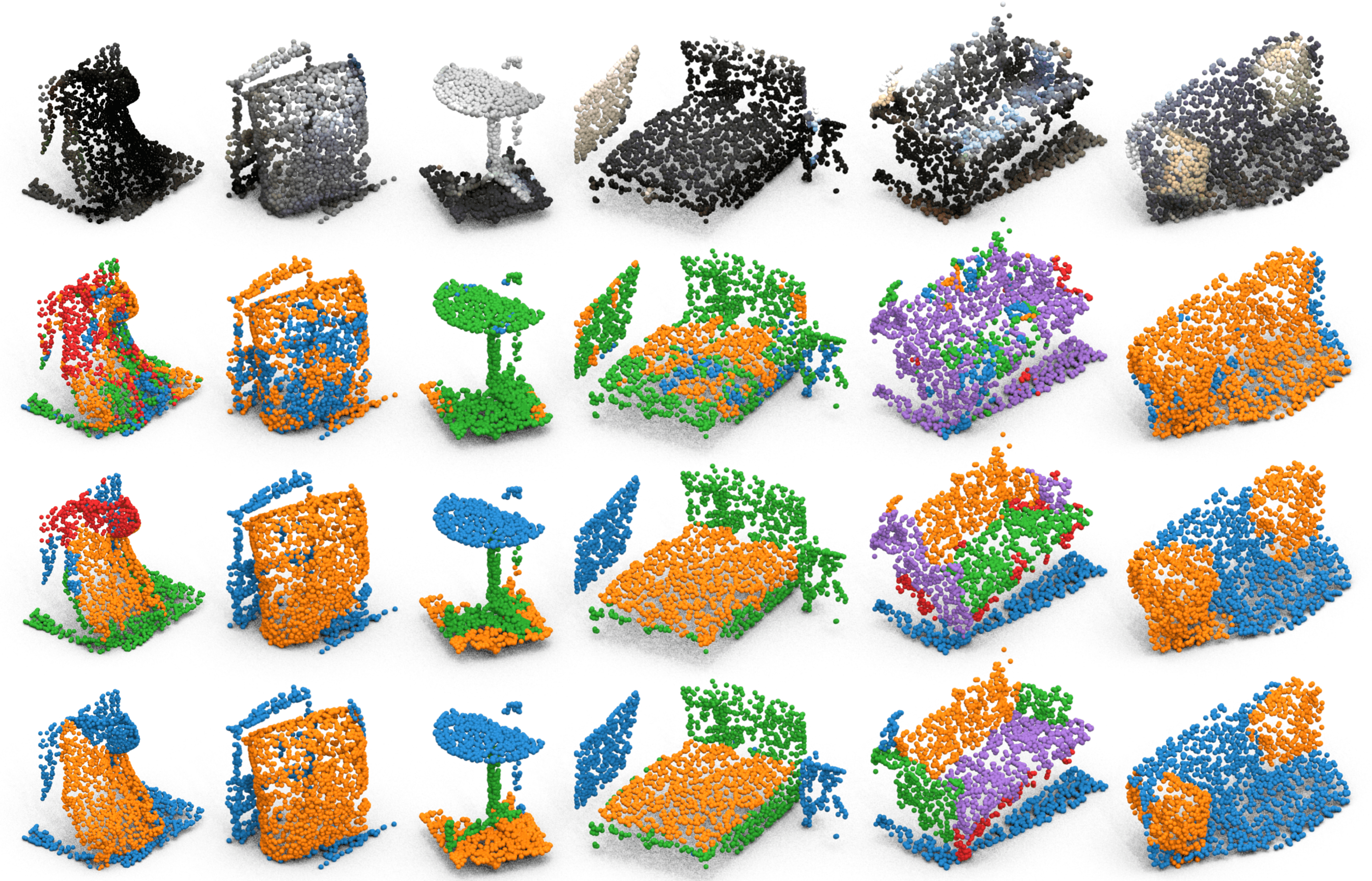}
    \put(-1,5){\rotatebox{90}{\footnotesize GT}}
    \put(-1,18.5){\rotatebox{90}{\footnotesize \acronym}}
    \put(-1,31){\rotatebox{90}{\footnotesize PointCLIPv2}}
    \put(-1,52){\rotatebox{90}{\footnotesize Input}}
\end{overpic}

\vspace{-1mm}
\caption{
Qualitative results on ScanObjectNN~\cite{uy2019scanobjectnn}.
Top to bottom: input point cloud with RGB colours, PointCLIPv2 predictions, \acronym predictions, and ground-truth segmentation.
}%
\label{fig:qual_sonn}
\end{figure*}

\begin{figure*}[!htb]
\centering

\begin{overpic}[width=0.8\textwidth]{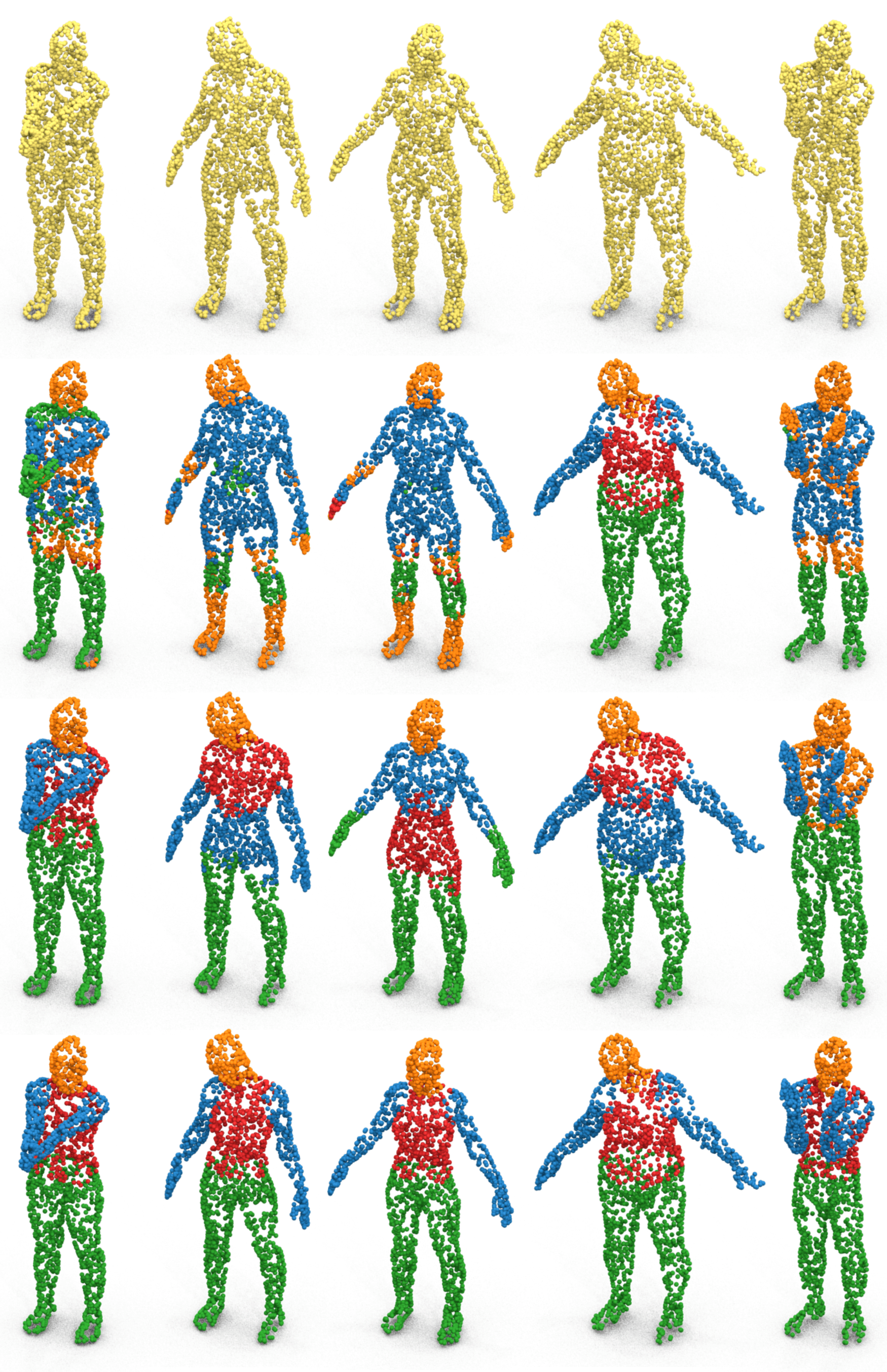}
    \put(-1,7.5){\rotatebox{90}{\footnotesize Ground-truth}}
    \put(-1,35){\rotatebox{90}{\footnotesize \acronym}}
    \put(-1,57){\rotatebox{90}{\footnotesize PointCLIPv2}}
    \put(-1,85){\rotatebox{90}{\footnotesize Input}}
\end{overpic}

\vspace{-3mm}
\caption{
Qualitative results on FAUST~\cite{Bogo:CVPR:2014}.
Top to bottom: input texture-less point cloud (coloured in yellow for visualisation purposes), PointCLIPv2 predictions, \acronym predictions, and ground-truth segmentation provided by SATR~\cite{abdelreheem2023satr}.
}
\label{fig:qual_faust}
\end{figure*}

Following our main paper, we show additional qualitative results on other datasets.
\cref{fig:qual_faust} shows qualitative results on FAUST~\cite{Bogo:CVPR:2014}, using annotations from SATR~\cite{abdelreheem2023satr}.
\cref{fig:qual_sonn} shows qualitative results on ScanObjectNN~\cite{uy2019scanobjectnn} in the most challenging OBJ-BG setting.
Objects such as ``bed'' or ``sofa'' pose challenges in distinguishing between the individual parts due to overlapping geometry or intricate designs.
Moreover, the real-world point clouds in ScanObjectNN~\cite{uy2019scanobjectnn} are noisy and can contain several occlusions, making it difficult to separate them into distinct parts.

    \section{Resources used}%
\label{sec:hardware}

Our training-free method does not require extensive computational resources and is designed to be computationally efficient. We run all our experiments on a consumer desktop NVIDIA RTX 3060 GPU with 12GB of VRAM and a laptop NVIDIA RTX 2070 Super Max-q with 8GB of VRAM.
Evaluation time on the NVIDIA RTX 3060 GPU took approximately:
40 seconds on FAUST~\cite{abdelreheem2023satr};
1 hour on ScanObjectNN~\cite{uy2019scanobjectnn};
3 hours on PartNetE~\cite{liu_partslip_2023};
9 hours on ShapeNetPart~\cite{yi2016shapenetpart};
12 hours on PartNet~\cite{Mo_2019_CVPR}.
All the reported inference times are doubled if running the inference on the consumer laptop with the NVIDIA RTX 2070 GPU with 8GB of VRAM.
Further optimisations, such as pre-rendering views for all objects, can be introduced to lower test times.
However, they can take up a large amount of storage space.

    {\small
    \bibliographystyle{wacv25toolkit/ieee_fullname}
    \bibliography{supp}
    }